\documentclass{article}

\usepackage[most]{tcolorbox}
\usepackage[preprint]{neurips_2026}

\newtcolorbox{promptbox}[1]{
  enhanced,
  colback=gray!10,
  colframe=purple!60!black,
  coltitle=white,
  fonttitle=\bfseries,
  title={#1},
  boxrule=0.8pt,
  arc=2pt,
  left=8pt,
  right=8pt,
  top=6pt,
  bottom=6pt
}
\newtcolorbox{insightbox}{
  colback=gray!5,
  colframe=gray!50,
  boxrule=0.4pt,
  arc=2pt,
  left=6pt,
  right=6pt,
  top=4pt,
  bottom=4pt,
  breakable
}

\usepackage{times}
\usepackage{algorithm}
\usepackage{algpseudocode}
\usepackage{hyperref}
\usepackage{url}
\usepackage{amsmath,amssymb}

\newtheorem{theorem}{Theorem}[section]
\newtheorem{proposition}[theorem]{Proposition}

\newtheorem{corollary}[theorem]{Corollary}

\usepackage{enumitem}
\usepackage{natbib}
\usepackage{booktabs}
\usepackage{multirow}
\usepackage{graphicx}
\usepackage{subcaption}
\usepackage{placeins}
\usepackage{float}
\usepackage{listings}
\usepackage{xcolor}

\title{When Less Latent Leads to Better Relay: Information-Preserving Compression for Latent Multi-Agent LLM Collaboration}

\author{
Yiping Li,\quad Zhiyu An,\quad Wan Du \\
Department of Computer Science and Engineering\\
University of California, Merced\\
Merced, CA 95343 \\
\texttt{\{yipingli, zan7, wdu3\}@ucmereced.edu}
}

\begin{document}
\maketitle

\begin{abstract}
Multi-agent LLM systems are moving beyond discrete-token messages toward richer relays that preserve internal state.
Recent work such as LatentMAS transmits full key-value (KV) caches between agents but pays a high memory and communication cost.
We adapt KV-cache eviction to this setting and introduce \textbf{Orthogonal BackFill (OBF)}, which injects a low-rank residual from the discarded KV states back into the retained ones, orthogonal to what is already kept.
With only $9.9\%$--$20.2\%$ of the prompt KV retained, compressed relay cuts bandwidth by $4.7\times$ and GPU memory by $8\%$ at under $5\%$ wall-clock overhead, and stays close to full relay in accuracy across nine benchmarks, ahead of it on several.
OBF matches or improves over headwise eviction on all nine, and its gain is proportional to the accuracy gap eviction opens against full relay ($r{=}0.78$ across three model scales), so it gives back part of what eviction takes.
Code is available at \url{https://github.com/markli404/When-Less-Latent-Leads-to-Better-Relay}.
\end{abstract}

\section{Introduction}
\label{sec:intro}

Large language model (LLM) based multi-agent systems (MAS) rely on communication to coordinate reasoning, critique intermediate outputs, and refine final decisions.
Early multi-agent frameworks use natural language as the communication medium, where agents exchange explicit messages under predefined roles or protocols \citep{li2023camel, hong2024metagpt, chen2024optima}.
Text-based communication is lossy. The model produces a probability distribution over the vocabulary, but only one sampled token is transmitted, discarding all other plausible alternatives.
This has pushed work toward richer communication channels that preserve more of the model's internal computation \citep{pham2024cipher, tang2025statedelta}.
LatentMAS (LMAS) pairs latent reasoning with lossless inter-agent communication. Each agent reasons in continuous latent space and relays its full key-value (KV) cache directly to the next agent, so downstream reasoning can continue from internal states rather than from text alone \citep{zou2025latentmas}.
That fidelity is expensive. Relaying full KV caches consumes memory and bandwidth that grow with the number of agents, rounds, and prompt tokens.

A natural solution is to compress the communicated KV states, as recent work on efficient LLM inference suggests that not all cached states are equally important. Methods such as StreamingLLM, H$_2$O, and Scissorhands show that selective retention or eviction of KV states can reduce memory usage while preserving generation quality in single-agent settings \citep{xiao2024streamingllm, zhang2023h2o, liu2023scissorhands}. However, these methods target online cache management during token-by-token generation within a single agent.
In LatentMAS-style communication, compression instead occurs at the relay boundary between agents, where the relevant question is whether the transmitted states remain useful for \emph{downstream continuation by another agent}.

In this work, we adapt KV-cache eviction methods to inter-agent relay and propose \textbf{Orthogonal BackFill (OBF)}. Under a fixed relay budget, we aim to make each retained state carry as much signal as possible for the downstream agent. OBF extracts from the discarded value states a low-rank component orthogonal to the retained ones and injects it into the retained values, so that information complementary to the retained set is preserved.

Our experiments show that compressed relay can match full KV relay at a fraction of the bandwidth, and that what governs relay quality is whether the preserved state stays useful to the downstream agent rather than how much of it survives. We further study how selector granularity and OBF design affect this trade-off across tasks.

Our contributions are as follows.
\begin{itemize}
    \item We give, to our knowledge, the first adaptation of KV-cache eviction to multi-agent latent relay, together with an explicit role decomposition of the cache that makes the eviction operator interpretable.

    \item We propose \textbf{OBF}, which mitigates information loss from eviction by injecting a low-rank vector extracted from the discarded value states and orthogonal to the retained span.

    \item We show empirically that compressed relay remains competitive with full relay while relaying only $9.9\%$--$20.2\%$ of the prompt KV across nine benchmarks, and surpasses full relay on a subset of the math and coding benchmarks.
\end{itemize}

\section{Related Work}
\label{sec:related}

\paragraph{Why Multi-Agent Systems.}
MAS outperform single-agent baselines on complex tasks such as software engineering, mathematical reasoning, and deliberative evaluation~\citep{hong2024metagpt, du2023multiagentdebate, chan2023chateval, liang2024encouraging}. They typically distribute reasoning across roles, prompts, or subtasks rather than forcing the entire problem into a single trajectory~\citep{li2023camel, wu2023autogen, zhou2023least}. All of these frameworks communicate through natural language, even though text is a lossy compression of each agent's internal state and is not strictly necessary for inter-agent exchange.

\paragraph{From Language to Latent Communication.}
Early MAS frameworks such as \textbf{CAMEL}~\cite{li2023camel} and \textbf{MetaGPT}~\cite{hong2024metagpt} organize collaboration through natural-language exchanges under predefined roles and workflows, while later systems such as \textbf{Optima}~\cite{chen2024optima} train agent policies to shorten messages and reduce redundant exchanges. More recent work moves beyond text-only communication and explores richer carriers for inter-agent information exchange. \textbf{CIPHER}~\cite{pham2024cipher} communicates through expected output embeddings, \textbf{StateDelta}~\cite{tang2025statedelta} augments language with hidden-state changes, and \textbf{COCONUT}~\cite{hao2025coconut} studies latent-space reasoning without decoding intermediate steps into text. These works motivate a broader shift from token-level to latent-state communication. Within this shift, we target the efficiency problem created by full KV-state relay, as proposed in LMAS.

\paragraph{LatentMAS and KV Relay.}
Our setting is most directly related to LMAS \cite{zou2025latentmas}, a training-free framework for pure latent collaboration among LLM agents. Instead of passing decoded intermediate messages, LMAS transfers the full KV cache from one agent to the next, so downstream reasoning continues directly from the predecessor's internal state. This yields higher fidelity than text-based communication, but it also turns communicated KV into the dominant communication object. We address this bottleneck directly by compressing at the relay boundary rather than modifying the reasoning procedure itself.

\paragraph{KV Compression, Eviction, and Compensation.}
A separate line of work studies how to reduce KV-cache cost in long-context inference. Quantization methods such as \textbf{KVQuant}~\cite{hooper2025kvquant} reduce the footprint of retained KV tensors, while eviction methods such as \textbf{H$_2$O}~\cite{zhang2023h2o}, \textbf{StreamingLLM}~\cite{xiao2024streamingllm}, and \textbf{Scissorhands}~\cite{liu2023scissorhands} selectively retain only the most useful past states. More recent methods mitigate the information loss from hard eviction through compensation or merging. \textbf{RazorAttention}~\cite{tang2024razorattention} summarizes removed context through compensation tokens in non-retrieval heads, while \textbf{CaM}~\cite{zhang2024cam} and \textbf{WeightedKV}~\cite{yuan2025weightedkv} merge discarded information into retained states instead of removing it outright. All of these target single-agent online inference, where compression interleaves with token generation rather than acting at an inter-agent boundary.

\paragraph{Positioning.}
In contrast, we compress at the relay step in a single block, removing multiple tokens at once. The criterion is whether the compressed states remain useful for downstream continuation by another agent. OBF further differs from these compensation methods in two ways. First, it injects only the residual orthogonal to the retained span, avoiding double counting of information already accessible through retained tokens. Second, it broadcasts a single low-rank correction to all retained values, rather than merging individual deleted tokens or appending compensation tokens.

\section{Preliminaries}

\subsection{KV Cache Notation}
\label{subsec:kv_cache}

For an autoregressive Transformer decoder processing a token sequence $Y=\{y_1,\dots,y_T\}$, at each layer $l \in \{1,\dots,L_{\text{layer}}\}$ with $H$ attention heads the self-attention cache contains
\[
\mathbf{K}^{(l)} \in \mathbb{R}^{T \times H \times d_k},
\qquad
\mathbf{V}^{(l)} \in \mathbb{R}^{T \times H \times d_v}.
\]
The per-head slices are
\[
\mathbf{K}^{(l,h)} := \mathbf{K}^{(l)}_{:,h,:} \in \mathbb{R}^{T \times d_k},
\qquad
\mathbf{V}^{(l,h)} := \mathbf{V}^{(l)}_{:,h,:} \in \mathbb{R}^{T \times d_v},
\qquad h \in \{1,\dots,H\}.
\]

\paragraph{Token-level and full-cache views.}
At a fixed layer, the cache forms a token-indexed sequence $\mathbf{S}^{(l)} := \{(\mathbf{K}^{(l)}_{t,:,:}, \mathbf{V}^{(l)}_{t,:,:})\}_{t=1}^{T}$. The full cache is $\mathrm{KV}(Y) = \{\mathbf{K}^{(l)}, \mathbf{V}^{(l)}\}_{l=1}^{L_{\text{layer}}}$, and we write $\mathbf{S} := \mathrm{KV}(Y)$ when treating it as a single token-indexed object, with $\mathbf{S}[t]$ the state at position $t$.

\paragraph{Index selection and concatenation.}
Any retained cache is a subsequence of $\mathbf{S}$ selected by an index list. For an ordered index list $I=(i_1,\dots,i_m)$ with $1\le i_1<\cdots<i_m\le T$, $\mathbf{S}_{I}[r] = \mathbf{S}[i_r]$ and $\mathbf{S}_I \in \mathbb{R}^{m \times D_{KV}}$. Row-wise concatenation $[\mathbf{S}_{I_a};\mathbf{S}_{I_b}]$ corresponds to index list $I_a \Vert I_b$.

\subsection{Agentic Communication in LMAS}
In LMAS, agents communicate through latent messages represented by Transformer KV-cache states produced during autoregressive latent inference. Let $\pi_i$ denote the $i$-th agent in the communication chain. After completing its local computation, $\pi_i$ transmits a message $\mathcal{M}_i$, where $\mathcal{M}_i = [\mathcal{M}_{i-1}; \mathrm{KV}(Y_i)]$.

Here, $\mathrm{KV}(Y_i)$ denotes the token-level KV-cache states produced by agent $\pi_i$. We decompose these states as $\mathrm{KV}(Y_i) = [\mathbf{S}_{P_i}; \mathbf{S}_{G_i}]$, where $\mathbf{S}_{P_i}$ is induced by the local prompt $P_i$ and $\mathbf{S}_{G_i}$ by autoregressive latent generation.

The receiver $\pi_{i+1}$ continues decoding by using $\mathcal{M}_i$ as \texttt{PastKV}. Because each $\mathrm{KV}(Y_k)$ is appended in full, message length grows linearly with the number of agents. We formalize the resulting compression problem in Section~\ref{subsec:formulation}.

\section{Methodology}
\label{sec:method}

We extend single-agent KV-cache eviction to multi-agent latent communication. Building on a four-part cache decomposition, we adapt single-agent eviction rules such as StreamingLLM and H$_2$O to the relay setting, and propose \textbf{OBF} to address the limitations of direct eviction.

\subsection{Problem Formulation}
\label{subsec:formulation}

We formulate inter-agent relay as a KV compression problem. Since $|\mathcal{M}_i|$ grows linearly with the number of agents under full relay ($|\mathcal{M}_i| \propto \sum_{k=0}^{i} |\mathrm{KV}(Y_k)|$), each agent compresses its prompt states before transmission. Let $\Phi$ denote a compression operator acting on the prompt states $\mathbf{S}_{P_i}$, optionally conditioned on the reasoning states $\mathbf{S}_{G_i}$:
\[
    \mathcal{M}_i \;=\; [\mathcal{M}_{i-1};\ \Phi(\mathbf{S}_{P_i};\,\mathbf{S}_{G_i});\ \mathbf{S}_{G_i}].
\]
Under a per-agent budget that retains a ratio $\rho \in (0,1]$ of the prompt positions, our goal is to design $\Phi$ that maximizes downstream task accuracy at any given $\rho$.

\subsection{From Single-Agent KV-Cache Eviction to Multi-Agent Cache Decomposition}
\label{subsec:decomposition}

Existing single-agent eviction methods implicitly decompose the KV cache into three functional parts: an \textbf{attention sink} of initial tokens that stabilize the softmax distribution \cite{xiao2024streamingllm}; a \textbf{candidate pool} of middle tokens targeted for importance-based eviction \cite{zhang2023h2o,liu2023scissorhands}; and a \textbf{local window} of recent tokens preserving local context \cite{xiao2024streamingllm,zhang2023h2o}.

\begin{figure}[t]
    \centering
    \includegraphics[width=\linewidth]{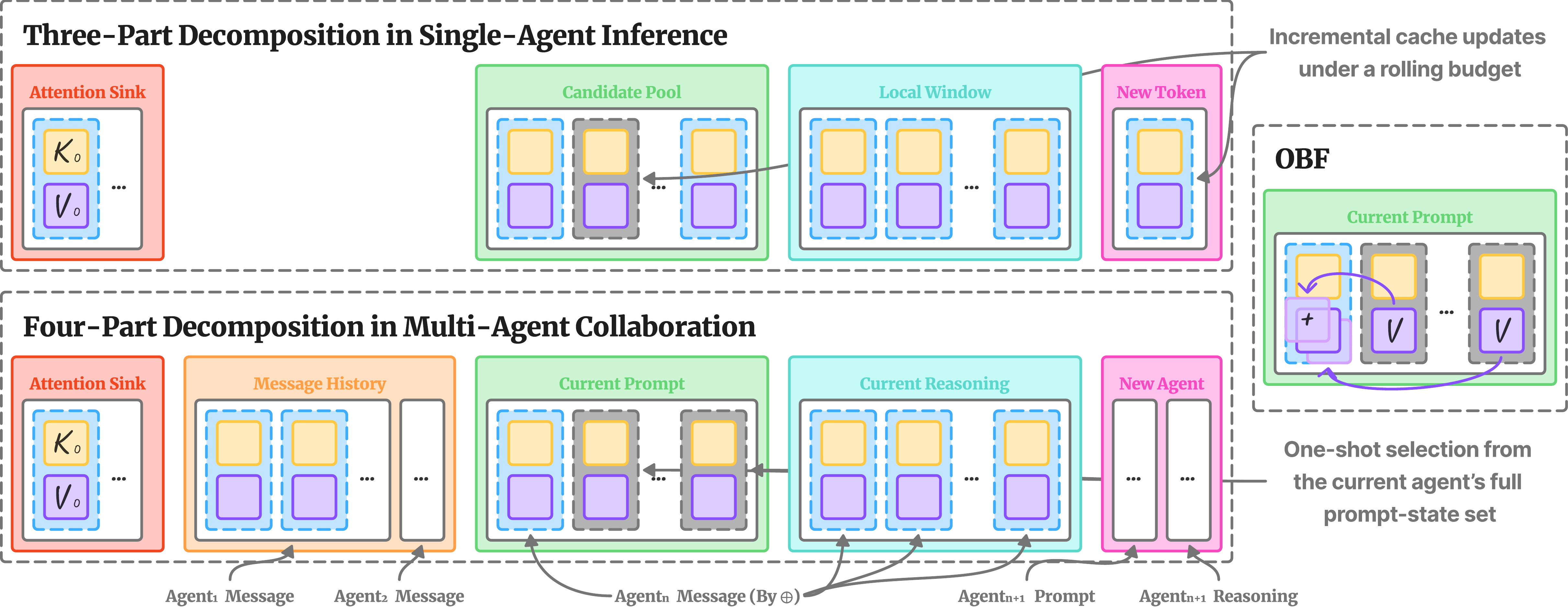}
    \caption{\textbf{Overview of role decomposition and the OBF compression operator.} Top: single-agent inference uses three roles (sink, candidate pool, local window) and updates the cache incrementally under a rolling budget. Bottom: our multi-agent setup adds inherited message history and selects retained prompt states in a single shot. Right inset: OBF previews residual injection into the retained values. See Figure~\ref{fig:selection_methods} for eviction granularities and Figure~\ref{fig:obf_figure} for OBF mechanics.}
    \label{fig:kv_decomposition}
\end{figure}

In multi-agent collaboration, latent memory is no longer produced by a single homogeneous stream. Instead, it accumulates across agent boundaries as alternating prompt and generation segments, e.g., $[\mathbf{S}_{P_1}; \mathbf{S}_{G_1}; \mathbf{S}_{P_2}; \mathbf{S}_{G_2}; \dots] = [\mathcal{M}_{i-1}; \mathbf{S}_{P_i}; \mathbf{S}_{G_i}]$.
This structure makes the candidate region heterogeneous. Inherited memory must be preserved as transmitted context, so we separate it from the current agent's local prompt context, and the role of preserving recent context is now filled by the current agent's active latent reasoning states. We therefore use a four-part decomposition (Figure~\ref{fig:kv_decomposition}):
\begin{enumerate}
    \item \textbf{Attention Sink ($\mathcal{S} \subset \mathcal{M}_{i-1}$):} A persistent anchor, originated by $\pi_0$ and inherited across the chain, that maintains numerical stability.

    \item \textbf{Inherited Message History ($\mathcal{H} \subset \mathcal{M}_{i-1}$):} The compressed historical context transmitted by upstream agents.

    \item \textbf{Current Prompt Context ($\mathcal{P} = \mathbf{S}_{P_i}$):} The latent states induced by the task-specific instructions of the current agent.

    \item \textbf{Current Latent Reasoning ($\mathcal{G} = \mathbf{S}_{G_i}$):} The latent states generated by the current agent's ongoing reasoning, which form the immediate causal basis for $\pi_i$'s generation.
\end{enumerate}

\subsection{Eviction Baselines in LatentMAS (LMAS)}
\label{subsec:baselines}

We adapt standard KV-cache eviction rules to LMAS by compressing only the current agent's local prompt states $\mathcal{P}$, while the current reasoning states $\mathcal{G}$ and the inherited memory $\mathcal{M}_{i-1}$ are forwarded unchanged. The message transmitted by agent $\pi_i$ is
\[
    \mathcal{M}_i = [\mathcal{M}_{i-1} ;\ \Phi(\mathcal{P}\,;\,\mathcal{G}) ;\ \mathcal{G}],
\]
where $\Phi(\mathcal{P}\,;\,\mathcal{G})$ returns a retained (and optionally augmented) subset of the prompt states, conditioned on the reasoning states $\mathcal{G}$ that determine which prompt positions are informative. We take $\mathcal{M}_0=\mathcal{S}$, so the shared sink instantiated by $\pi_0$ is carried forward inside $\mathcal{M}_{i-1}$ thereafter.

\subsubsection{LMAS-Sink: Generation-Only Retention}

LMAS-Sink instantiates the retention rule with a simple policy that preserves only the shared \textbf{Attention Sink} $\mathcal{S}$ (initialized by the first agent) and treats each local prompt $\mathcal{P}_i$ as transient context. The retention rule is $\Phi_{\mathrm{Gen}}(\mathcal{P}_i\,;\,\mathcal{G}_i) = \emptyset$, so only $\mathcal{G}_i$ is appended to $\mathcal{M}_{i-1}$ at each round.

\subsubsection{LMAS-Attn: Attention-Based Selection}
\label{subsec:h2o_selection}

LMAS-Attn retains prompt tokens that receive high attention from the current latent reasoning states and evicts the rest under a fixed budget, following the importance-based retention principle of \citep{zhang2023h2o}.

\paragraph{Attention mass from the current reasoning states.}
For a given layer $l$ and a given attention head $h$, let $a_{t,j}^{(l,h)}$ denote the attention weight from current reasoning position $t \in \mathcal{G}_i$ to prompt position $j \in \mathcal{P}_i$. We define the attention mass assigned to prompt position $j$ as
\begin{equation}
A_j^{(l,h)}
\;=\;
\sum_{t \in \mathcal{G}_i} a_{t,j}^{(l,h)} .
\label{eq:h2o_mass_head}
\end{equation}

\begin{figure}[htbp]
    \centering
    \includegraphics[width=0.95\linewidth]{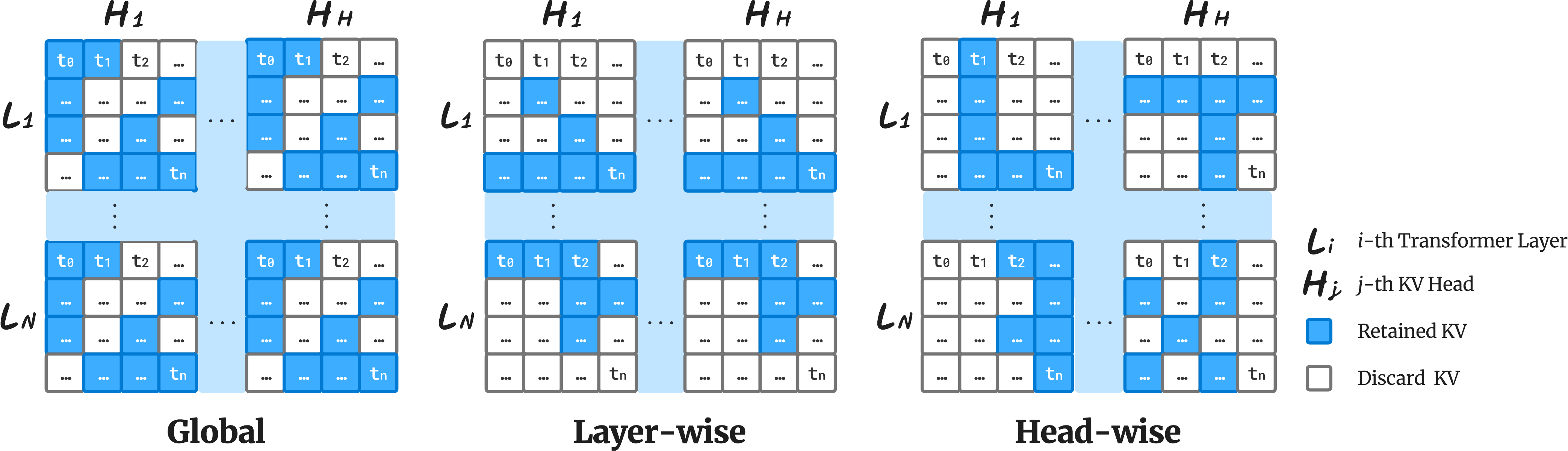}
    \caption{\textbf{Comparison of KV cache eviction granularities.}
    (1) \textbf{Layerwise}: shared across heads within each layer; and
    (2) \textbf{Headwise}: independent for each KV head and each layer.
    Qualitative examples of the retained prompt tokens under different granularities are shown in Appendix~\ref{app:token_selection}.}
    \label{fig:selection_methods}
\end{figure}

\paragraph{Retention rule.}
Let $\Omega_i \subseteq \mathcal{P}_i$ denote the eligible prompt indices (excluding padding and, when $i=1$, the shared sink $\mathcal{S}$). Under headwise and layerwise granularity (Figure~\ref{fig:selection_methods}), the retained prompt set is
\[
\mathcal{P}_{\mathrm{keep},i}^{\mathrm{head}} = \operatorname{TopK}_B\!\big(\{A_j^{(l,h)}\}_{j\in\Omega_i}\big),
\qquad
\mathcal{P}_{\mathrm{keep},i}^{\mathrm{layer}} = \operatorname{TopK}_B\!\Big(\big\{\textstyle\sum_{h=1}^{H} A_j^{(l,h)}\big\}_{j\in\Omega_i}\Big),
\]
where $B$ corresponds to \texttt{kv\_budget} and ties are broken deterministically. Writing $\mathcal{P}_{\mathrm{keep},i}$ for either choice, the retention rule is $\Phi_{\mathrm{Attn}}(\mathcal{P}_i\,;\,\mathcal{G}_i) = \mathcal{P}_{\mathrm{keep},i}$.

\subsection{Orthogonal BackFill: Residual Principal-Subspace Injection for Hard Eviction}
\label{subsec:obf}

\begin{figure}[htbp]
    \centering
    \includegraphics[width=1\linewidth]{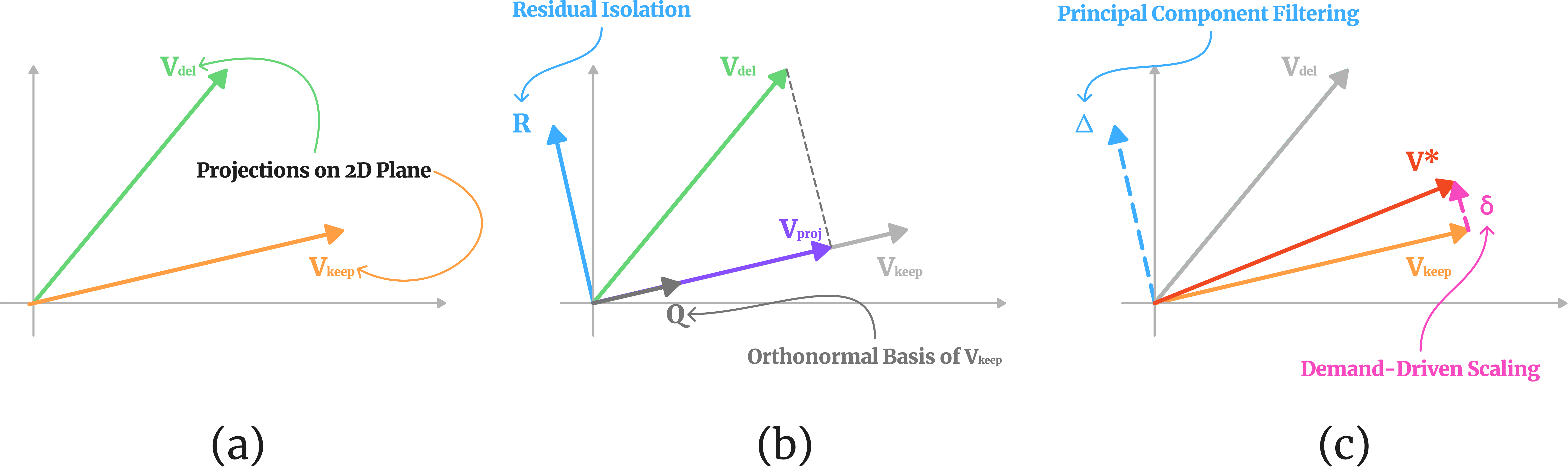}
\caption{\textbf{Geometric illustration of Orthogonal BackFill (OBF).} 2D sketch (not a literal high-dimensional mapping). From discarded values $\mathbf{V}_\mathrm{del}$, OBF isolates the residual $\mathbf{R}$ orthogonal to the retained span $\mathbf{Q}$ (Eq.~\ref{eq:pca_extra_residual}), keeps its principal subspace $\Delta$ (Eq.~\ref{eq:delta}), scales it by the demand ratio (Eq.~\ref{eq:demand_ratio}), and fuses the result $\delta$ uniformly into $\mathbf{V}_\mathrm{keep}$ (Eq.~\ref{eq:pca_extra_injection_uniform}), adding information orthogonal to the retained span without changing which positions are kept.}
    \label{fig:obf_figure}
\end{figure}

In LMAS, eviction is applied block-wise to each agent's local prompt states, so a single compression step can delete a large block of prompt KV at once. We propose \textbf{Orthogonal BackFill (OBF)}, which extracts a low-rank residual from the deleted values and injects it back into the retained ones (illustrated geometrically in Figure~\ref{fig:obf_figure}). Given any base selector producing a retained prompt set $\mathcal{P}_{\mathrm{keep},i}\subset\mathcal{P}_i$ (e.g., the LMAS-Attn rule above), let $\mathcal{P}_{\mathrm{del},i}=\mathcal{P}_i\setminus\mathcal{P}_{\mathrm{keep},i}$ denote the deleted positions.

\paragraph{Retained and deleted value states.}
For each layer $l$ and KV head $h$, let $\mathbf{V}^{(l,h)}\in\mathbb{R}^{L_{\mathrm{prompt}}\times d}$ denote the value states of the prompt tokens, whose $t$-th row is $\mathbf{V}_t^{(l,h)}$. All of the following operations are applied independently for each $(l,h)$, so we omit these superscripts for the remainder of this section. We treat $\mathcal{P}_{\mathrm{keep}}$ and $\mathcal{P}_{\mathrm{del}}$ as ordered index lists specifying the kept and deleted portions of the value matrix, and write $\mathbf{V}_{\mathrm{keep}} = \mathbf{V}[\mathcal{P}_{\mathrm{keep}},:]$ and $\mathbf{V}_{\mathrm{del}} = \mathbf{V}[\mathcal{P}_{\mathrm{del}},:]$.

\paragraph{Residual relative to the retained span.}
\label{sec:projection}
We construct an orthonormal basis for $\mathrm{span}(\mathbf{V}_{\mathrm{keep}})$ via a reduced QR decomposition on $\mathbf{V}_{\mathrm{keep}}^{\top}$:
\begin{equation}
\mathbf{Q}=\mathrm{orth}\!\big(\mathbf{V}_{\mathrm{keep}}^{\top}\big)\in\mathbb{R}^{d\times r},
\qquad
\mathbf{Q}^{\top}\mathbf{Q}=\mathbf{I}_r,
\qquad
r\le \min(B,d).
\end{equation}
Subtracting from $\mathbf{V}_{\mathrm{del}}$ its projection onto the span of retained values yields the residual
\begin{equation}
\mathbf{R}
=
\mathbf{V}_{\mathrm{del}}
-
(\mathbf{V}_{\mathrm{del}}\mathbf{Q})\mathbf{Q}^{\top}.
\label{eq:pca_extra_residual}
\end{equation}
The residual $\mathbf{R}$ is therefore orthogonal to the span of retained values.

\begin{insightbox}
\textbf{Why inject only the orthogonal component?} After compression, attention that would have been assigned to deleted tokens is redistributed to the retained ones, so any in-span component of the deleted values is partially recovered through the retained states. Injecting the in-span component again would double count this information. OBF therefore injects only the residual orthogonal to the span of retained values.
\end{insightbox}

\paragraph{Injection direction.}
\label{sec:pca}
We first denoise $\mathbf{R}$ by applying Singular Value Decomposition (SVD) and retaining its top $k$ right singular vectors as the residual principal basis $\mathbf{C}\in\mathbb{R}^{k\times d}$, where $k=\min(\texttt{pca\_rank},\ \mathrm{rank}(\mathbf{R}))$. We fix $\texttt{pca\_rank}=4$ for L-OBF and $\texttt{pca\_rank}=2$ for H-OBF throughout (Section~\ref{sec:experiments}); Appendix~\ref{app:pca_rank} sweeps the rank. We then summarize the deleted residual content by an attention-weighted average: writing $\mathcal{P}_{\mathrm{del}}=\{t_j\}_{j=1}^{N}$ for the deleted positions in the same order as the rows of $\mathbf{R}$,
\begin{equation}
\bar{\mathbf{r}}
=
\sum_{j=1}^{N}w_j\,\mathbf{R}[j,:]
\in\mathbb{R}^{d},
\qquad
w_j
=
\frac{A_{t_j}}{\sum_{t\in\mathcal{P}_{\mathrm{del}}}A_t+\epsilon},
\label{eq:pca_extra_rbar}
\end{equation}
where $A_t$ denotes the attention mass $A_t^{(l,h)}$ from Eq.~\ref{eq:h2o_mass_head}. Projecting $\bar{\mathbf{r}}$ onto the residual principal subspace yields the injection direction
\begin{equation}
\Delta=\big(\bar{\mathbf{r}}\mathbf{C}^{\top}\big)\mathbf{C}\in\mathbb{R}^{d}.
\label{eq:delta}
\end{equation}

\begin{insightbox}
\textbf{Why aggregate via PCA?} A single discarded token is unimportant by construction, but many of them taken together can still encode meaningful structure. PCA on the residual matrix extracts the dominant direction these tokens form while dropping each token's noise. The injected residual then acts as a gentle steering of retained values rather than a large modification.
\end{insightbox}

\paragraph{Fast SVD alternative.}
Since $\texttt{pca\_rank}$ is small ($k \leq 32$ in all experiments) but $\mathbf{R}$ can have thousands of rows, the exact SVD in Eq.~\ref{eq:delta} can be replaced by a Gram-matrix subspace iteration: form the $d\times d$ Gram matrix $\mathbf{R}^{\top}\mathbf{R}$ and run a few iterations of subspace iteration to extract the top-$k$ right singular vectors directly. This alternative reproduces the injection direction $\Delta$ to numerical precision at roughly $5\times$–$8\times$ lower wall-clock cost. Main-text results use the exact SVD; timing and accuracy verification of the fast alternative are in Appendix~\ref{app:fast_svd}.

\paragraph{Dynamic scaling by deleted-versus-retained demand.}
\label{sec:dynamic_scale}
We then scale the injection by the deleted-versus-retained demand ratio:
\begin{equation}
\delta
=
\frac{\tilde{A}_{\mathrm{del}}}{\tilde{A}_{\mathrm{keep}}+\epsilon}\,\Delta,
\qquad
\text{where }
\tilde{A}_{\mathrm{keep}}=\sum_{t\in\mathcal{P}_\mathrm{keep}}A_t,
\quad
\tilde{A}_{\mathrm{del}}=\sum_{t\in\mathcal{P}_\mathrm{del}}A_t.
\label{eq:demand_ratio}
\end{equation}

\begin{insightbox}
\textbf{Why scale by demand?} The PCA direction tells us \emph{what} to inject; the demand ratio tells us \emph{how much}. When deleted positions collectively receive little attention ($\tilde{A}_{\mathrm{del}}$ small), the missing signal matters little to the downstream query and OBF attenuates the injection accordingly. Strong backfill therefore requires both a coherent residual direction and high deleted demand.
\end{insightbox}

\paragraph{Injection into retained prompt values.}
OBF modifies \emph{only} the value states and leaves keys unchanged.\footnote{If $\mathcal{P}_{\mathrm{del}}=\emptyset$, $k=0$, or $\|\mathbf{R}\|_F$ is negligible, OBF is skipped for that $(i,l,h)$; we use $\epsilon=10^{-12}$ throughout for numerical stability.} In this work, we use a uniform injection strategy: we add the same residual vector $\delta$ to all retained value states,
\begin{equation}
\mathbf{V}_t^* \leftarrow \mathbf{V}_t + \delta,
\qquad \forall\, t\in\mathcal{P}_{\mathrm{keep}}.
\label{eq:pca_extra_injection_uniform}
\end{equation}
The retention rule is $\Phi_{\mathrm{OBF}}(\mathcal{P}_i\,;\,\mathcal{G}_i) = \mathcal{P}_{\mathrm{keep},i} \oplus \delta_i$, where $\oplus$ denotes adding $\delta_i$ to the values at retained positions while keeping keys unchanged. We ablate each of the three design choices (projection, filtering, and scaling) in Appendix~\ref{app:obf_ablation}.

\section{Experiments}
\label{sec:experiments}

\subsection{Setup}

\paragraph{Tasks and Datasets.}
Following the evaluation protocol of LatentMAS~\cite{zou2025latentmas}, we evaluate on a suite of benchmarks organized into four groups:
(i) \textbf{Math}: GSM8K~\cite{cobbe2021gsm8k} and AIME 2024/2025~\cite{aime2024,aime2025};
(ii) \textbf{Expert QA}: GPQA~\cite{rein2023gpqa} and MedQA~\cite{jin2021medqa};
(iii) \textbf{Commonsense QA}: ARC (Easy/Challenge)~\cite{clark2018arc};
(iv) \textbf{Coding}: MBPP-Plus and HumanEval-Plus~\cite{liu2023plus}.

\paragraph{Implementation Details.}
We build on the official LatentMAS implementation~\cite{zou2025latentmas} with Qwen3-4B~\cite{qwen3} as the backbone, evaluating only the sequential multi-agent setting. All compared methods share the backbone, prompts, agent-chain configuration, and decoding setup, differing only in the compression operator. The inter-agent communication protocol and prompt templates are in Appendices~\ref{app:latentmas_setting} and~\ref{app:prompt}. Algorithm pseudocode is in Appendix~\ref{app:algorithm}. All hyperparameter settings are listed in Appendix~\ref{app:hyper}, and hardware details in Appendix~\ref{app:hardware}.

\paragraph{Compared Methods.}
\begin{itemize}[leftmargin=1.2em]
    \item \textbf{Full-KV LatentMAS (Full).} Each agent relays the complete KV cache of the full history.
    \item \textbf{LMAS-Attn (H).} Attention-based hard eviction inspired by importance-based retention~\cite{zhang2023h2o}. For each $(l,h)$, we rank prompt positions by that head's attention mass from $\mathcal{G}_i$ and keep the top-$B$ independently, so retained positions vary across heads and layers.
    \item \textbf{LMAS-Attn (L).} A layerwise variant that aggregates attention across heads within each layer before selection; all heads in a layer share the retained positions.
    \item \textbf{LMAS-Attn + OBF (H-OBF, L-OBF).} The H and L selectors followed by Orthogonal BackFill injection (Section~\ref{subsec:obf}) into the retained values. We fix $\texttt{pca\_rank}=4$ for L-OBF and $\texttt{pca\_rank}=2$ for H-OBF throughout the paper, chosen on Qwen3-4B before any Qwen3-8B or Qwen3-14B run; a per-dataset rank sweep is in Appendix~\ref{app:pca_rank}.
    \item \textbf{LMAS-Sink (Gen).} A sink and generation-only relay~\cite{xiao2024streamingllm} that preserves a global attention sink and forwards only each agent's newly generated KV states. LMAS-Sink has no H/L variant because it performs no position-wise selection.
\end{itemize}

\paragraph{Metrics and KV Budget.}
We report \textbf{accuracy} using the standard metric of each benchmark. For efficiency, we impose a fixed \textbf{KV token budget} $B$ on each agent's relayed prompt states $\mathcal{P}_i$, retaining at most $B$ positions per agent. We also report the \textbf{total relayed prompt length} $L$, summed across agents and averaged over samples, and the \textbf{compression ratio} $\rho$, defined as the fraction of prompt KV states retained at the end of a rollout relative to those fed in across all rounds.

\begin{table*}[!htbp]
\centering
\footnotesize
\setlength{\tabcolsep}{4pt}
\renewcommand{\arraystretch}{1.08}
\resizebox{\textwidth}{!}{%
\begin{tabular}{l|ccc|cc|cc|cc}
\toprule
Method & \multicolumn{3}{c|}{Math} & \multicolumn{2}{c|}{Expert QA} & \multicolumn{2}{c|}{Commonsense QA} & \multicolumn{2}{c}{Coding} \\
\cmidrule(lr){2-4} \cmidrule(lr){5-6} \cmidrule(lr){7-8} \cmidrule(lr){9-10}
 & GSM8K & AIME24 & AIME25 & GPQA & MedQA & ARC-E & ARC-C & MBPP+ & HumanEval+ \\
\midrule
$L$ / $\rho$ (\%) & 524 / 19.1 & 663 / 15.1 & 819 / 12.2 & 942 / 10.6 & 1012 / 9.9 & 496 / 20.2 & 524 / 19.1 & 778 / 12.8 & 828 / 12.1 \\
\midrule
Full & \textbf{88.98} & 48.89 & 43.33 & \textbf{39.26} & \textbf{61.67} & \textbf{91.98} & \textbf{87.03} & 58.20 & \underline{68.90} \\
L & 86.58 & \textbf{53.33} & \textbf{52.22} & 33.16 & 59.89 & 90.59 & 85.13 & 59.35 & 65.45 \\
L-OBF & \underline{87.74} & \underline{52.22} & 45.56 & 31.98 & \underline{61.56} & 90.24 & 85.30 & 59.26 & \textbf{69.11} \\
H & 86.76 & 51.11 & 45.56 & 33.33 & 59.00 & 90.64 & 85.69 & 59.08 & 62.40 \\
H-OBF & 87.52 & \underline{52.22} & 46.67 & \underline{34.01} & 61.44 & \underline{90.78} & \underline{85.75} & \underline{59.70} & 68.29 \\
Gen & 82.21 & \textbf{53.33} & \underline{47.78} & 29.78 & 57.22 & 85.70 & 79.72 & \textbf{61.73} & 64.63 \\
\bottomrule
\end{tabular}%
}
\caption{\textbf{Downstream accuracy across benchmarks.} Each entry is the task's standard accuracy metric (\%). Best per benchmark is in \textbf{bold}.}
\label{tab:qwen4b_accuracy}
\end{table*}

\subsection{Results}

We report downstream accuracy (Table~\ref{tab:qwen4b_accuracy}) and defer a full end-to-end cost breakdown to Appendix~\ref{app:cost_breakdown}. Across benchmarks, $L$ ranges from $496$ to $1012$ tokens and $\rho$ from $9.9\%$ to $20.2\%$.

\paragraph{Downstream accuracy.}
We compare each OBF variant against its own eviction baseline at the same relay budget, H-OBF against H and L-OBF against L. \textbf{H-OBF improves over H on all nine benchmarks} (sign test $p{=}0.008$), and L-OBF improves over L on four. The asymmetry follows the selectors. Headwise eviction is the more lossy of the two at 4B, so H-OBF has more room to recover, while layerwise eviction already sits closer to Full. Against all six methods, at least one OBF variant places in the top two on eight of nine benchmarks, and on HumanEval+ L-OBF matches Full while relaying $12.1\%$ of the prompt KV.

\paragraph{Inference efficiency.}
Compressed relay produces roughly $10\%$ more final-agent tokens for only $5\%$ more decoding time because shorter relayed prompts make each step cheaper; a full cost breakdown is in Appendix~\ref{app:cost_breakdown}.

\paragraph{Scaling behavior.}
Extending the same protocol to Qwen3-8B and Qwen3-14B (Appendix~\ref{app:scaling}) shows one pattern at every scale: OBF recovers eviction damage in proportion to its magnitude (Appendix~\ref{app:recovery_law}). At 4B headwise eviction costs $2.73$~pp against Full on average and OBF recovers about half of it, which is why H-OBF improves over H on all nine benchmarks. At 8B and 14B the headwise gap falls to $0.45$ and $0.86$~pp, and H-OBF turns near-neutral.

\subsection{Additional Analyses}
\label{subsec:additional_analyses}
We defer supplementary analyses to the appendix. Appendix~\ref{app:obf_ablation} ablates OBF's design in two groups. Component ablations show that each internal step contributes to the full L-OBF result, and alternative designs such as attention-weighted injection and adaptive rank do not improve over OBF at the same relay budget. A comparison against external KV compression baselines (cache merging, int8/int4 quantization) is in Appendix~\ref{app:external_baselines}. A full end-to-end cost breakdown is in Appendix~\ref{app:cost_breakdown}, and a fast Gram-based SVD alternative for OBF is characterized in Appendix~\ref{app:fast_svd}. Sweeps over the relay budget and the PCA rank (Appendices~\ref{app:budget_sweep} and~\ref{app:pca_rank}) reveal no universally optimal setting and often non-monotonic per-dataset accuracy. Per-benchmark trade-offs between accuracy and generation cost (Appendix~\ref{app:downstream_cost}) suggest that excessive compression causes information loss, which manifests as the final agent generating more tokens to compensate. Layer-level diagnostics (Appendix~\ref{app:internal_diagnostics}) show that roughly half of the deleted information lies orthogonal to the retained span, and that L-OBF differs from H-OBF mainly in injection magnitude.

\section{Discussion}
Selective relay can match or exceed full relay because relay quality depends on usefulness to the next agent, not on the amount of state preserved. The benchmarks where Full leads (GSM8K~\cite{cobbe2021gsm8k}, GPQA~\cite{rein2023gpqa}, MedQA~\cite{jin2021medqa}, ARC-E, ARC-C~\cite{clark2018arc}) reward factual retention, where the prompt acts as evidence and any deletion costs information. The benchmarks where compressed variants lead (MBPP+, HumanEval+~\cite{liu2023plus}) demand code generation, where the prompt acts more like a scratchpad and redundant context distracts. One conjecture is that eviction acts as denoising on redundant prompt context; verifying this mechanism directly is left to future work.

H-OBF matches or improves over headwise eviction on all nine benchmarks, and L-OBF matches or improves over layerwise on four of nine (Table~\ref{tab:qwen4b_accuracy}, sign test $p{=}0.008$). This asymmetry follows the recovery law of Appendix~\ref{app:recovery_law}. OBF's gain scales with how much information the selector removed, and headwise removes more at 4B, so H-OBF has more to recover. Corollary~\ref{cor:obf_dominates} in Appendix~\ref{app:error_bound} shows that $\|\mathbf{e}_{\mathrm{OBF}}\| \le \|\mathbf{e}_{\mathrm{evict}}\|$, so OBF weakly dominates its selector by construction.

Layerwise and headwise selection show complementary strengths across tasks, and the stronger OBF variant depends on the underlying selector. This is consistent with prior evidence of head-level functional specialization~\cite{jo2020roles,li2023interpreting} and with KV compression methods that exploit head-level importance and task-aware adaptation~\cite{fu2024not,taniguchi2026adaptive,zhou2025dynamickv}. Task-adaptive selection across layers and heads is a natural next step.

Even where Full keeps the top accuracy, the case for compressed relay is its communication cost. Compressed variants cut per-round prompt KV bandwidth by $4.7\times$ for about $10\%$ more final-agent tokens and $5\%$ more inference time (Appendix~\ref{app:cost_breakdown}), and OBF's overhead drops further with the fast subspace iteration of Appendix~\ref{app:fast_svd}. Our timings understate the saving. The chain runs in a single process, the relay is an in-memory handoff, and Full is never charged for moving its cache, so the $4.7\times$ reduction is a property of the relayed volume rather than of our wall clock. Once agents sit far enough apart that communication is paid for, compressed relay is the rational default.

We view OBF as a first design point. The relay budget could also be allocated to the inherited history $\mathcal{H}$ or the latent reasoning states $\mathcal{G}$ with per-region compression rules. Realizing such extensions would benefit from sharper mechanistic understanding of which retained KV components drive downstream reasoning.

\section{Limitations}
\label{sec:limitations}

Our results use the Qwen3 dense grouped-query attention family (4B, 8B, and 14B) within the sequential four-agent chain of~\cite{zou2025latentmas}. Extension to other backbone families (MoE, non-GQA) and to other multi-agent topologies remains to be verified. Several OBF design choices (PCA rank, demand-ratio scaling, uniform injection) are empirical defaults rather than principled, and we do not fully explain why compressed relay helps on some benchmarks but not others. The denoising interpretation and OBF's gain over plain eviction are consistent with our results but unverified at the mechanism level. OBF's residual SVD adds per-relay compute beyond the base selector; a fast Gram-based alternative is characterized in Appendix~\ref{app:fast_svd}. Finally, our evaluation uses reasoning, QA, and code benchmarks; performance on real-world multi-agent applications such as long-form collaboration, tool use, or full agentic workflows remains untested.

\clearpage
\bibliographystyle{plain}
\bibliography{example_paper}

\clearpage
\appendix
\section{Experiment Settings}
\subsection{Sequential LatentMAS Setting}
\label{app:latentmas_setting}

We follow the sequential LatentMAS (LMAS) setting of Zou et al.~\cite{zou2025latentmas} without modification. Four agents are arranged in a pipeline: Planner, Critic, Refiner, and Judger. Each non-Judger agent performs latent reasoning under its role-specific prompt and relays its full KV cache to the next agent as \texttt{past\_key\_values}; the downstream agent conditions on its text input (role prompt and question) and the transferred KV cache. Only the Judger decodes a textual response.

\subsection{Prompt Templates}
\label{app:prompt}

We reuse the LMAS prompt templates verbatim. Let $\mathcal{Q}$ denote the input question shared across agents.

\paragraph{System Prompt.}
All agents share the same system prompt:
\begin{quote}
\texttt{You are Qwen, created by Alibaba Cloud. You are a helpful assistant.}
\end{quote}

\paragraph{Planner.}
\begin{quote}
\texttt{You are a Planner Agent. Given an input question, design a clear, step-by-step plan for how to solve the question.}

\texttt{Question: $\mathcal{Q}$}

\texttt{Your outlined plan should be concise with a few bulletpoints for each step. Do not produce the final answer.}\\
\texttt{Now output your plan to solve the question below:}
\end{quote}

\paragraph{Critic.}
\begin{quote}
\texttt{Question: $\mathcal{Q}$}

\texttt{You are a Critic Agent to evaluate the correctness of the input plan for the given question and provide helpful feedback for improving the plan.}\\
\texttt{The plan information is provided in latent KV representation format. Review the plan and question and output:}\\
\texttt{(1) original plan contents}\\
\texttt{(2) constructive feedback on the original plan.}

\texttt{Format your response as follows:}\\
\texttt{Original Plan: [Copy the provided Planner Agent's plan here]}\\
\texttt{Feedback: [Your detailed feedback to improve the plan here]}

\texttt{Now, output your response below:}
\end{quote}

\paragraph{Refiner.}
\begin{quote}
\texttt{Question: $\mathcal{Q}$}

\texttt{You are a Refiner Agent to provide a refined step-by-step plan for solving the given question.}\\
\texttt{You are provided with:}\\
\texttt{(1) latent-format information: a previous plan with feedback}\\
\texttt{(2) text-format information: the input question you need to solve.}

\texttt{Based on the input, write a refined and improved plan to solve the question. Make sure your output plan is correct and concise.}

\texttt{Now, output your refined plan below:}
\end{quote}

\paragraph{Judger.}
For all tasks, the Judger is told that latent information is available as reference and may contain irrelevant content. The task-specific output constraint differs by benchmark family.

\textbf{(1) Math / numeric-answer tasks} (\texttt{gsm8k}, \texttt{aime2024}, \texttt{aime2025}):
\begin{quote}
\texttt{Target Question: $\mathcal{Q}$}

\texttt{You are a helpful assistant. You are provided with latent information for reference and a target question to solve.}

\texttt{The latent information might contain irrelevant contents. Ignore it if it is not helpful for solving the target question.}

\texttt{You must reason step-by-step to solve the provided Target Question without outputting other irrelevant information.}

\texttt{Now, reason step by step and output the final answer inside \textbackslash boxed\{YOUR\_FINAL\_ANSWER\}.}
\end{quote}

\textbf{(2) Multiple-choice tasks} (\texttt{arc\_easy}, \texttt{arc\_challenge}, \texttt{gpqa}, \texttt{medqa}):
\begin{quote}
\texttt{Target Question: $\mathcal{Q}$}

\texttt{You are a helpful assistant. You are provided with latent information for reference and a target question to solve.}

\texttt{The latent information might contain irrelevant contents. Ignore it if it is not helpful for solving the target question.}

\texttt{You must reason step-by-step to solve the provided Target Question without outputting other irrelevant information.}\\
\texttt{Your final answer must be selected from A,B,C,D. For example \textbackslash boxed\{A\}. Do not add any other contents inside the box.}

\texttt{Now, reason step by step and output the final answer inside \textbackslash boxed\{YOUR\_FINAL\_ANSWER\}.}
\end{quote}

\textbf{(3) Code-generation tasks} (\texttt{mbppplus}, \texttt{humanevalplus}):
\begin{quote}
\texttt{Target Question: $\mathcal{Q}$}

\texttt{You are a helpful assistant. You are provided with latent information for reference and a target question to solve.}

\texttt{The latent information might contain irrelevant contents. Ignore it if it is not helpful for solving the target question.}

\texttt{You must reason step-by-step to solve the provided Target Question without outputting other irrelevant information.}\\
\texttt{You must put all python code as self-contained Python function in markdown code blocks. Do not add any other contents inside the markdown code block.}

\texttt{Now, reason step by step and output the final answer inside a Python markdown code block.}
\end{quote}

\subsection{Hardware Environment}
\label{app:hardware}

Unless otherwise specified, all experiments run on NVIDIA A100-PCIE-40GB nodes.

\subsection{Maximum Generated Tokens and Batch Sizes}
\label{app:max_token_and_batch_size}

Table~\ref{tab:max_token_and_batch_size} lists the maximum generated tokens and batch sizes used throughout the paper, following the LMAS defaults.

\begin{table}[H]
\centering
\small
\begin{tabular}{lcc}
\toprule
Datasets & Max generated tokens & Batch size \\
\midrule
GSM8K, ARC-E, ARC-C & 2048  & 4 \\
MBPP+, HumanEval+, MedQA & 4096 & 2 \\
GPQA & 8192 & 2 \\
AIME2024, AIME2025 & 20000 & 2 \\
\bottomrule
\end{tabular}
\caption{\textbf{Task-specific token generation budgets and batch sizes.}}
\label{tab:max_token_and_batch_size}
\end{table}

\subsection{Shared Experimental Configuration}
\label{app:ablation_setup}
\label{app:hyper}

All experiments in the paper share the sequential LMAS setting (Appendix~\ref{app:latentmas_setting}), prompt templates (Appendix~\ref{app:prompt}), hardware (Appendix~\ref{app:hardware}), and per-task decoding budgets (Appendix~\ref{app:max_token_and_batch_size}). We use the Qwen3 family~\cite{qwen3} as the backbone throughout; the specific model is stated per-table. Decoding uses \texttt{bfloat16} precision with stochastic sampling (temperature $0.6$, top-$p$ $0.95$).

\paragraph{Fixed hyperparameters.}
Every experiment uses $\texttt{kv\_budget}=32$ prompt tokens retained per non-judger agent per round, $\texttt{sink\_size}=4$ tokens retained once across the rollout, and $\texttt{latent\_steps}=40$ latent iterations per round. OBF variants use fixed ranks $\texttt{pca\_rank}=4$ for L-OBF and $\texttt{pca\_rank}=2$ for H-OBF, chosen on Qwen3-4B before any Qwen3-8B or Qwen3-14B run.

\paragraph{Seeding.}
All main and appendix results are three-seed means unless the row explicitly says otherwise. Seed sets are chosen per backbone:
\begin{itemize}[leftmargin=1.2em]
    \item \textbf{Qwen3-4B}: seeds $\{4, 44, 444\}$.
    \item \textbf{Qwen3-8B}: seeds $\{8, 88, 888\}$.
    \item \textbf{Qwen3-14B}: seeds $\{14, 1414, 141414\}$.
\end{itemize}

\paragraph{AIME reporting convention.}
AIME24 and AIME25 each contain $30$ problems, and per-seed accuracy swings above $20$~pp are common. We quote their numbers alongside the other benchmarks but treat their margins as inconclusive, and cross-benchmark averages are computed on the seven non-AIME benchmarks (or the five used in the compression and ablation sweeps).

\subsection{Algorithm}
\label{app:algorithm}

\begingroup
\setlength{\floatsep}{4pt plus 1pt minus 1pt}
\setlength{\textfloatsep}{4pt plus 1pt minus 1pt}
\setlength{\intextsep}{4pt plus 1pt minus 1pt}

We organize the algorithm into one unified relay procedure (Algorithm~\ref{alg:relay}) that treats the compression step as a black box, and two sub-algorithms (Algorithms~\ref{alg:attn_selector} and \ref{alg:obf}) that instantiate the compression operator $\Phi$. Setting $\Phi=\Phi_{\mathrm{Gen}}$ is trivial ($\Phi_{\mathrm{Gen}}(\mathcal{P}_i)=\emptyset$) and omitted.

\begin{algorithm}[htbp]
\caption{Unified relay at agent $\pi_i$.}
\label{alg:relay}
\begin{algorithmic}[1]
\Require Prompt $P_i$; previous message $\mathcal{M}_{i-1}$; latent-step count $p$; compression operator $\Phi$.
\Ensure Message $\mathcal{M}_i$.
\State $(\mathbf{K},\,\mathbf{V},\,\mathcal{G}_i,\,a) \gets \textsc{LatentAutoregression}\bigl(\mathrm{past\_kv}{=}\mathcal{M}_{i-1},\,\mathrm{prompt}{=}P_i,\,\mathrm{steps}{=}p\bigr)$
\State $\mathcal{P}_i^{\star} \gets \Phi(\mathbf{K},\mathbf{V},a)$ \Comment{see Alg.~\ref{alg:attn_selector} / Alg.~\ref{alg:obf}}
\State $\mathcal{M}_i \gets [\mathcal{M}_{i-1}\,;\ \mathcal{P}_i^{\star}\,;\ \mathcal{G}_i]$
\State \Return $\mathcal{M}_i$
\end{algorithmic}
\end{algorithm}

The following two sub-algorithms use the notation of Section~\ref{sec:method}: $\mathcal{P}_{\mathrm{keep}}/\mathcal{P}_{\mathrm{del}}$ are retained / deleted prompt index sets, and $\mathbf{V}_{\mathrm{keep}}/\mathbf{V}_{\mathrm{del}}$ the corresponding value slices.

\begin{algorithm}[htbp]
\caption{$\Phi_{\mathrm{Attn}}$: attention-based selector (Section~\ref{subsec:h2o_selection}).}
\label{alg:attn_selector}
\begin{algorithmic}[1]
\Require $\mathbf{K},\mathbf{V}$; attention tensor $a$; budget $B$; granularity $\in\{\mathrm{H},\mathrm{L}\}$.
\State $A_j^{(l,h)} \gets \sum_{t=1}^{p} a_{t,j}^{(l,h)}$ for $j\in\mathcal{P}_i$
\If{granularity $=\mathrm{H}$}
    \State $\mathcal{P}_{\mathrm{keep}}^{(l,h)} \gets \operatorname{TopK}_B\bigl(\{A_j^{(l,h)}\}_{j\in\mathcal{P}_i}\bigr)$
\ElsIf{granularity $=\mathrm{L}$}
    \State $\mathcal{P}_{\mathrm{keep}}^{(l,h)} \gets \operatorname{TopK}_B\bigl(\{\textstyle\sum_h A_j^{(l,h)}\}_{j\in\mathcal{P}_i}\bigr)$
\EndIf
\State $\mathbf{K}_{\mathrm{keep}},\mathbf{V}_{\mathrm{keep}} \gets \mathbf{K}[\mathcal{P}_{\mathrm{keep}}^{(l,h)},:],\,\mathbf{V}[\mathcal{P}_{\mathrm{keep}}^{(l,h)},:]$
\State \Return $(\mathbf{K}_{\mathrm{keep}},\mathbf{V}_{\mathrm{keep}},\mathcal{P}_{\mathrm{keep}}^{(l,h)})$
\end{algorithmic}
\end{algorithm}

\begin{algorithm}[htbp]
\caption{$\Phi_{\mathrm{OBF}}$: Orthogonal BackFill (Section~\ref{subsec:obf}).}
\label{alg:obf}
\begin{algorithmic}[1]
\Require $\mathbf{K},\mathbf{V}$; attention tensor $a$; budget $B$; granularity $\in\{\mathrm{H},\mathrm{L}\}$; PCA rank $k$.
\State $(\mathbf{K}_{\mathrm{keep}},\mathbf{V}_{\mathrm{keep}},\mathcal{P}_{\mathrm{keep}}^{(l,h)}) \gets \Phi_{\mathrm{Attn}}(\mathbf{K},\mathbf{V},a,B,\text{granularity})$ \Comment{Alg.~\ref{alg:attn_selector}}
\For{each $(l,h)$}
    \State $\mathcal{P}_{\mathrm{del}}^{(l,h)} \gets \mathcal{P}_i \setminus \mathcal{P}_{\mathrm{keep}}^{(l,h)}$;\quad $\mathbf{V}_{\mathrm{del}} \gets \mathbf{V}[\mathcal{P}_{\mathrm{del}}^{(l,h)},:]$
    \If{$\mathcal{P}_{\mathrm{del}}^{(l,h)}=\emptyset$} \textbf{continue} \EndIf
    \State $\tilde{A}_{\mathrm{keep}} \gets \sum_{j\in\mathcal{P}_{\mathrm{keep}}^{(l,h)}} A_j^{(l,h)}$;\quad $\tilde{A}_{\mathrm{del}} \gets \sum_{j\in\mathcal{P}_{\mathrm{del}}^{(l,h)}} A_j^{(l,h)}$
    \State $\mathbf{Q} \gets \textsc{qr}(\mathbf{V}_{\mathrm{keep}}^\top)$;\quad $\mathbf{R} \gets \mathbf{V}_{\mathrm{del}} - \mathbf{V}_{\mathrm{del}}\mathbf{Q}\mathbf{Q}^\top$
    \State $\mathbf{C} \gets \textsc{svd}_k(\mathbf{R})$
    \State $\tilde{w}_t \gets A_t^{(l,h)}/(\tilde{A}_{\mathrm{del}}+\epsilon)$ for $t\in\mathcal{P}_{\mathrm{del}}^{(l,h)}$;\quad $\bar{\mathbf{r}}\gets\sum_t \tilde{w}_t\,\mathbf{R}_t$
    \State $\delta \gets (\bar{\mathbf{r}}\mathbf{C}^\top)\mathbf{C}$;\quad $\alpha\gets\tilde{A}_{\mathrm{del}}/(\tilde{A}_{\mathrm{keep}}+\epsilon)$
    \State $\mathbf{V}_{\mathrm{keep}}\gets\mathbf{V}_{\mathrm{keep}}+\alpha\,\mathbf{1}\delta$ \Comment{$\mathbf{1}\!\in\!\mathbb{R}^{B\times 1}$, broadcast $\alpha\delta$ to every retained row}
\EndFor
\State \Return $(\mathbf{K}_{\mathrm{keep}},\mathbf{V}_{\mathrm{keep}})$
\end{algorithmic}
\end{algorithm}

\endgroup

\subsection{Implementation Details}
\label{app:implementation}

Our implementation builds on the official LatentMAS codebase~\cite{zou2025latentmas}. Supporting compressed relay under batched HuggingFace inference requires four adjustments to the inference path.

\paragraph{Right padding.}
We use right padding. Our compressors protect sink tokens as the first $s$ indices of the KV cache (following StreamingLLM~\cite{xiao2024streamingllm}), so position $0$ must be a real token for every sample; left padding would place a pad KV in that slot for short samples. Because the last position of a padded row is now a pad token, we gather the hand-off hidden state at each sample's last valid index using the attention mask, not \texttt{[:,\,-1,\,:]}.

\paragraph{Cross-agent attention mask.}
At each hand-off, the inherited KV cache $\mathcal{M}_{i-1}$ is concatenated with the freshly encoded prompt KV $\mathcal{P}_i$, and an attention mask covering both ranges is fed to the model. We propagate the real attention mask across hand-offs so that no downstream token can attend to a pad-origin KV slot, which matters at $N_b{>}1$ with mixed prompt lengths. When a compressor changes the cache length, the inherited mask no longer aligns with the new cache, and we fall back to an all-ones past mask for that hand-off. This is safe because the retained slots are the highest-attention ones and are dominated by real-token entries.

\paragraph{Per-sample position tracking.}
Under right padding and compression, samples in a batch accumulate different numbers of real KV entries across agents, latent steps, and (per layer and head) evictions. HuggingFace's single \texttt{cache\_position} shared across the batch is only correct when all samples are at the same next position. We therefore maintain a per-sample position cursor $\mathbf{c}\in\mathbb{Z}^{N_b}$ and pass explicit \texttt{position\_ids} of shape $(N_b, L)$ at every forward pass. For the prompt of each agent, real tokens receive consecutive offsets $\mathbf{c},\,\mathbf{c}{+}1,\,\dots$ using the cumulative sum of the attention mask, while pad tokens continue the same schedule so no two KV entries share a position id. For the $t$-th latent step and for text generation, sample $b$ uses position $\mathbf{c}_b+t$. After each stage, $\mathbf{c}$ advances by the number of valid slots actually appended to each sample's cache, using the compressor-reported retained lengths where applicable.

\paragraph{Manual decoding rollout.}
\texttt{model.generate()} exposes no API for injecting per-sample \texttt{position\_ids} at every decoding step, so both the latent-step loop and text generation are manual decoding loops. At each step we set $\text{position\_ids}=\mathbf{c}+t$ per sample, extend the attention mask by one, run a single-token forward with explicit \texttt{past\_key\_values}, and select the next token via greedy decoding. Per-sample EOS and max-new-tokens conditions are tracked independently, so finished sequences stop advancing $\mathbf{c}$ while others continue.

\clearpage
\section{Scaling Behavior at Larger Backbones}
\label{app:scaling}

All runs follow Appendix~\ref{app:ablation_setup}, except we change the backbone to Qwen3-8B and Qwen3-14B.

\subsection{Qwen3-8B}
\label{app:qwen8b}

\begin{table*}[!htbp]
\centering
\footnotesize
\setlength{\tabcolsep}{4pt}
\renewcommand{\arraystretch}{1.08}
\resizebox{\textwidth}{!}{%
\begin{tabular}{l|ccc|cc|cc|cc}
\toprule
Method & \multicolumn{3}{c|}{Math} & \multicolumn{2}{c|}{Expert QA} & \multicolumn{2}{c|}{Commonsense QA} & \multicolumn{2}{c}{Coding} \\
\cmidrule(lr){2-4} \cmidrule(lr){5-6} \cmidrule(lr){7-8} \cmidrule(lr){9-10}
 & GSM8K & AIME24 & AIME25 & GPQA & MedQA & ARC-E & ARC-C & MBPP+ & HumanEval+ \\
\midrule
Full & 91.96 & 66.67 & 47.78 & \textbf{52.12} & 74.11 & \textbf{98.39} & \textbf{95.02} & \textbf{75.57} & \textbf{83.33} \\
L & 92.01 & 65.56 & \textbf{58.89} & 50.25 & \underline{75.33} & 98.26 & 94.48 & 73.19 & 79.47 \\
L-OBF & \textbf{92.70} & \underline{72.22} & \textbf{58.89} & 50.76 & \textbf{75.56} & 98.32 & 94.43 & 73.72 & 80.08 \\
H & 92.01 & \underline{72.22} & 53.33 & 50.76 & 74.11 & \underline{98.34} & \underline{94.82} & 73.99 & \textbf{83.33} \\
H-OBF & \underline{92.12} & 65.56 & \underline{55.56} & \underline{51.61} & 74.78 & 98.33 & 94.62 & \underline{74.34} & \underline{81.10} \\
Gen & 89.82 & \textbf{73.33} & \textbf{58.89} & 46.87 & 74.67 & 97.77 & 93.37 & 72.66 & 79.67 \\
\bottomrule
\end{tabular}%
}
\caption{\textbf{Qwen3-8B accuracy across benchmarks} (\%, three-seed mean on A100; setup in Appendix~\ref{app:ablation_setup}). Best per benchmark is in \textbf{bold}, runner-up \underline{underlined}.}
\label{tab:qwen8b_accuracy}
\end{table*}

\paragraph{Reading.}
The pattern inverts from 4B. At 8B the headwise selector is nearly lossless, so H-OBF has little to recover and is close to neutral; the layerwise selector still loses ground, and L-OBF matches or improves on it on eight of nine benchmarks. Scaling did not weaken OBF; it removed the damage OBF exists to undo, on one selector but not the other. Appendix~\ref{app:recovery_law} makes this quantitative.

\subsection{Qwen3-14B}
\label{app:qwen14b}

\begin{table*}[!htbp]
\centering
\footnotesize
\setlength{\tabcolsep}{5pt}
\renewcommand{\arraystretch}{1.08}
\resizebox{\textwidth}{!}{%
\begin{tabular}{l|c|cc|cc|cc}
\toprule
Method & Math & \multicolumn{2}{c|}{Expert QA} & \multicolumn{2}{c|}{Commonsense QA} & \multicolumn{2}{c}{Coding} \\
\cmidrule(lr){2-2} \cmidrule(lr){3-4} \cmidrule(lr){5-6} \cmidrule(lr){7-8}
 & GSM8K & GPQA & MedQA & ARC-E & ARC-C & MBPP+ & HumanEval+ \\
\midrule
Full & 91.89 & \textbf{59.90} & 80.67 & \textbf{98.67} & \textbf{95.79} & \underline{77.43} & 86.99 \\
L & \textbf{92.82} & \underline{56.68} & \textbf{81.11} & 98.54 & 95.59 & 77.60 & 85.57 \\
L-OBF & \underline{92.60} & 56.51 & 80.78 & 98.51 & 95.53 & 77.16 & 84.55 \\
H & 92.34 & 56.85 & 79.22 & 98.53 & 95.51 & \textbf{78.13} & 84.76 \\
H-OBF & 92.67 & \underline{57.36} & 79.89 & \underline{98.63} & \underline{95.62} & 76.90 & \textbf{88.01} \\
Gen & 91.51 & 52.96 & \underline{80.78} & 98.12 & 94.14 & 75.57 & 83.74 \\
\bottomrule
\end{tabular}%
}
\caption{\textbf{Qwen3-14B accuracy across benchmarks} (\%, three-seed mean on H200; setup in Appendix~\ref{app:ablation_setup}). AIME24 and AIME25 are omitted here, since their $30$-problem sets give per-seed swings too large to read at this scale (Appendix~\ref{app:ablation_setup}). Best per benchmark is in \textbf{bold}, runner-up \underline{underlined}.}
\label{tab:qwen14b_accuracy}
\end{table*}

\paragraph{Reading.}
The same pattern holds at 14B. Both selectors are nearly lossless here, so OBF stays near neutral on most benchmarks. The exception is HumanEval+, where headwise still gives up ground and H-OBF recovers more than the gap to finish above Full. Across the three scales, the room OBF has to work in shrinks exactly as eviction becomes less damaging.

\clearpage
\section{The Effectiveness of OBF}
\label{app:recovery_law}

\begin{figure}[!htbp]
    \centering
    \includegraphics[width=\textwidth]{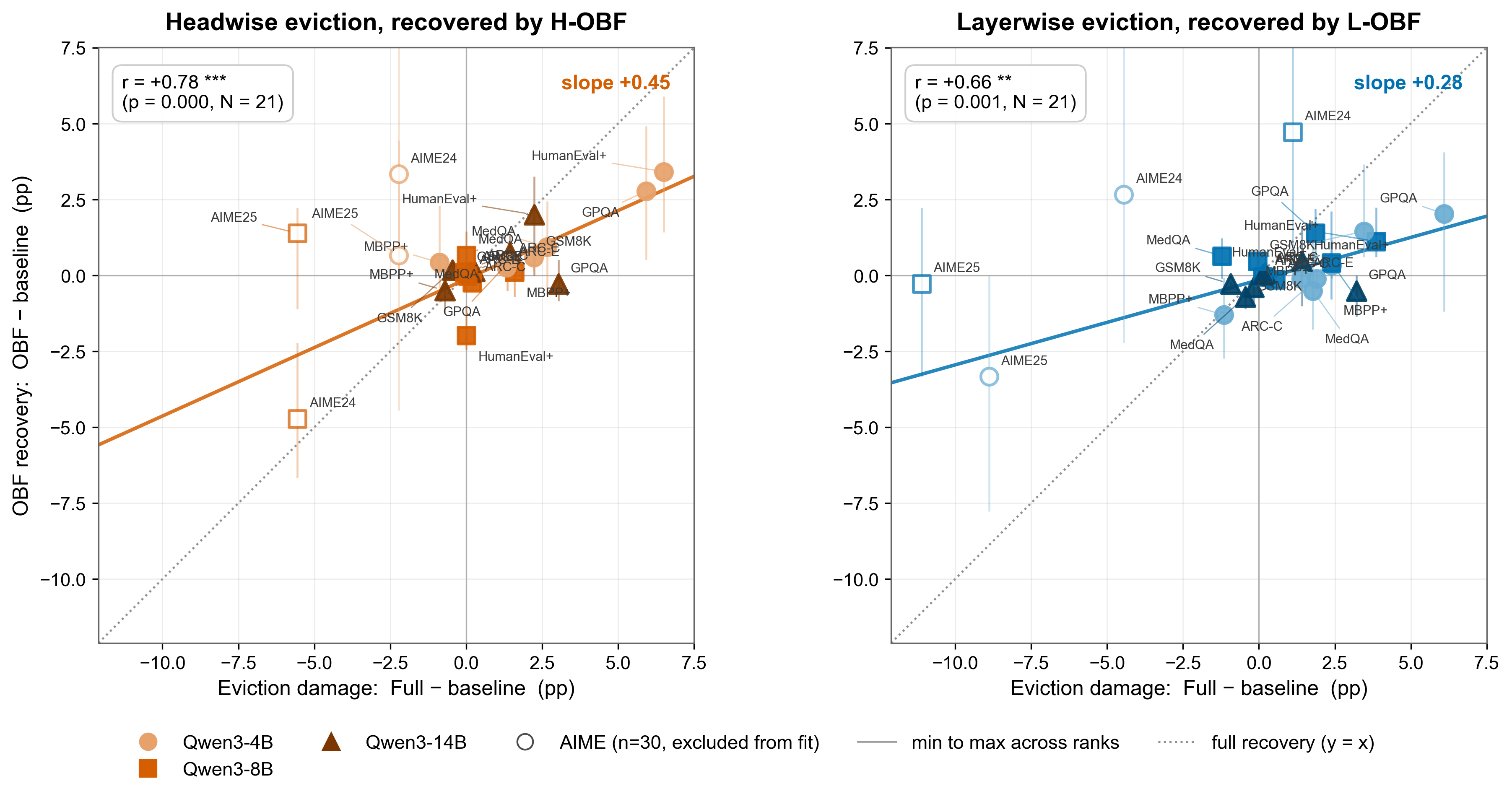}
    \caption{\textbf{OBF recovers eviction damage in proportion to its magnitude.} Each point is a (model, benchmark) pair; the $x$-axis is the eviction gap against Full and the $y$-axis is OBF's recovery of that gap (both in percentage points). Vertical bars span min-to-max across the ranks swept at that scale; markers plot the rank-averaged value. Marker shape and shade identify the backbone, shading darker with model size. Solid lines are OLS fits on the non-AIME cells at all three scales ($N{=}21$ per selector). AIME24/25 (30 problems each) are shown as hollow markers and excluded from the fit. The dashed diagonal $y{=}x$ marks full recovery of eviction damage.}
    \label{fig:recovery_law_app}
\end{figure}

We test whether OBF restores accuracy in proportion to what eviction discards. For each (model, benchmark) pair and each selector, define
\[
\text{damage} \;=\; \text{Acc}_{\text{Full}} - \text{Acc}_{\text{baseline}}, \qquad
\text{recovery} \;=\; \frac{1}{|R|}\sum_{r \in R} \text{Acc}_{\text{OBF}}(r) - \text{Acc}_{\text{baseline}},
\]
where $R$ is the \texttt{pca\_rank} set swept at that scale ($\{2,4,8,16,32\}$ at 4B, $\{2,4,8,32\}$ at 8B, $\{2,4,8\}$ at 14B), baseline $\in$ \{H, L\}, and OBF $\in$ \{H-OBF, L-OBF\}. Recovery averages over the whole sweep so no rank is selected anywhere. Data are from Qwen3-4B (Table~\ref{tab:qwen4b_accuracy}), Qwen3-8B (Table~\ref{tab:qwen8b_accuracy}) and Qwen3-14B (Table~\ref{tab:qwen14b_accuracy}); AIME24 and AIME25 (hollow markers in Figure~\ref{fig:recovery_law_app}) are excluded from the fit following Appendix~\ref{app:ablation_setup}, leaving $N=21$ non-AIME cells per selector.

OLS regression of recovery on damage yields:

\begin{center}
\begin{tabular}{lcccc}
\toprule
Selector & Slope & Intercept & Pearson $r$ & $p$-value \\
\midrule
Headwise (H-OBF vs.\ H) & $+0.45$ & $\approx 0$ & $+0.78$ & $<10^{-4}$ \\
Layerwise (L-OBF vs.\ L) & $+0.28$ & $\approx 0$ & $+0.66$ & $0.001$ \\
\bottomrule
\end{tabular}
\end{center}

Both slopes are positive and significant. Both intercepts are statistically indistinguishable from zero, so where eviction does not damage accuracy, OBF neither helps nor hurts. Headwise OBF recovers roughly half of what headwise eviction removes; layerwise OBF recovers roughly a third.

\subsection{An independent test on the relay budget}
\label{app:recovery_law_budget}

The fit above varies the eviction gap by changing the benchmark. The relay budget varies it directly, since a smaller \texttt{kv\_budget} discards more of the prompt and widens the gap on a fixed benchmark. Appendix~\ref{app:budget_sweep} sweeps $B \in \{4, 8, 16, 32, 64\}$ on four benchmarks at three seeds, so it tests the relationship on runs the fit above never saw. We refit at each budget separately across the four benchmarks.

\begin{center}
\begin{tabular}{lccccc}
\toprule
\texttt{kv\_budget} & $4$ & $8$ & $16$ & $32$ & $64$ \\
\midrule
Pearson $r$ & $-0.07$ & $+0.21$ & $\mathbf{+0.92}$ & $\mathbf{+0.91}$ & $-0.94$ \\
Slope & $-0.16$ & $+0.32$ & $+1.15$ & $+0.84$ & $-1.57$ \\
\bottomrule
\end{tabular}
\end{center}

The relationship reproduces at $B{=}16$ and $B{=}32$, and pooling the three larger budgets (twelve cells) gives $r = +0.79$, slope $+0.80$. The two smallest budgets fall outside it. At $B \in \{4, 8\}$ the retained set is too small for the correction to be absorbed, and H-OBF is below plain eviction on every benchmark (Appendix~\ref{app:budget_sweep}), so there is no recovery to explain. The per-budget fit at $B{=}64$ is unreliable on its own, because its four gaps span only $0.9$ to $2.3$~pp against $1.9$ to $6.5$~pp at $B{=}32$, and one benchmark sets the sign. We read the sweep as confirming the relationship wherever OBF is in its working range, not as extending it to every budget.

\clearpage
\section{Ablation Analysis of OBF Design Choices}
\label{app:obf_ablation}

We ablate OBF's design choices in two groups: \emph{component ablations} remove or simplify one internal step of full L-OBF, and \emph{alternative designs} replace one step with a different implementation of the same intent. All runs follow Appendix~\ref{app:ablation_setup}; AIME24 and AIME25 are excluded following the convention there.

\subsection{Component Ablations}

\begin{itemize}[leftmargin=1.2em]
    \item \textbf{No Projection}: skip the residual extraction in Section~\ref{sec:projection} that removes from $\mathbf{V}_{\mathrm{del}}$ the component representable in the retained-value span.
    \item \textbf{Max-P}: skip the low-rank principal subspace extraction in Section~\ref{sec:pca} and use all components of the residual.
    \item \textbf{No Scaling}: remove the demand-aware dynamic scaling in Section~\ref{sec:dynamic_scale}.
    \item \textbf{Naive Aggregation}: form the backfill vector by directly averaging the deleted value states, bypassing all three steps.
\end{itemize}

\begin{figure}[!htbp]
    \centering
    \includegraphics[width=\textwidth]{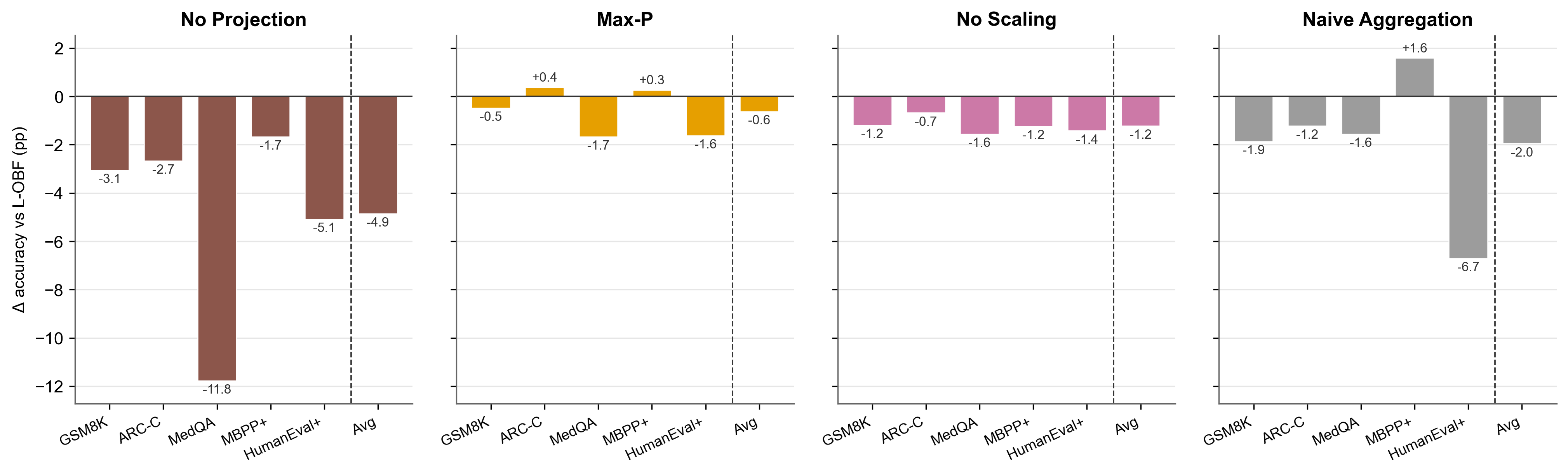}
    \caption{\textbf{Component ablations of full L-OBF.} Bars show $\Delta$ accuracy (pp) versus the L-OBF reference; the L-OBF reference itself is at zero and omitted. Negative bars mean the variant loses accuracy.}
    \label{fig:obf_ablation_components}
\end{figure}

\begin{table}[!htbp]
\centering
\setlength{\tabcolsep}{6pt}
\renewcommand{\arraystretch}{1.1}
{\footnotesize
\begin{tabular}{l|c}
\toprule
Variant & Avg $\Delta$ vs L-OBF (pp) \\
\midrule
No Projection      & $-4.86$ \\
Naive Aggregation  & $-1.96$ \\
No Scaling         & $-1.22$ \\
Max-P              & $-0.63$ \\
\bottomrule
\end{tabular}%
}
\setlength{\abovecaptionskip}{12pt}
\caption{\textbf{Component ablations: average $\Delta$ accuracy vs L-OBF} across the five benchmarks. Ordered by average degradation.}
\label{tab:obf_ablation_components}
\end{table}

\paragraph{Analysis.}
Removing projection causes the largest and most consistent drop across benchmarks, so we treat it as a key step of the design. Scaling shows the same pattern, milder. Naive Aggregation swings unpredictably across tasks, so unstructured averaging cannot substitute for projection plus scaling. Max-P is the mildest case by design, because OBF already allows any rank up to the swept maximum, and setting the rank to its cap is a valid configuration rather than a removed component.

\subsection{Alternative Designs}

\begin{itemize}[leftmargin=1.2em]
    \item \textbf{Attn-inject}: replace the uniform injection $\mathbf{V}_t^{\star} \leftarrow \mathbf{V}_t + \alpha\delta$ with an attention-weighted injection
    \[
        \mathbf{V}_t^{\star} \leftarrow \mathbf{V}_t + B\,\tilde{w}_t^{\mathrm{keep}}\,\alpha\delta,
        \qquad
        \tilde{w}_t^{\mathrm{keep}} = \frac{A_t^{(l,h)}}{\sum_{s\in\mathcal{P}_{\mathrm{keep}}} A_s^{(l,h)}+\epsilon},
    \]
    so each retained position receives a share of the backfill proportional to its attention mass from the downstream agent (rather than an equal share).

    \item \textbf{Per-token}: instead of a single aggregated direction $\delta$ shared across all retained slots, form a per-slot direction $\delta_t$ for each $t \in \mathcal{P}_{\mathrm{keep}}$ (analogous to Eq.~\ref{eq:delta} but with per-$t$ residual weighting) and inject
    \[
        \mathbf{V}_t^{\star} \leftarrow \mathbf{V}_t + \alpha\,\delta_t, \qquad t \in \mathcal{P}_{\mathrm{keep}}.
    \]
    This preserves finer-grained information but multiplies the extra state per relay by $|\mathcal{P}_{\mathrm{keep}}|$.

    \item \textbf{EVR-adaptive ($\tau{=}0.35$)}: replace the fixed rank $r=4$ with an adaptive rank
    \[
        k^{\star} \;=\; \min\Bigl\{k \;:\; \sum_{i=1}^{k}\sigma_i^{2}\,\Big/\,\sum_{i}\sigma_i^{2} \;\geq\; \tau\Bigr\},
    \]
    where $\sigma_i$ are the singular values of the residual matrix $\mathbf{R}$. On our data this criterion is well approximated by $\texttt{pca\_rank}=35$, which is what we sweep.
\end{itemize}

\begin{figure}[!htbp]
    \centering
    \includegraphics[width=\textwidth]{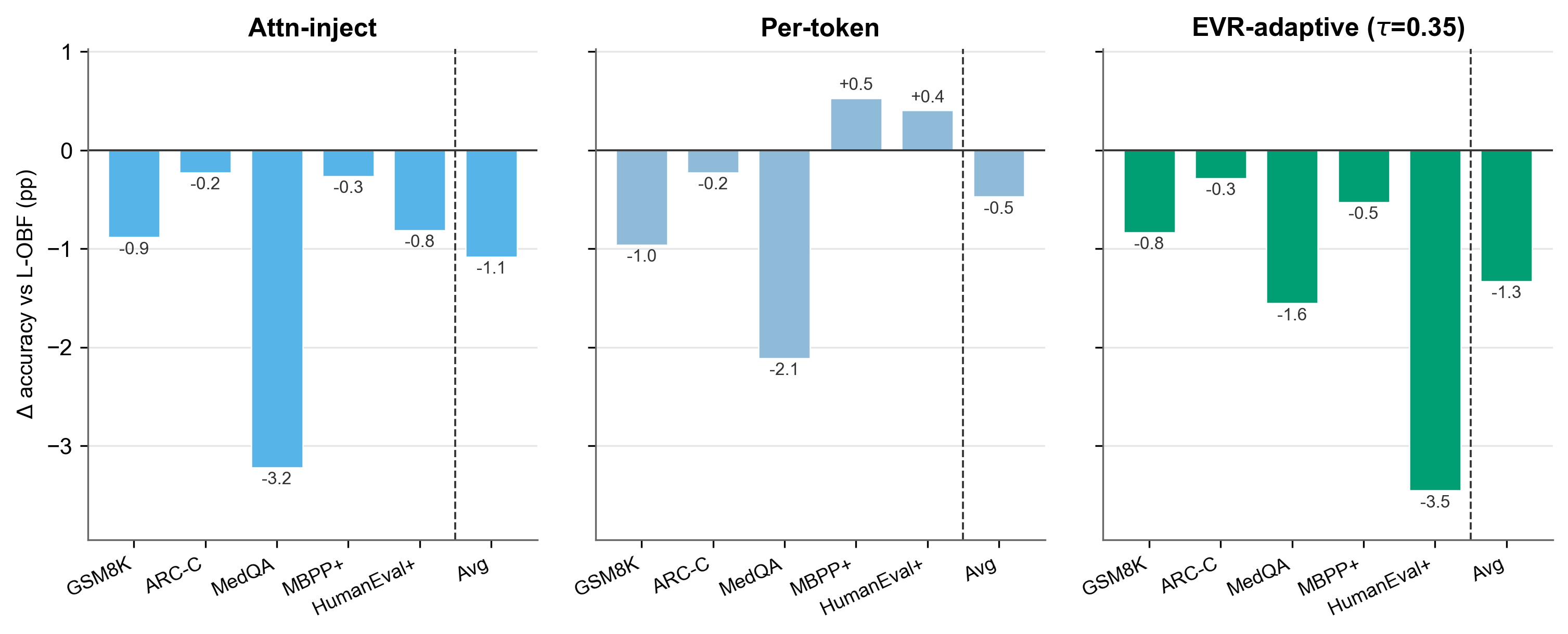}
    \caption{\textbf{Alternative-design ablations of full L-OBF.} Bars show $\Delta$ accuracy (pp) versus the L-OBF reference; the L-OBF reference itself is at zero and omitted. Negative bars mean the variant loses accuracy.}
    \label{fig:obf_ablation_alternatives}
\end{figure}

\begin{table}[!htbp]
\centering
\setlength{\tabcolsep}{6pt}
\renewcommand{\arraystretch}{1.1}
{\footnotesize
\begin{tabular}{l|c}
\toprule
Variant & Avg $\Delta$ vs L-OBF (pp) \\
\midrule
EVR-adaptive ($\tau{=}0.35$) & $-1.33$ \\
Attn-inject                  & $-1.09$ \\
Per-token                    & $-0.48$ \\
\bottomrule
\end{tabular}%
}
\setlength{\abovecaptionskip}{12pt}
\caption{\textbf{Alternative designs: average $\Delta$ accuracy vs L-OBF} across the five benchmarks. Ordered by average degradation.}
\label{tab:obf_ablation_alternatives}
\end{table}

\paragraph{Summary.}
No alternative improves over L-OBF at the same relay budget. Per-token residuals come closest ($-0.48$~pp on average) but require storing a separate residual per retained token. Attn-inject re-emphasizes content the retained states already carry, and EVR-adaptive rank pulls in noisier residual directions that low-rank truncation was designed to drop.

\clearpage
\section{Comparison with External KV Compression Baselines}
\label{app:external_baselines}

We compare L-OBF against two other KV compression families on Qwen3-4B: a cache-merging baseline in the spirit of CaM~\cite{zhang2024cam}, which merges to-be-evicted KV entries into retained ones instead of dropping them, and post-hoc int8/int4 quantization in the spirit of KVQuant~\cite{hooper2025kvquant}, applied either to the full KV cache or on top of L-OBF. Neither is the exact code from the cited work; we describe the concrete cache-merging implementation below. All runs follow Appendix~\ref{app:ablation_setup}. The sweep covers five benchmarks, one per capability in the suite (GSM8K reasoning, MedQA domain knowledge, ARC-C commonsense, MBPP+ and HumanEval+ code); ARC-E is dropped as a near duplicate of ARC-C, and GPQA groups with the AIME sets as long reasoning and is deferred per the AIME convention.

\paragraph{Cache-merging implementation.}
At every relay boundary and every $(l,h)$, we do three things:

\textbf{(i) Select.} The same attention selector as the eviction baselines picks the retained set of size $B$:
\[
\mathcal{P}_{\mathrm{keep}} \;=\; \mathrm{TopK}_B\bigl(\{A_t^{(l,h)}\}_{t\in\mathcal{P}}\bigr), \qquad \mathcal{P}_{\mathrm{del}} \;=\; \mathcal{P}\setminus\mathcal{P}_{\mathrm{keep}}.
\]

\textbf{(ii) Assign.} Each deleted position is routed to its nearest retained slot by key cosine similarity:
\[
\pi(j) \;=\; \arg\max_{t\in\mathcal{P}_{\mathrm{keep}}}\frac{\langle\mathbf{K}_j,\mathbf{K}_t\rangle}{\lVert\mathbf{K}_j\rVert\,\lVert\mathbf{K}_t\rVert},
\qquad j\in\mathcal{P}_{\mathrm{del}}.
\]

\textbf{(iii) Merge.} The routed value states are added into the retained slots, weighted by attention mass:
\[
\mathbf{V}_t^{\star} \;=\; \mathbf{V}_t \;+\; \sum_{j\,:\,\pi(j)=t}\frac{A_j^{(l,h)}}{A_t^{(l,h)}+\epsilon}\,\mathbf{V}_j,
\qquad t\in\mathcal{P}_{\mathrm{keep}}.
\]
Keys are unchanged. The retained-set size stays at $B$, so the row shares the $62.4$~MB budget of the other selection-based baselines. This differs from CaM in the routing rule and the weighting but implements the same principle: recycle discarded KV content into retained slots rather than dropping it.

\begin{table*}[!htbp]
\centering
\setlength{\tabcolsep}{5pt}
\renewcommand{\arraystretch}{1.1}
{\footnotesize
\begin{tabular}{l|c|ccccc|cc}
\toprule
Method & Compression & MedQA & GSM8K & ARC-C & MBPP+ & HEval+ & Mean & Worst \\
\midrule
Full (uncompressed) & $1.0\times$ & -- & -- & -- & -- & -- & -- & -- \\
\midrule
Layerwise       & $4.6\times$ & $-1.78$ & $-2.40$ & $-1.91$ & $+1.15$ & $-3.46$ & $-1.68$ & $-3.46$ \\
L-OBF ($r{=}4$) & $4.6\times$ & $-0.11$ & $-1.24$ & $-1.73$ & $+1.06$ & $\mathbf{+0.20}$ & $\mathbf{-0.36}$ & $\mathbf{-1.73}$ \\
Cache Merging   & $4.6\times$ & $-2.89$ & $-1.72$ & $-2.10$ & $\mathbf{+3.35}$ & $-1.42$ & $-0.96$ & $-2.89$ \\
\midrule
Full + int8 & $1.9\times$ & $\mathbf{+0.22}$ & $-0.99$ & $\mathbf{+0.80}$ & $-1.85$ & $-2.44$ & $-0.85$ & $-2.44$ \\
Full + int4 & $3.7\times$ & $-0.33$ & $\mathbf{-0.48}$ & $+0.09$ & $+0.18$ & $-6.71$ & $-1.45$ & $-6.71$ \\
\midrule
L-OBF + int8 & $\mathbf{9.0\times}$ & $-2.89$ & $-1.74$ & $-1.54$ & $+2.38$ & $-3.05$ & $-1.37$ & $-3.05$ \\
\bottomrule
\end{tabular}%
}
\setlength{\abovecaptionskip}{12pt}
\caption{\textbf{External KV compression baselines on Qwen3-4B.} Each cell is the accuracy delta (pp) versus uncompressed Full relay. \emph{Compression} is the relay-size reduction versus Full. \emph{Mean} is the cross-benchmark average delta. \emph{Worst} is the largest single-benchmark drop. Best per column is in \textbf{bold}.}
\label{tab:external_baselines}
\end{table*}

\paragraph{Analysis.}
At the $4.6\times$ relay budget, L-OBF has the best Mean, the mildest Worst, and the tightest spread across benchmarks. Quantization and cache merging each score highest on one task but drop sharply on another, most visibly on HumanEval+. Stacking int8 on L-OBF pushes the ratio to $9.0\times$ at a modest average loss, higher than any single-family method here reaches.

\clearpage
\section{Effect of \texttt{kv\_budget}}
\label{app:budget_sweep}

\begin{figure}[!htbp]
    \centering
    \includegraphics[width=\textwidth]{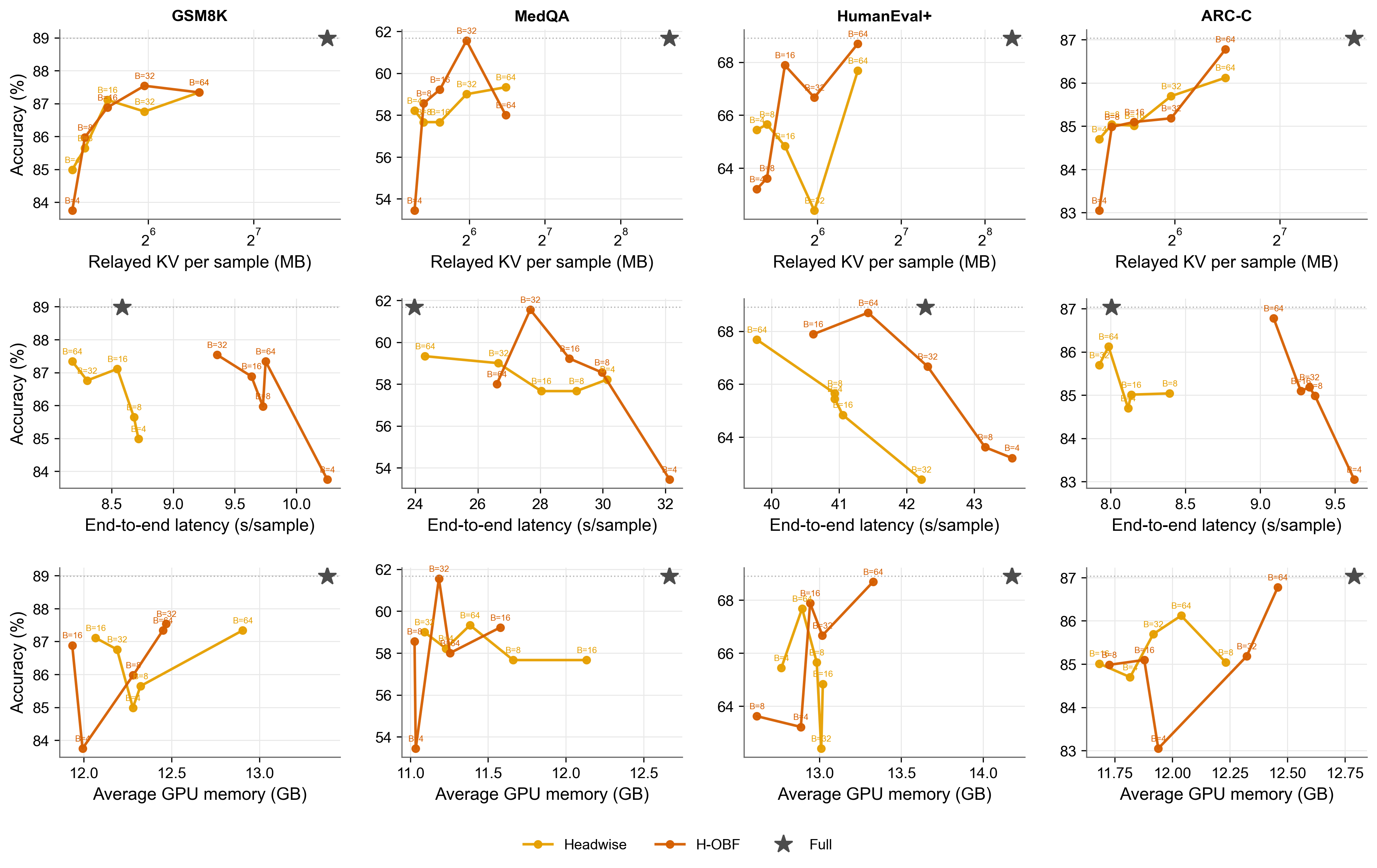}
    \caption{\textbf{Accuracy vs cost trade-offs.} For each benchmark, accuracy is plotted against three cost axes: relayed KV per sample (top row, log scale), end-to-end latency (middle row), and average GPU memory (bottom row). Each marker labelled $B{=}k$ is one budget in $\{4,8,16,32,64\}$; the star is Full relay.}
    \label{fig:budget_tradeoff}
\end{figure}

We sweep \texttt{kv\_budget} $B\in\{4,8,16,32,64\}$ for headwise eviction (H) and H-OBF at the paper's fixed rank $r{=}2$, on four benchmarks (GSM8K, ARC-C, MedQA, HumanEval+). All runs follow Appendix~\ref{app:ablation_setup}. H-OBF here uses the fast SVD path of Appendix~\ref{app:fast_svd}, so the latency axis in Figure~\ref{fig:budget_tradeoff} already reflects the cheaper factorization.

\begin{figure}[!htbp]
    \centering
    \includegraphics[width=\textwidth]{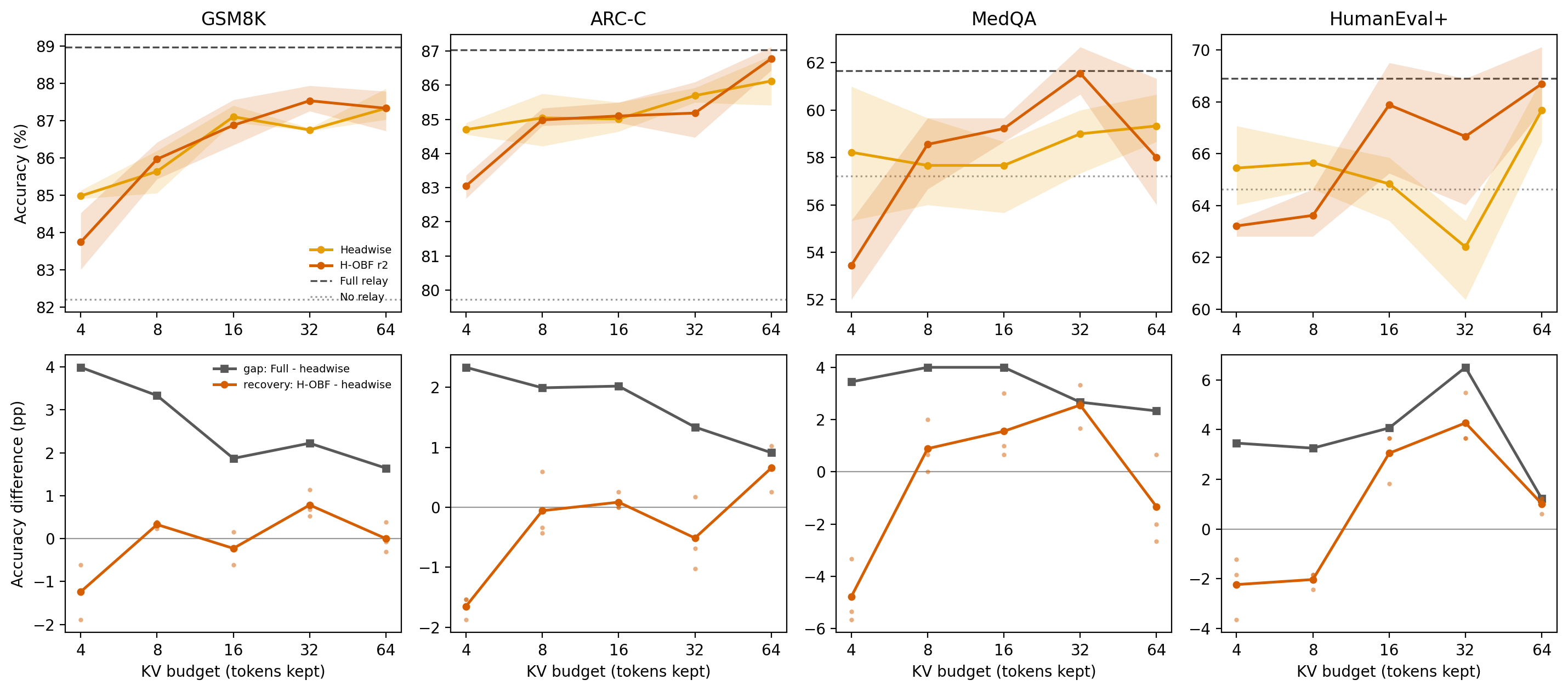}
    \caption{\textbf{Accuracy and recovery as functions of \texttt{kv\_budget}.} Top row: accuracy of H (yellow) and H-OBF (orange) at $B\in\{4,8,16,32,64\}$; shaded bands are three-seed standard deviation. Full-relay and no-relay baselines are marked. Bottom row: eviction gap ($\text{Full}-\text{H}$, grey) and OBF recovery ($\text{H-OBF}-\text{H}$, orange) as functions of budget.}
    \label{fig:budget_curve}
\end{figure}

\paragraph{Accuracy and recovery.}
OBF requires a minimum budget to work. At $B{=}4$ recovery is negative on all four benchmarks, most severely on MedQA, so injecting into a retained set that small hurts rather than helps. Recovery turns positive from $B{=}8$ onward and peaks in the middle of the sweep, where the retained span is wide enough to absorb the injected residual but eviction still discards enough for the correction to matter. The largest recoveries appear on MedQA and HumanEval+, the two benchmarks where headwise eviction opens the widest gap against Full. Budget is therefore a separate axis from the cross-benchmark recovery law of Appendix~\ref{app:recovery_law}, which holds at fixed $B{=}32$: within a benchmark, recovery does not follow the gap monotonically as $B$ varies.

One confound is that we hold $\texttt{pca\_rank}{=}2$ across the entire sweep, having selected it at $B{=}32$ (Appendix~\ref{app:pca_rank}). The retained value states span a subspace of dimension at most $B$, so the same rank is a very different intervention at different budgets: at $B{=}4$ the injected residual occupies half the available span, while at $B{=}64$ it occupies a few percent of it. The failure at $B{=}4$ is thus consistent with two readings we cannot separate here, either that OBF needs a minimum retained set regardless of rank, or that the rank should scale with the budget and $r{=}2$ over-injects when $B$ is small. Sweeping rank jointly with budget would distinguish them, and we leave that to future work.

\paragraph{Cost trade-offs.}
At equal budget H-OBF relays the same KV volume as H and uses comparable GPU memory, but its residual extraction adds latency. Headwise eviction therefore reaches a given accuracy at lower end-to-end cost on GSM8K and ARC-C, while on MedQA and HumanEval+ the accuracy gain is large enough that H-OBF sits above H at every latency it occupies.

The latency axis should be read with one caveat. Our multi-agent chain is simulated within a single process on one device, so the relayed KV cache is handed to the next agent in memory and never crosses a link. Full relay consequently pays no transfer cost in these measurements, and its latency reflects only the extra attention over a longer prompt. Any deployment that places agents on separate devices or nodes would charge Full for the full relay volume, which is roughly an order of magnitude larger than the compressed variants (top row of Figure~\ref{fig:budget_tradeoff}). The relayed-KV and GPU-memory axes are hardware independent and are unaffected by this, but the latency comparison is the most favorable reading Full can receive.

\clearpage
\section{Effect of \texttt{pca\_rank}}
\label{app:pca_rank}

We sweep \texttt{pca\_rank} over $\{2,4,8,16,32\}$ for both L-OBF and H-OBF on all nine benchmarks, with $\texttt{pca\_rank}=0$ marking the no-OBF baseline (the plain layerwise or headwise selector alone). All runs follow Appendix~\ref{app:ablation_setup}.

\begin{figure}[H]
    \centering
    \includegraphics[width=\linewidth]{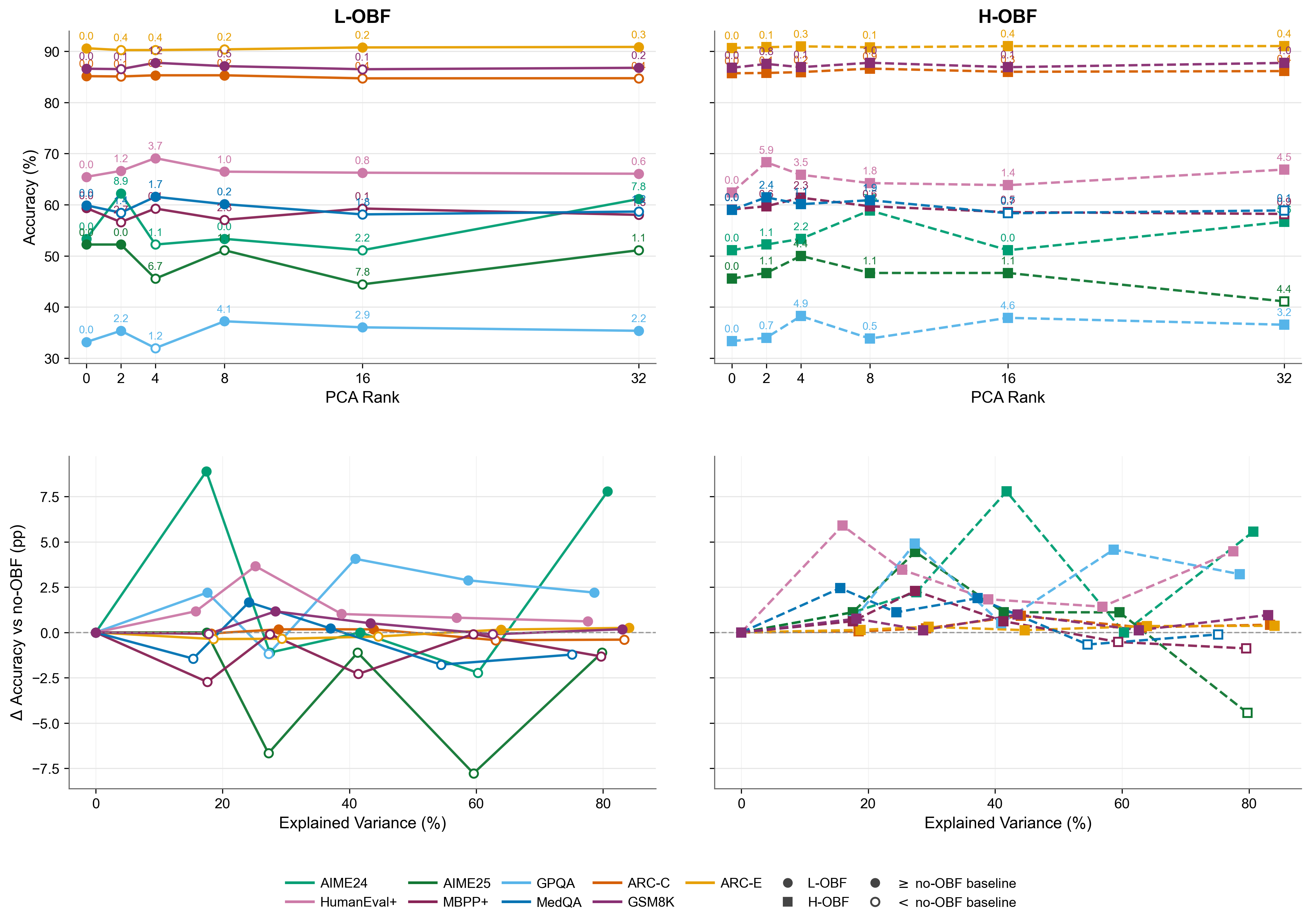}
    \caption{\textbf{Effect of \texttt{pca\_rank} on OBF accuracy across nine benchmarks.} L-OBF (left) and H-OBF (right) across ranks $\{0, 2, 4, 8, 16, 32\}$. Top row: absolute accuracy. Bottom row: gain over the rank-0 no-OBF baseline plotted against the PCA explained variance ratio. Filled markers are at or above the baseline, hollow markers below.}
    \label{fig:pca_rank_sweep}
\end{figure}

\paragraph{Analysis.}
At a given \texttt{pca\_rank}, the PCA explained variance ratio is comparable across benchmarks, so rank and EVR are largely equivalent axes. H-OBF matches or exceeds its no-OBF baseline on most benchmarks and ranks (filled markers on the right panel). Table~\ref{tab:pca_rank_scoring} aggregates the per-dataset rank orderings via a Borda count with positional weights $5, 4, 3, 2, 1$. No single rank dominates, but the $2$--$8$ range wins the majority of cells while rank $16$ is consistently near the bottom. This motivates the small fixed ranks used throughout the paper (L-OBF at $r{=}4$, H-OBF at $r{=}2$).

\begin{table}[!ht]
    \centering
    \small
    \begin{tabular}{cccc}
    \toprule
    \texttt{pca\_rank} & L-OBF & H-OBF & Combined \\
    \midrule
    4 & 31 & \textbf{31} & \textbf{62} \\
    8 & \textbf{32} & 29 & 61 \\
    2 & 26 & 28 & 54 \\
    32 & 26 & 28 & 54 \\
    16 & 20 & 19 & 39 \\
    \bottomrule
    \end{tabular}
    \caption{\textbf{Borda-count scores for each \texttt{pca\_rank}.} Higher is better; \textbf{bold} marks the top rank per column.}
    \label{tab:pca_rank_scoring}
\end{table}

\clearpage
\section{Fast SVD Path for OBF Compression}
\label{app:fast_svd}

Section~\ref{sec:pca} defines the injection direction $\Delta$ via the top-$k$ right singular vectors of the residual matrix $\mathbf{R}\in\mathbb{R}^{N\times d}$. In our experiments $k = \texttt{pca\_rank} \leq 32$ while $N$ is thousands of tokens per relay boundary, so full SVD is dominated by rows the algorithm never uses. We replace it with a Gram-matrix subspace iteration.

\paragraph{Algorithm.}
Instead of decomposing $\mathbf{R}$ directly, we form the $d\times d$ Gram matrix $\mathbf{G} = \mathbf{R}^{\top}\mathbf{R}$, initialize $\mathbf{C}_0 \in \mathbb{R}^{k\times d}$ with orthonormal rows, and iterate
\[
\mathbf{C}_{t+1} \leftarrow \mathrm{orth}\big(\mathbf{C}_t \, \mathbf{G}\big), \qquad t = 0, 1, 2,
\]
where $\mathrm{orth}(\cdot)$ is a thin QR. After three iterations $\mathbf{C}$ spans the top-$k$ right singular subspace of $\mathbf{R}$ to numerical precision; the injection direction $\Delta = (\bar{\mathbf{r}}\mathbf{C}^{\top})\mathbf{C}$ is then computed as in Eq.~\ref{eq:delta}. The cost drops from $O(N d \min(N, d))$ to $O(N d^2)$ for the Gram formation plus $O(k d^2)$ per iteration, and the $O(N d^2)$ term parallelizes as a single dense matmul.

\paragraph{Setup for timing.}
We run the exact and fast paths on Qwen3-4B, matched on (benchmark, seed) at the setup in Appendix~\ref{app:ablation_setup}. \emph{Compression time per sample} is the wall clock spent building $\Delta$ and injecting it, averaged across all relay boundaries in a rollout. AIME24 and AIME25 are excluded because their small sample size makes per-sample time estimates unstable.

\begin{figure}[!htbp]
    \centering
    \includegraphics[width=\textwidth]{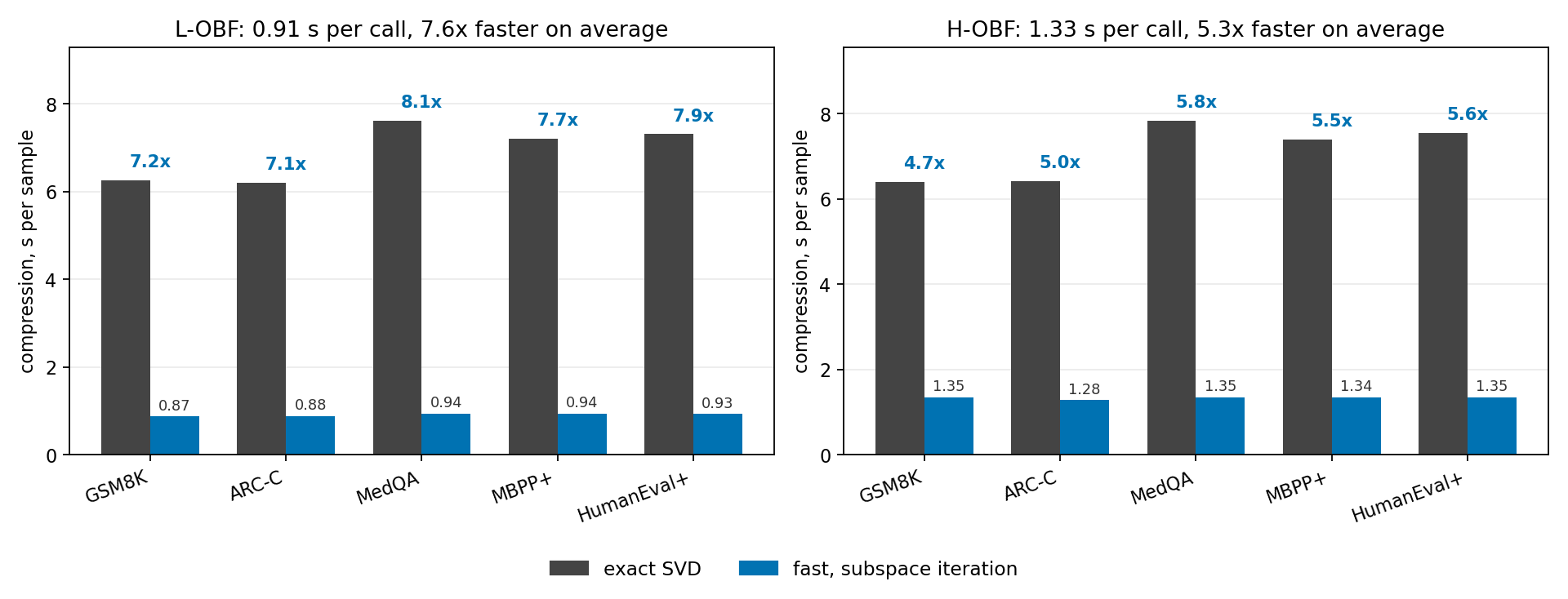}
    \caption{\textbf{Fast SVD path compression time.} Per-sample compression time (three-seed mean) for the exact and fast paths on Qwen3-4B (A100 40GB PCIe). Annotations above each pair are the per-benchmark speedup.}
    \label{fig:fast_compression_time}
\end{figure}

\paragraph{Timing.}
Figure~\ref{fig:fast_compression_time} shows compression time per sample for the exact and fast paths on each benchmark. Averaged across the five benchmarks, the fast path drops compression from $\sim 7.1$~s to $0.91$~s per sample for L-OBF ($7.6\times$ on average, $7.1$ to $8.1\times$ across benchmarks) and from $\sim 7.1$~s to $1.33$~s for H-OBF ($5.3\times$ on average, $4.7$ to $5.8\times$).

The end-to-end impact of this speedup, together with a full comparison against Full and the eviction-only baselines, is reported in Appendix~\ref{app:cost_breakdown}.

\begin{figure}[!htbp]
    \centering
    \includegraphics[width=\textwidth]{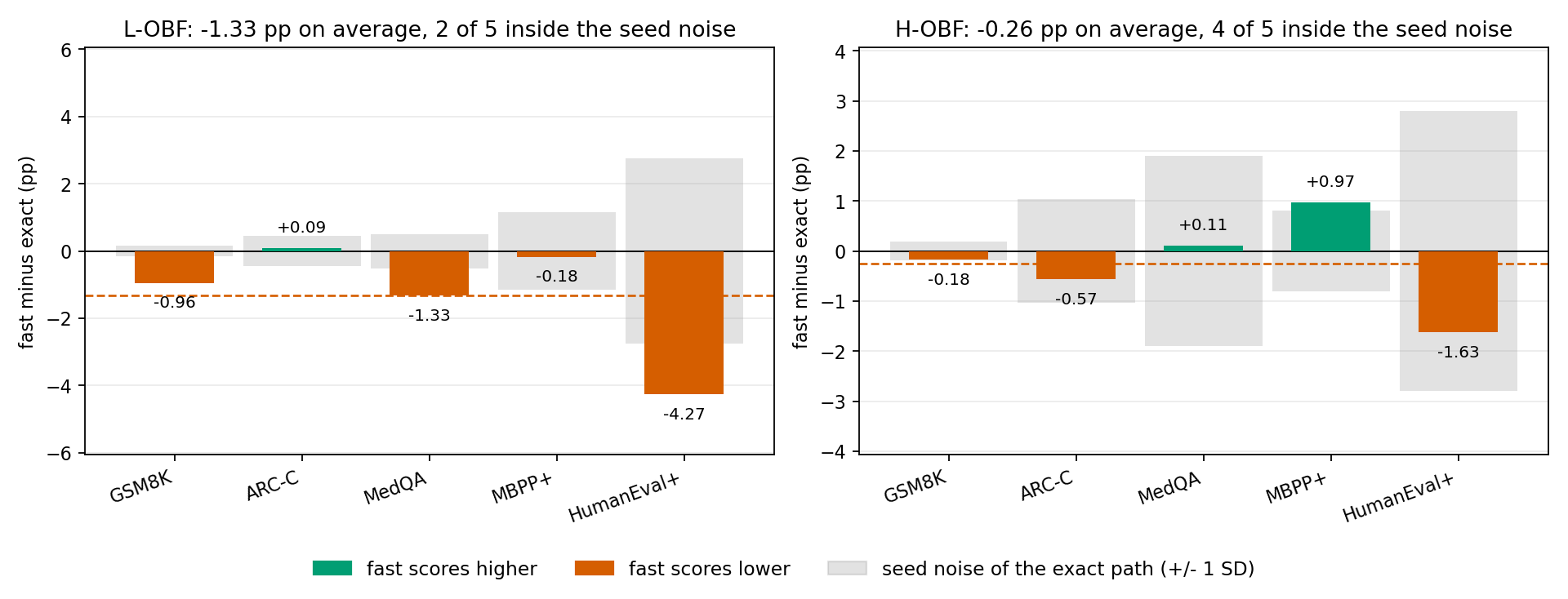}
    \caption{\textbf{Fast SVD path accuracy against exact.} Per-benchmark fast-minus-exact accuracy delta (pp, three-seed mean). Grey bands are $\pm 1$~SD of the exact path across its own seeds; bars inside the band are indistinguishable from sampling noise.}
    \label{fig:fast_accuracy}
\end{figure}

\paragraph{Accuracy verification.}
The subspace-iteration output matches the exact top-$k$ right singular vectors to $\sim 10^{-4}$ relative error, but a perturbation this small can still flip a sampled token during decoding, so we verify accuracy directly rather than relying on numerical agreement. Figure~\ref{fig:fast_accuracy} plots the fast-minus-exact accuracy delta per benchmark against the exact path's own seed noise band ($\pm 1$~SD).

For H-OBF, the fast path shifts accuracy by $-0.26$~pp on average, with four of five benchmarks inside the seed noise band; the fast-vs-exact delta is inside the range the exact path already covers on its own.

For L-OBF, the average shift is $-1.33$~pp, concentrated on HumanEval+ ($-4.27$~pp seed-averaged; the other four benchmarks lie between $-1.33$ and $+0.09$~pp). HumanEval+ has the longest generation ($\sim 2000$ tokens), the smallest test set (164 problems), and a pass@1 metric that is sensitive to single-token substitutions in code, all of which amplify any per-step drift. We keep the fast path as the main-text default because the accuracy conclusions on the other benchmarks are unchanged, and flag this residual drift here.

\clearpage
\section{End-to-End Cost Breakdown}
\label{app:cost_breakdown}

We decompose the total cost of one rollout into four quantities: relay bandwidth per non-judger agent, compression work per relay boundary, tokens generated at the final agent, and end-to-end wall clock. All runs follow Appendix~\ref{app:ablation_setup}, restricted to the five benchmarks with three-seed coverage for every method (GSM8K, ARC-C, MedQA, MBPP+, HumanEval+).

\subsection{Cost Across Compression Methods}
\label{app:cost_methods}

\begin{table*}[!htbp]
\centering
\setlength{\tabcolsep}{4pt}
\renewcommand{\arraystretch}{1.1}
\resizebox{\textwidth}{!}{%
{\footnotesize
\begin{tabular}{l|cc|c|c|cc|c}
\toprule
Method & Relay (MB) & Relay ratio & Compression (s) & Output tokens & End-to-end (s) & E2E ratio & Avg GPU (GB) \\
\midrule
Full           & $290.1$ & $1.00\times$ & $0.00$ & $1149$ & $24.67$ & $1.00\times$ & $13.0$ \\
\midrule
L              & $62.4$  & $4.65\times$ & $0.05$ & $1247$ & $\mathbf{24.31}$ & $\mathbf{0.99\times}$ & $12.0$ \\
L-OBF (fast)   & $62.4$  & $4.65\times$ & $0.91$ & $1274$ & $25.79$ & $1.05\times$ & $\mathbf{11.8}$ \\
H              & $61.6$  & $4.71\times$ & $0.07$ & $1245$ & $24.85$ & $1.01\times$ & $12.0$ \\
H-OBF (fast)   & $61.4$  & $4.72\times$ & $1.30$ & $1239$ & $25.97$ & $1.05\times$ & $12.0$ \\
Gen            & $\mathbf{35.4}$ & $\mathbf{8.19\times}$ & $0.00$ & $1269$ & $25.97$ & $1.05\times$ & $12.0$ \\
\bottomrule
\end{tabular}%
}%
}
\setlength{\abovecaptionskip}{12pt}
\caption{\textbf{End-to-end cost of one rollout across compression methods.} \emph{Relay} is the KV size transferred per non-judger agent; ratio columns divide by Full's value. OBF uses the fast SVD path (Appendix~\ref{app:fast_svd}). Bold marks the minimum in each cost column.}
\label{tab:cost_methods}
\end{table*}

\paragraph{Analysis.}
Eviction-based methods transfer roughly $62$~MB against Full's $290$~MB, a $4.7\times$ reduction, and Gen goes further to $35$~MB ($8.2\times$). Full is the fastest end-to-end because it pays no compression cost. Every compressed method sits within $5\%$ of Full's wall clock even after paying for compression, with L essentially free and OBF adding about $1$~s per sample under the fast SVD path (Appendix~\ref{app:fast_svd}). This wall-clock parity should be read against the setup. Our multi-agent chain runs in a single process on one device, so the relayed cache is handed to the next agent in memory and Full never pays to move its $290$~MB. The relay and GPU-memory columns are properties of the method and carry over to any deployment; the end-to-end column is specific to a co-located chain and is the most favorable reading Full can receive.

\subsection{Fast SVD vs Exact SVD for OBF}
\label{app:cost_fast_vs_exact}

Relay bandwidth and output tokens are unchanged between the exact and fast SVD paths, so Table~\ref{tab:cost_fast_vs_exact} focuses on compression time and end-to-end wall clock.

\begin{table}[!htbp]
\centering
\setlength{\tabcolsep}{6pt}
\renewcommand{\arraystretch}{1.1}
{\footnotesize
\begin{tabular}{l|c|c|cc}
\toprule
Method & Compression (s) & Compression speedup & End-to-end (s) & E2E ratio \\
\midrule
Full          & $0.00$ & --           & $24.67$ & $1.00\times$ \\
\midrule
L-OBF (exact) & $6.93$ & $1.0\times$   & $31.46$ & $1.28\times$ \\
L-OBF (fast)  & $\mathbf{0.91}$ & $\mathbf{7.6\times}$   & $25.79$ & $\mathbf{1.05\times}$ \\
\midrule
H-OBF (exact) & $7.14$ & $1.0\times$   & $31.78$ & $1.29\times$ \\
H-OBF (fast)  & $\mathbf{1.30}$ & $\mathbf{5.5\times}$   & $25.97$ & $\mathbf{1.05\times}$ \\
\bottomrule
\end{tabular}%
}
\setlength{\abovecaptionskip}{12pt}
\caption{\textbf{Fast SVD path against exact SVD for OBF.} \emph{Compression speedup} is exact time divided by fast time. \emph{E2E ratio} divides end-to-end wall clock by Full's.}
\label{tab:cost_fast_vs_exact}
\end{table}

\paragraph{Analysis.}
Under the exact SVD, OBF's $\sim 7$~s per-sample compression cost pushes end-to-end wall clock to $1.28\times$--$1.29\times$ Full, which is not a competitive operating point. The fast path (Appendix~\ref{app:fast_svd}) cuts compression by $5.5\times$--$7.6\times$ and brings end-to-end to $1.05\times$ Full for both selectors, so the $4.7\times$ bandwidth reduction from Table~\ref{tab:cost_methods} is delivered at essentially matched wall clock.

\clearpage

\section{Additional Analysis of Downstream Generation Cost}
\label{app:downstream_cost}
We examine how accuracy correlates with per-agent runtime cost, using the main-experiment runs (Appendix~\ref{app:ablation_setup}). Each panel overlays a dashed least-squares fit and reports the Pearson correlation coefficient $r$ in the dataset label.

\subsection{Latent-Side Cost}
\label{app:acc_vs_latent_time}

\begin{figure}[H]
    \centering
    \includegraphics[width=\textwidth]{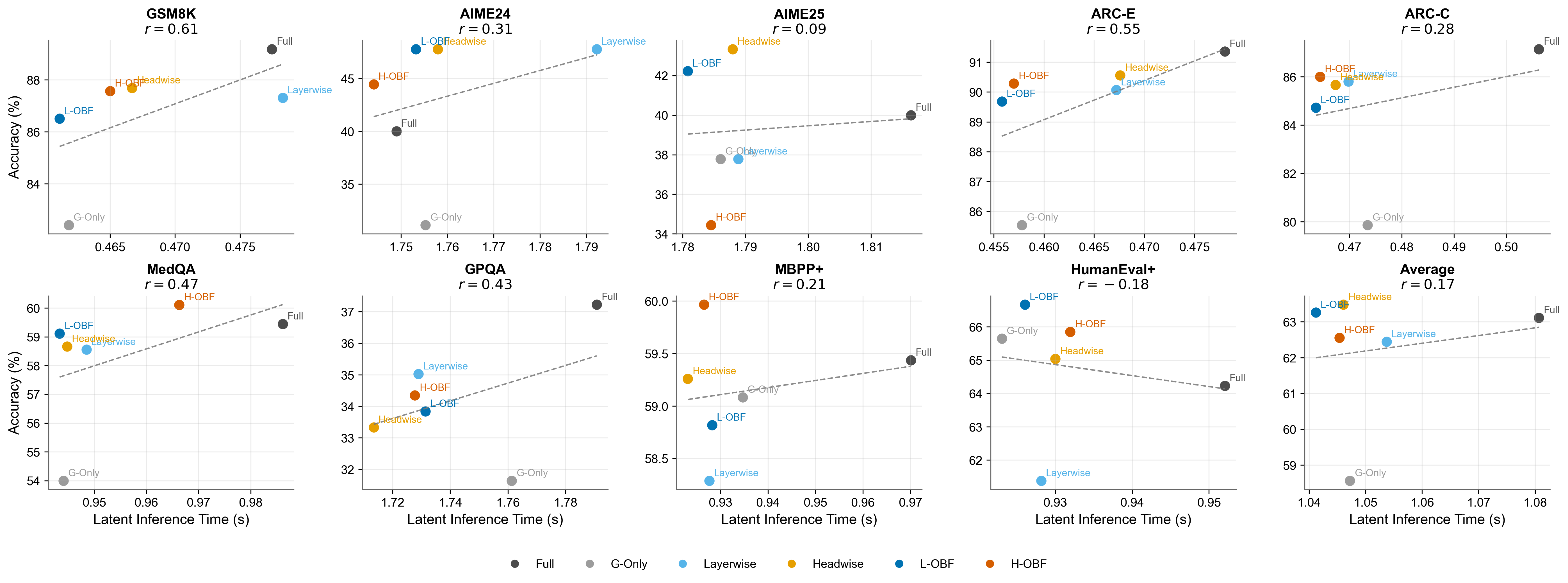}
    \caption{\textbf{Accuracy versus latent inference time.} Latent inference time is reported as the average runtime across all agents that perform latent inference.}
    \label{fig:acc_vs_latent_time}
\end{figure}

\paragraph{Analysis.}
The correlations are positive but weak (cross-dataset average $r=+0.17$), so latent inference time is a poor predictor of downstream accuracy.

\subsection{Text-Side Cost}
\label{app:text_side_cost}
\begin{figure}[H]
    \centering
    \includegraphics[width=\textwidth]{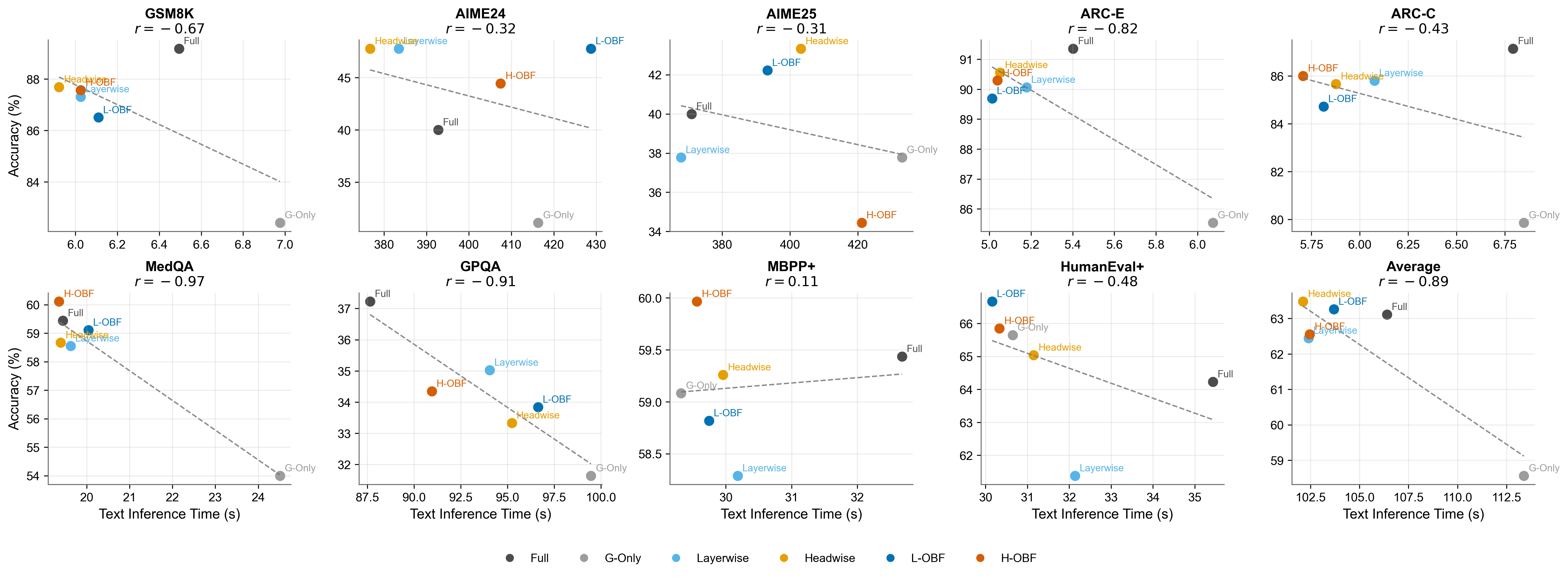}
    \caption{\textbf{Accuracy versus text inference time.} This figure measures the runtime of text generation by the final agent.}
    \label{fig:acc_vs_text_time}
\end{figure}

\begin{figure}[H]
    \centering
    \includegraphics[width=\textwidth]{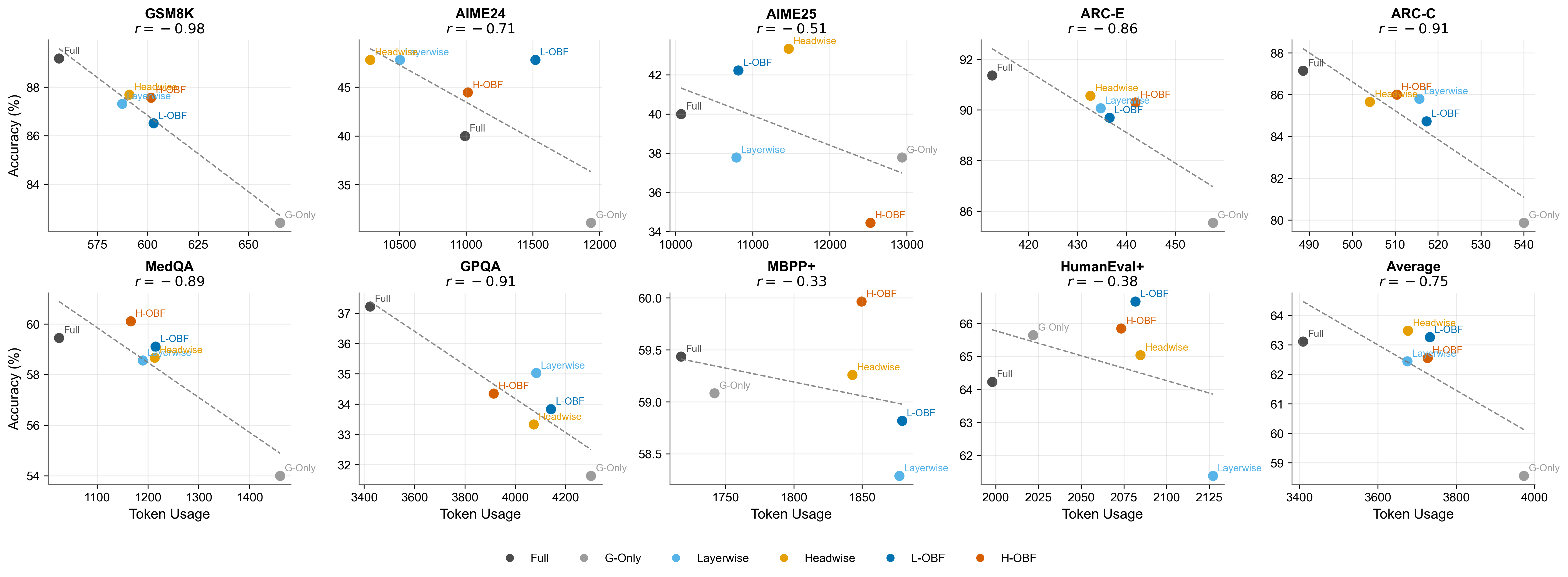}
    \caption{\textbf{Accuracy versus final-agent token usage.} Token usage counts only the tokens generated by the final agent.}
    \label{fig:acc_vs_token_usage}
\end{figure}

\paragraph{Analysis.}
Both text inference time and final-agent token usage correlate strongly negatively with accuracy (cross-dataset averages $r=-0.89$ and $r=-0.75$), so text-side cost is a reliable accuracy signal whereas latent time is not. On reasoning and QA benchmarks, shorter generations co-occur with higher accuracy, suggesting extra tokens reflect compensation for compression loss rather than productive reasoning; the pattern weakens on coding. On average, L-OBF and H-OBF sit closer to the upper-left of both text-side panels than their non-OBF counterparts, indicating OBF retains accuracy while reducing text-side cost.

\clearpage
\section{Internal Behavior of OBF}
\label{app:internal_diagnostics}
\subsection{Diagnostic Setup and Representative Example}

This appendix examines how OBF behaves internally across layers and heads, complementing the benchmark-level results in the main text. We report a set of diagnostic curves that characterize the magnitude, recoverability, and relative contribution of the injected residual vector throughout the network. All figures in this section follow the shared ablation configuration in Appendix~\ref{app:ablation_setup}. Figure~\ref{fig:obf_internal_metrics_gsm8k} shows the diagnostic curves of L-OBF on GSM8K as a running example, run on the NVIDIA A100-PCIE-40GB setup following Appendix~\ref{app:hardware}.

\begin{figure}[H]
    \centering
    \includegraphics[width=\linewidth]{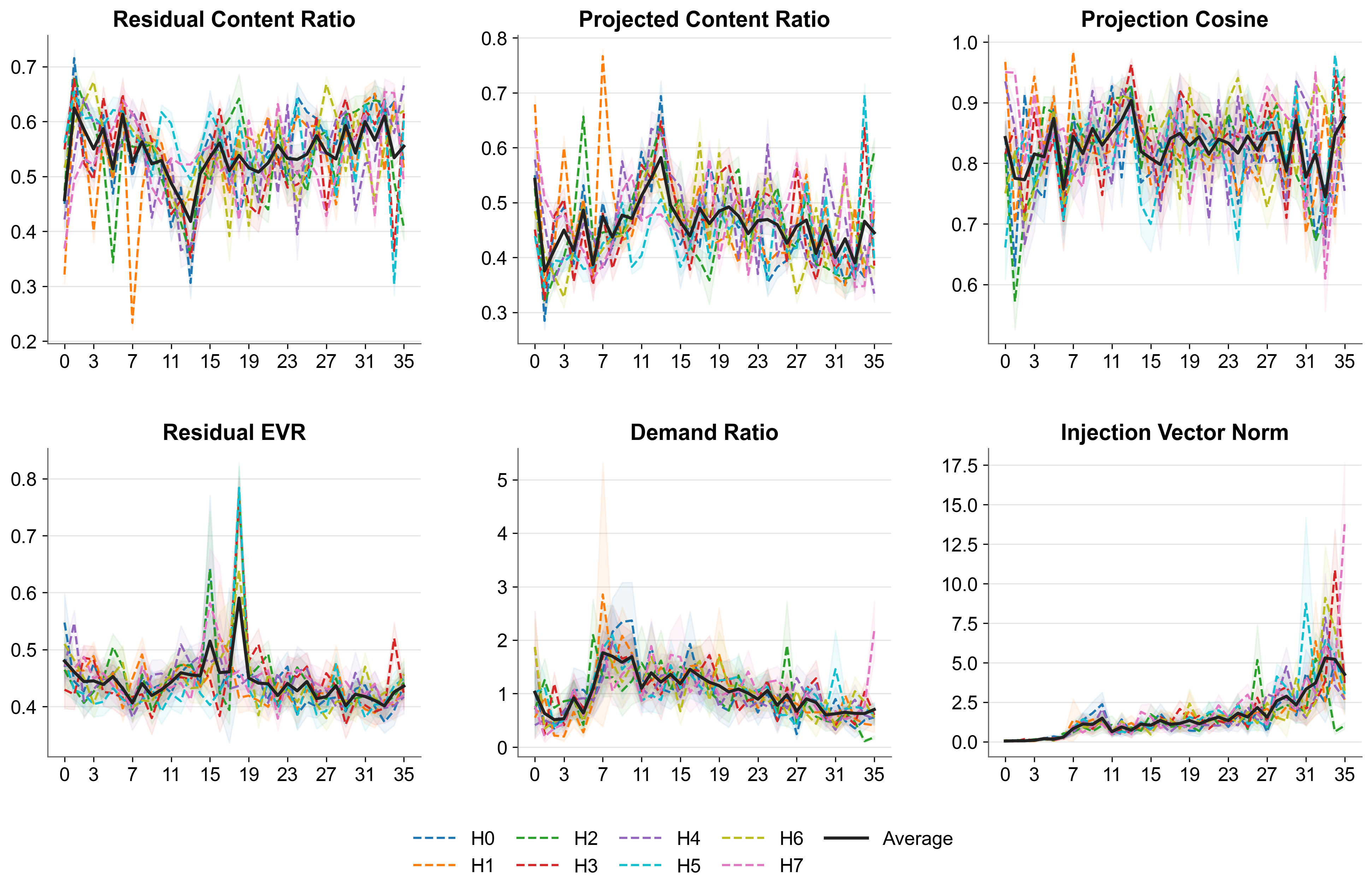}
    \caption{\textbf{Representative internal diagnostics of L-OBF on GSM8K.} The six panels report the metrics defined in Appendix~\ref{app:internal_metrics}; the horizontal axis is the layer index. Colored dashed curves correspond to individual attention head groups, the solid black curve shows the across-head average, and the shaded region indicates variability across samples.}
    \label{fig:obf_internal_metrics_gsm8k}
\end{figure}

\subsection{Diagnostic Metrics}
\label{app:internal_metrics}

We define the internal diagnostic quantities used throughout this appendix. All metrics are computed for each layer-head pair $(l,h)$ on the residual matrix $\mathbf{R}$.

\paragraph{Projected Content Ratio (PCR) and Residual Content Ratio (RCR).}
Recall from Eq.~\ref{eq:pca_extra_residual} that the deleted value states can be decomposed as
\[
\mathbf{V}_{\mathrm{del}} = \mathbf{V}_{\parallel} + \mathbf{R},
\qquad\text{where }
\mathbf{V}_{\parallel} = (\mathbf{V}_{\mathrm{del}}\mathbf{Q})\mathbf{Q}^{\top}.
\]
We define
\[
\mathrm{PCR}
=
\frac{\|\mathbf{V}_{\parallel}\|_F^2}{\|\mathbf{V}_{\mathrm{del}}\|_F^2+\epsilon},
\qquad
\mathrm{RCR}
=
\frac{\|\mathbf{R}\|_F^2}{\|\mathbf{V}_{\mathrm{del}}\|_F^2+\epsilon}.
\]
Since $\mathbf{V}_{\parallel}$ is the projection of $\mathbf{V}_{\mathrm{del}}$ onto the span of the kept value states, $\mathbf{V}_{\parallel}$ and $\mathbf{R}$ form an orthogonal decomposition of $\mathbf{V}_{\mathrm{del}}$, so $\mathrm{PCR}+\mathrm{RCR}\approx 1$. PCR measures how much of the deleted information is representable by that span; RCR measures how much is not.

\paragraph{Projection Cosine (PC).}
We also define
\[
\mathrm{PC}
=
\mathrm{CosSim}\!\left(
\sum_{j=1}^{|\mathcal{P}_{\mathrm{del}}|}\mathbf{V}_{\mathrm{del}}[j,:],
\sum_{j=1}^{|\mathcal{P}_{\mathrm{del}}|}\mathbf{V}_{\parallel}[j,:]
\right),
\]
which measures whether the representable part remains directionally consistent with the deleted information.

\paragraph{Residual Explained Variance Ratio (REVR).}
Recall from Section~\ref{sec:pca} that OBF applies SVD to the residual matrix $\mathbf{R}$ and retains its top-$k$ right singular vectors. Let $\sigma_1,\ldots,\sigma_m$ denote the singular values of $\mathbf{R}$. We define
\begin{equation}
\mathrm{REVR}
=
\frac{\sum_{j=1}^{p}\sigma_j^2}{\sum_{j=1}^{m}\sigma_j^2+\epsilon}.
\end{equation}
REVR measures how much of the residual information the top-$p$ principal subspace retains; a high value indicates low-rank structure.

\paragraph{Deleted-versus-Retained Demand Ratio (DR).}
Recall from Eq.~\ref{eq:demand_ratio} that OBF uses the deleted-versus-retained demand ratio for dynamic scaling. We define
\[
\mathrm{DR}
=
\frac{\tilde{A}_{\mathrm{del}}}{\tilde{A}_{\mathrm{keep}}+\epsilon}.
\]
DR measures how much attention falls on deleted positions relative to kept ones.

\paragraph{Injection Vector Norm (IVN).}
Recall from Eq.~\ref{eq:demand_ratio} that OBF forms the final injected correction $\delta$ after dynamic scaling. We define
\[
\mathrm{IVN}=\|\delta\|_2.
\]
Under the uniform injection rule in Eq.~\ref{eq:pca_extra_injection_uniform}, the same $\delta$ is added to all kept positions, so IVN captures the overall strength of OBF's correction.

\subsection{L-OBF vs.\ H-OBF}
\label{app:internal_compare_methods}

\begin{figure*}[t]
    \centering
    \includegraphics[width=\linewidth]{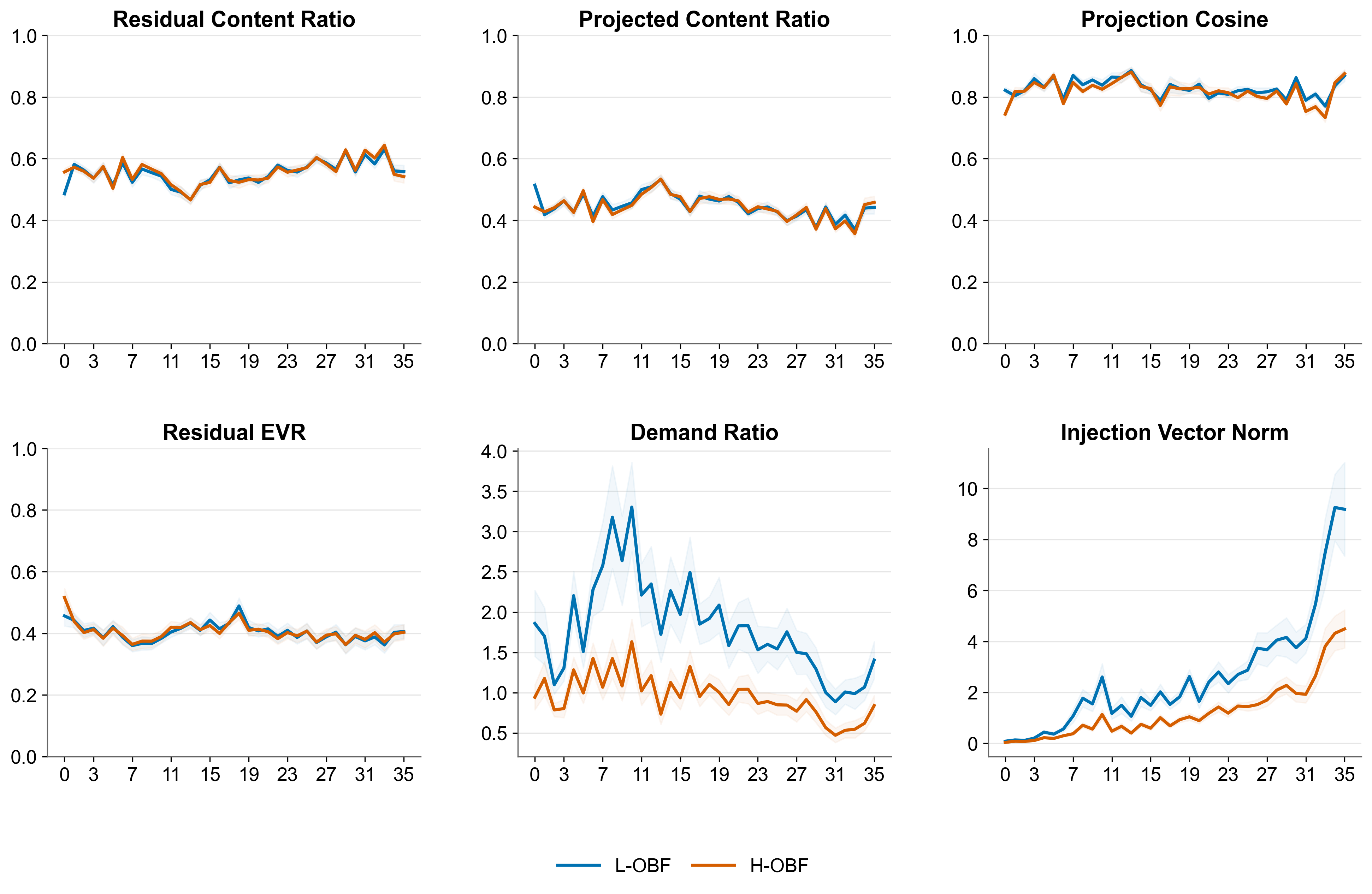}
    \caption{\textbf{Internal diagnostics of L-OBF versus H-OBF}, averaged across the seven main-text benchmarks. The first four metrics (PCR, RCR, PC, REVR) are nearly identical between the two methods; L-OBF shows a larger Demand Ratio and Injection Vector Norm, especially in later layers.}
    \label{fig:internal_states_compare}
\end{figure*}

Figure~\ref{fig:internal_states_compare} compares L-OBF and H-OBF averaged across the seven main-text benchmarks, using the default configuration in Appendix~\ref{app:ablation_setup}.

\paragraph{Representability metrics.} Across both methods, PCR is about $40\%$--$50\%$, PC stays around $0.8$ or above, and REVR is also about $40\%$--$50\%$, meaning that roughly half of the deleted information is representable by the kept value states, that representable part is directionally consistent, and the first $p=8$ residual components capture about half of the residual energy.

\paragraph{Injection-side gap.} The gap between L-OBF and H-OBF concentrates in Demand Ratio and Injection Vector Norm, especially in later layers. A plausible reason is that layerwise selection uses a retained set shared across heads: for some heads the kept tokens are not the ones receiving the highest attention, which increases deleted demand and decreases kept demand, thereby raising DR and the injection strength.

\subsection{Effect of \texttt{kv\_budget}}
\label{app:internal_budget_sweep}

\begin{figure*}[t]
    \centering
    \includegraphics[width=\linewidth]{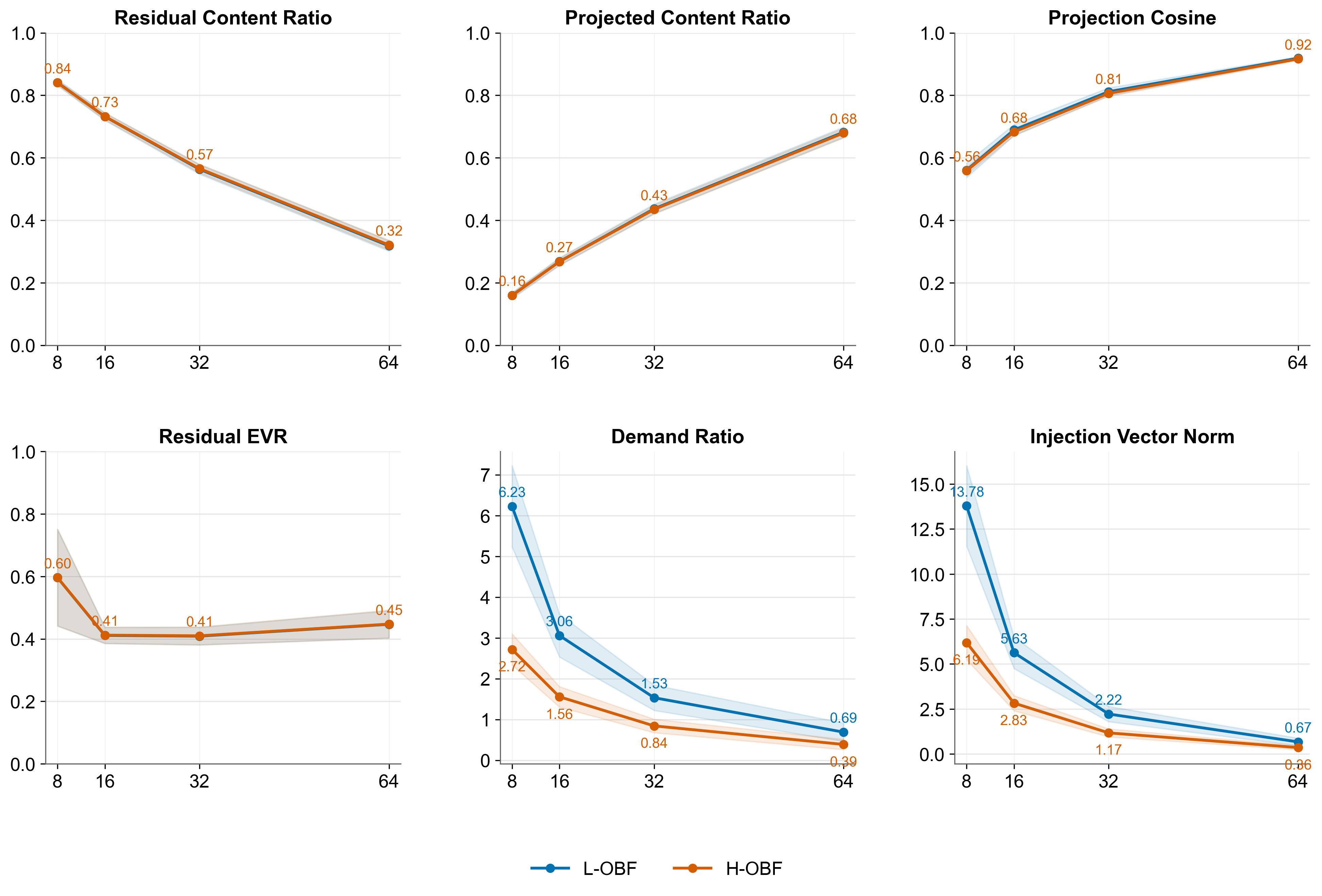}
    \caption{\textbf{Internal diagnostics under different keep budgets} ($\texttt{kv\_budget}\in\{8,16,32,64\}$). As the budget grows, more deleted content becomes representable by the kept value states (PCR$\uparrow$, RCR$\downarrow$, PC$\uparrow$); the L-OBF--H-OBF gap in Demand Ratio and Injection Vector Norm persists but narrows.}
    \label{fig:internal_states_budget_sweep}
\end{figure*}

Figure~\ref{fig:internal_states_budget_sweep} shows the internal diagnostics under different keep budgets, using the setting described in Appendix~\ref{app:budget_sweep}.

\paragraph{First four panels.} Increasing the keep budget makes more deleted information representable by the kept value states. The Projected Content Ratio rises from about $0.16$ at \(\texttt{kv\_budget}=8\) to about $0.68$ at \(\texttt{kv\_budget}=64\), with the Residual Content Ratio decreasing accordingly. The Projection Cosine also increases, indicating stronger directional consistency. Residual EVR, by contrast, follows a rough U-shape: at small budgets less deleted information is absorbed by the projection, so more structure remains in the residual and PCA captures a relatively larger fraction of it; at large budgets the residual itself shrinks, which again makes it easier to cover.

\paragraph{Last two panels.} Both Demand Ratio and Injection Vector Norm decrease monotonically as the keep budget grows, because a larger budget raises $A_{\mathrm{keep}}$ and lowers $A_{\mathrm{del}}$ in Eq.~\ref{eq:demand_ratio}, so the deleted-versus-retained demand ratio shrinks and less correction is needed. The L-OBF--H-OBF gap from Section~\ref{app:internal_compare_methods} persists across all budgets.

\subsection{Effect of \texttt{pca\_rank}}
\label{app:internal_pca_rank_sweep}

\begin{figure*}[t]
    \centering
    \includegraphics[width=\linewidth]{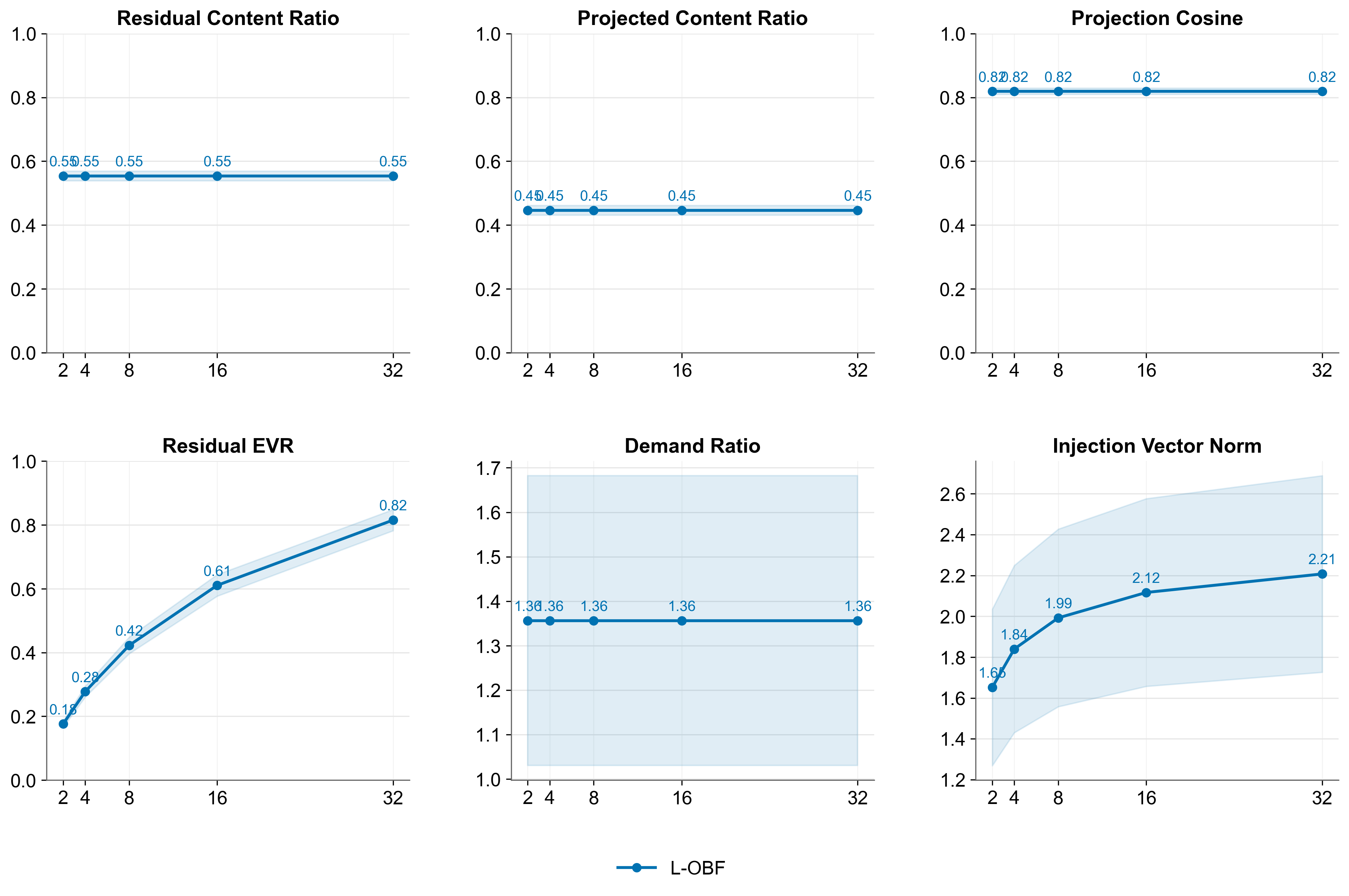}
    \caption{\textbf{Internal diagnostics under different PCA ranks} ($\texttt{pca\_rank}\in\{2,4,8,16,32\}$). Only Residual EVR and Injection Vector Norm increase monotonically with rank; PCR, RCR, PC, and Demand Ratio are essentially unchanged. REVR remains below $0.95$ even at rank $32$.}
    \label{fig:internal_states_pca_rank_sweep}
\end{figure*}

Figure~\ref{fig:internal_states_pca_rank_sweep} shows the internal diagnostics under different PCA ranks, using the setting of Appendix~\ref{app:pca_rank}.
Only Residual EVR and Injection Vector Norm increase monotonically with the PCA rank; the remaining metrics stay nearly constant. REVR rises from about $0.18$ at \texttt{pca\_rank=2} to about $0.82$ at \texttt{pca\_rank=32}, and IVN increases from about $1.65$ to about $2.21$.

This is expected: changing the PCA rank does not change which tokens are kept or deleted, so the projection step is unchanged and PCR, RCR, PC, and Demand Ratio remain nearly fixed. Retaining more residual principal directions directly raises REVR, and because the injected vector is built from the retained residual subspace, IVN rises with it. Even at \texttt{pca\_rank=32}, REVR stays below $0.95$, so the residual information is spread over a broad subspace rather than concentrated in a few directions.

\clearpage
\section{Attention-Output Error Bounds for Prompt Compression}
\label{app:error_bound}

We derive, for a single layer-head, the attention-output error introduced when a downstream query attends to a compressed prompt KV. The analysis makes explicit the \emph{softmax renormalization} that amplifies retained positions after deletion, and uses that fact to explain why OBF injects \emph{orthogonally} to the retained value subspace. For brevity we fix a layer-head and drop the $(l,h)$ index throughout.

\subsection{Setup}
\label{app:error_bound:setup}

Consider a downstream query $\mathbf{q}\in\mathbb{R}^{d_k}$ attending to prompt positions $\mathcal{P}=\{1,\dots,T\}$ with keys $\mathbf{K}\in\mathbb{R}^{T\times d_k}$ and values $\mathbf{V}\in\mathbb{R}^{T\times d_v}$. Let $s_j = \mathbf{q}\mathbf{K}_{j,:}^{\top}/\sqrt{d_k}$ and define the attention mass $\alpha_j = \exp(s_j)/Z$ with $Z = \sum_{j\in\mathcal{P}}\exp(s_j)$. The full attention output is
\[
\mathbf{o}_{\mathrm{full}} \;=\; \sum_{j\in\mathcal{P}} \alpha_j\,\mathbf{v}_j .
\]
Partition $\mathcal{P} = \mathcal{P}_{\mathrm{keep}} \sqcup \mathcal{P}_{\mathrm{del}}$ and write
\[
A_{\mathrm{keep}} \;=\; \sum_{j\in\mathcal{P}_{\mathrm{keep}}}\alpha_j,
\qquad
A_{\mathrm{del}} \;=\; \sum_{j\in\mathcal{P}_{\mathrm{del}}}\alpha_j \;=\; 1 - A_{\mathrm{keep}} .
\]
Define the attention-weighted conditional means
\[
\bar{\mathbf{v}}_{\mathrm{keep}} \;=\; \frac{1}{A_{\mathrm{keep}}}\sum_{j\in\mathcal{P}_{\mathrm{keep}}}\alpha_j\,\mathbf{v}_j,
\qquad
\bar{\mathbf{v}}_{\mathrm{del}} \;=\; \frac{1}{A_{\mathrm{del}}}\sum_{j\in\mathcal{P}_{\mathrm{del}}}\alpha_j\,\mathbf{v}_j ,
\]
so that $\mathbf{o}_{\mathrm{full}} = A_{\mathrm{keep}}\bar{\mathbf{v}}_{\mathrm{keep}} + A_{\mathrm{del}}\bar{\mathbf{v}}_{\mathrm{del}}$.

\subsection{Softmax renormalization amplifies retained positions}
\label{app:error_bound:amplify}

Deleting positions removes the logits $\{s_j\}_{j\in\mathcal{P}_{\mathrm{del}}}$ from the partition function, so the softmax over the remaining keys renormalizes.

\begin{proposition}[Amplification factor]
\label{prop:amplification}
After eviction, the attention weight on each retained position is
\[
\tilde{\alpha}_j \;=\; \frac{\exp(s_j)}{\sum_{k\in\mathcal{P}_{\mathrm{keep}}}\exp(s_k)} \;=\; \frac{\alpha_j}{A_{\mathrm{keep}}} \;=\; \frac{\alpha_j}{1 - A_{\mathrm{del}}}, \qquad j\in\mathcal{P}_{\mathrm{keep}} .
\]
Every retained weight is amplified by the factor $1/(1-A_{\mathrm{del}})$ relative to the full softmax. Consequently, the pure-eviction output collapses to the conditional mean over retained positions,
\[
\mathbf{o}_{\mathrm{evict}} \;=\; \sum_{j\in\mathcal{P}_{\mathrm{keep}}} \tilde{\alpha}_j \mathbf{v}_j \;=\; \bar{\mathbf{v}}_{\mathrm{keep}} .
\]
\end{proposition}

The amplification is load-bearing for the rest of the analysis: any correction $\mathbf{r}_j$ added to a retained value contributes $\tilde{\alpha}_j\mathbf{r}_j = \alpha_j\mathbf{r}_j/(1-A_{\mathrm{del}})$ to the output, so a compressor that writes into the retained value space inherits the same amplification factor.

\subsection{Hard-eviction error}
\label{app:error_bound:evict}

Proposition~\ref{prop:amplification} immediately yields the error of hard eviction.

\begin{proposition}[Hard-eviction error]
\label{prop:evict_error}
\[
\mathbf{e}_{\mathrm{evict}} \;:=\; \mathbf{o}_{\mathrm{full}} - \mathbf{o}_{\mathrm{evict}} \;=\; A_{\mathrm{del}}\bigl(\bar{\mathbf{v}}_{\mathrm{del}} - \bar{\mathbf{v}}_{\mathrm{keep}}\bigr),
\]
so $\|\mathbf{e}_{\mathrm{evict}}\| \le A_{\mathrm{del}}\,\bigl\|\bar{\mathbf{v}}_{\mathrm{del}} - \bar{\mathbf{v}}_{\mathrm{keep}}\bigr\|$.
\end{proposition}

The error factors into two quantities: (i) the deleted attention mass $A_{\mathrm{del}}$, which attention-based selection ($\Phi_{\mathrm{Attn}}$) explicitly minimizes given a budget $B$; and (ii) the conditional-mean gap $\|\bar{\mathbf{v}}_{\mathrm{del}} - \bar{\mathbf{v}}_{\mathrm{keep}}\|$, which attention-based selection does not control. When the retained values are geometrically similar to the deleted values, the gap is small and eviction alone is nearly lossless; otherwise, small $A_{\mathrm{del}}$ is insufficient.

\subsection{OBF error bound}
\label{app:error_bound:obf}

OBF adds a value-space residual $\mathbf{r}_j$ to each retained position so that the compressed output becomes
\[
\mathbf{o}_{\mathrm{OBF}} \;=\; \sum_{j\in\mathcal{P}_{\mathrm{keep}}}\tilde{\alpha}_j\bigl(\mathbf{v}_j + \mathbf{r}_j\bigr) \;=\; \bar{\mathbf{v}}_{\mathrm{keep}} + \sum_{j\in\mathcal{P}_{\mathrm{keep}}}\tilde{\alpha}_j\mathbf{r}_j .
\]
Comparing with Proposition~\ref{prop:evict_error}, zero-error reconstruction would require
\begin{equation}
\sum_{j\in\mathcal{P}_{\mathrm{keep}}}\tilde{\alpha}_j\mathbf{r}_j \;=\; A_{\mathrm{del}}\bigl(\bar{\mathbf{v}}_{\mathrm{del}} - \bar{\mathbf{v}}_{\mathrm{keep}}\bigr) \;=:\; \boldsymbol{\tau} .
\label{eq:obf_target}
\end{equation}
We call $\boldsymbol{\tau}\in\mathbb{R}^{d_v}$ the \emph{correction target}. Under a rank-$k$ budget on the injection directions, OBF realizes only the projection of $\boldsymbol{\tau}$ onto a $k$-dimensional subspace $U_k\subseteq\mathbb{R}^{d_v}$.

\begin{proposition}[Rank-$k$ OBF error]
\label{prop:obf_error}
Let $\Pi_{U_k}$ denote orthogonal projection onto $U_k$. If the injection $\sum_j\tilde{\alpha}_j\mathbf{r}_j$ equals $\Pi_{U_k}\boldsymbol{\tau}$, then
\[
\mathbf{e}_{\mathrm{OBF}} \;:=\; \mathbf{o}_{\mathrm{full}} - \mathbf{o}_{\mathrm{OBF}} \;=\; (I - \Pi_{U_k})\,\boldsymbol{\tau},
\qquad
\|\mathbf{e}_{\mathrm{OBF}}\| \;=\; \|(I - \Pi_{U_k})\,\boldsymbol{\tau}\| .
\]
Choosing $U_k$ as the top-$k$ left singular subspace of the retained-orthogonal deleted-value matrix
\[
\mathbf{V}_{\mathrm{del}}^{\perp} \;:=\; (I - \Pi_{\mathrm{keep}})\,\mathbf{V}_{\mathrm{del}}, \qquad \Pi_{\mathrm{keep}} := \text{orthogonal projection onto } \mathrm{span}\{\mathbf{v}_j : j\in\mathcal{P}_{\mathrm{keep}}\},
\]
gives the tail-energy bound
\begin{equation}
\|\mathbf{e}_{\mathrm{OBF}}\| \;\le\; A_{\mathrm{del}}\,\sqrt{\sum_{j>k}\sigma_j^{\,2}\!\bigl(\mathbf{V}_{\mathrm{del}}^{\perp}\bigr)} ,
\label{eq:obf_tail_bound}
\end{equation}
where $\sigma_j(\cdot)$ denotes the $j$-th singular value.
\end{proposition}

Comparing Propositions~\ref{prop:evict_error} and~\ref{prop:obf_error}: hard eviction pays the full norm $A_{\mathrm{del}}\|\bar{\mathbf{v}}_{\mathrm{del}} - \bar{\mathbf{v}}_{\mathrm{keep}}\|$, while OBF pays only the tail energy of $\mathbf{V}_{\mathrm{del}}^{\perp}$. When the deleted-value information concentrates in a few directions outside $\mathrm{span}\{\mathbf{v}_j : j\in\mathcal{P}_{\mathrm{keep}}\}$, small $k$ suffices; when it is spread over many directions, OBF's advantage shrinks and its error approaches hard eviction's. The empirical Residual EVR reported in Appendix~\ref{app:internal_pca_rank_sweep} tracks exactly the fraction of $\sum_{j}\sigma_j^{\,2}(\mathbf{V}_{\mathrm{del}}^{\perp})$ captured by the chosen rank, so Eq.~\eqref{eq:obf_tail_bound} is directly measurable.

An exact, per-query dominance result follows by decomposing the correction target along the retained value subspace. Write $\boldsymbol{\tau} = \Pi_{\mathrm{keep}}\boldsymbol{\tau} + \boldsymbol{\tau}^{\perp}$ with $\boldsymbol{\tau}^{\perp} := (I-\Pi_{\mathrm{keep}})\boldsymbol{\tau}$. Since the OBF injection subspace $U_k$ is chosen inside $\mathrm{span}\{\mathbf{v}_j : j\in\mathcal{P}_{\mathrm{keep}}\}^{\perp}$, it is orthogonal to $\Pi_{\mathrm{keep}}\boldsymbol{\tau}$, and the OBF residual decomposes as $\mathbf{e}_{\mathrm{OBF}} = \Pi_{\mathrm{keep}}\boldsymbol{\tau} + (I-\Pi_{U_k})\boldsymbol{\tau}^{\perp}$, with the two terms orthogonal. Hence

\begin{corollary}[OBF dominates hard eviction]
\label{cor:obf_dominates}
For every query $\mathbf{q}$,
\[
\|\mathbf{e}_{\mathrm{OBF}}\|^{2} \;=\; \|\mathbf{e}_{\mathrm{evict}}\|^{2} \;-\; \|\Pi_{U_k}\boldsymbol{\tau}^{\perp}\|^{2} \;\le\; \|\mathbf{e}_{\mathrm{evict}}\|^{2},
\]
with strict inequality whenever $\Pi_{U_k}\boldsymbol{\tau}^{\perp}\neq\mathbf{0}$, i.e., whenever the retained-orthogonal component of the correction target has non-trivial overlap with the chosen injection subspace.
\end{corollary}

Two properties are worth highlighting. First, the parallel component $\|\Pi_{\mathrm{keep}}\boldsymbol{\tau}\|^{2}$ is never corrected by OBF — this is the intrinsic price of the orthogonality constraint, but it is also the component that hard eviction would already partially represent through the retained values, so the untreated residual here is benign. Second, the advantage $\|\Pi_{U_k}\boldsymbol{\tau}^{\perp}\|^{2}$ is directly witnessed by the Injection Vector Norm diagnostic in Appendix~\ref{app:internal_metrics}: a non-zero injection norm is equivalent to $\Pi_{U_k}\boldsymbol{\tau}^{\perp}\neq\mathbf{0}$, and is therefore a sufficient empirical condition for strict dominance.

\subsection{Why orthogonal: decoupling the correction from the amplification}
\label{app:error_bound:ortho}

Proposition~\ref{prop:amplification} shows that anything written into the retained value space is amplified by $1/(1-A_{\mathrm{del}})$ in the output. This motivates the orthogonality constraint in OBF. Decompose any injection as
\[
\mathbf{r}_j \;=\; \mathbf{r}_j^{\parallel} + \mathbf{r}_j^{\perp},
\qquad
\mathbf{r}_j^{\parallel} \in \mathrm{span}\{\mathbf{v}_k : k\in\mathcal{P}_{\mathrm{keep}}\},\
\mathbf{r}_j^{\perp} \in \mathrm{span}\{\mathbf{v}_k : k\in\mathcal{P}_{\mathrm{keep}}\}^{\perp} .
\]
The parallel component reshapes the already-amplified signal $\bar{\mathbf{v}}_{\mathrm{keep}}$: a perturbation $\boldsymbol{\varepsilon}$ in the estimated $\boldsymbol{\tau}$ that lies in the parallel subspace translates into an output perturbation
\[
\bigl\| \boldsymbol{\varepsilon}^{\parallel} \bigr\| \;\cdot\; \frac{1}{1 - A_{\mathrm{del}}},
\]
i.e., a small misestimation distorts the retained signal by the same amplification factor. In contrast, the orthogonal component contributes additively to an otherwise-empty subspace and does not interact with the amplified retained signal.

Constraining $\mathbf{r}_j \in \mathrm{span}\{\mathbf{v}_k : k\in\mathcal{P}_{\mathrm{keep}}\}^{\perp}$ therefore (i) guarantees that $\bar{\mathbf{v}}_{\mathrm{keep}}$ and the correction lie in orthogonal subspaces, so their norms add in quadrature rather than interfering; and (ii) isolates any estimation error along $\boldsymbol{\tau}$ to the same orthogonal subspace, preventing the amplification factor from propagating into the retained signal. This is the mechanism behind the clean tail-energy bound in Eq.~\eqref{eq:obf_tail_bound}: the residual depends only on $\mathbf{V}_{\mathrm{del}}^{\perp}$, never on the retained values themselves.

\subsection{Take-aways}
\label{app:error_bound:summary}


\begin{itemize}[leftmargin=1.2em]
    \item \textbf{Softmax renormalization creates amplification.} After deletion, each retained weight is scaled by $1/(1-A_{\mathrm{del}})$. Any compressor that writes into the retained value space inherits this factor, so the design of the correction must account for it.
    \item \textbf{Hard eviction is controlled by two quantities, not one.} Attention-based selection bounds only the deleted mass $A_{\mathrm{del}}$, not the value-space gap $\|\bar{\mathbf{v}}_{\mathrm{del}} - \bar{\mathbf{v}}_{\mathrm{keep}}\|$.
    \item \textbf{OBF replaces the gap with a tail-energy term.} The error bound in Eq.~\eqref{eq:obf_tail_bound} depends only on the singular tail of $\mathbf{V}_{\mathrm{del}}^{\perp}$, and is directly tracked by the Residual EVR diagnostic.
    \item \textbf{Orthogonality is not cosmetic.} Injecting orthogonally to the retained value span is precisely what decouples the correction from the softmax amplification and yields an error bound that does not reference the retained values.
\end{itemize}
\clearpage

\section{Token Selection Visualization Across Granularities}
\label{app:token_selection}
We visualize the prompt tokens selected for relay under the layerwise (L) and headwise (H) granularities defined in Section~\ref{subsec:h2o_selection}, on a representative ARC-C sample. For each of the three intermediate agents (Planner, Critic, Refiner) and each granularity, we show an early, middle, and late layer to illustrate how retention varies with depth. Red highlights mark attention sinks; yellow highlights mark tokens ranked by attention mass. Note that this appendix uses the Qwen3-8B backbone (rather than the Qwen3-4B backbone used in the main experiments); all other settings follow Appendix~\ref{app:ablation_setup} except for the overrides below.
\begin{lstlisting}[basicstyle=\ttfamily\small, frame=single]
{
  "model_name": "Qwen/Qwen3-8B",
  "seed": 888,
  "device": "NVIDIA RTX 6000 Ada Generation GPU"
}
\end{lstlisting}

\FloatBarrier
\subsection{Layerwise Token Selection}

\begin{figure}[!htbp]
    \centering
    \begin{subfigure}{0.95\textwidth}
        \centering
        \includegraphics[width=\textwidth]{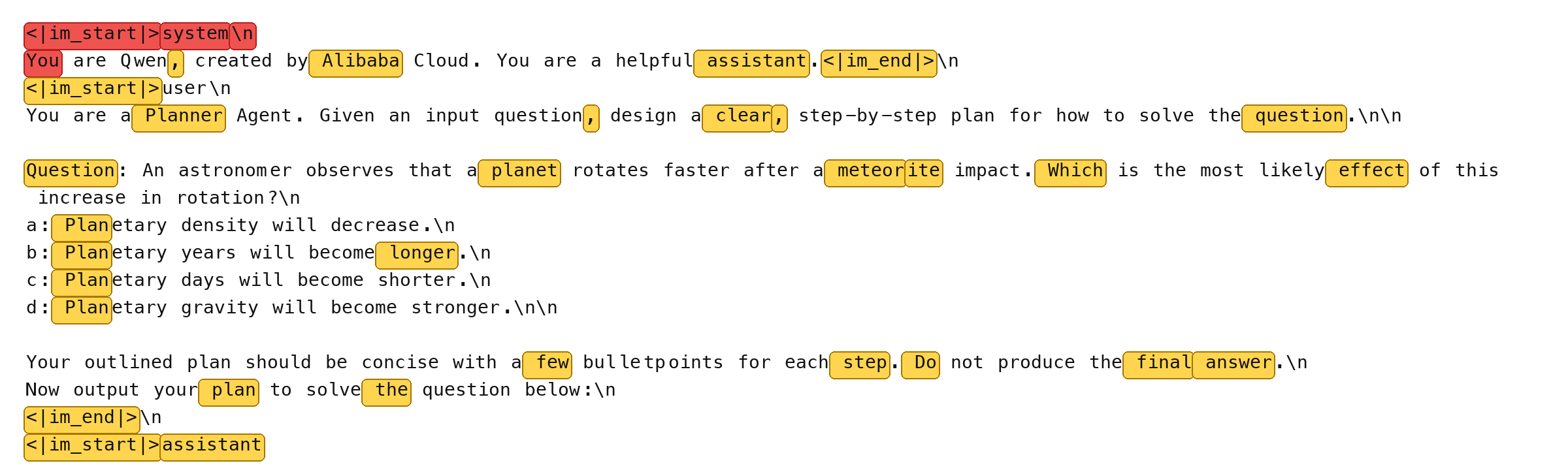}
        \caption{Early layer (layer 0).}
    \end{subfigure}

    \vspace{0.5em}

    \begin{subfigure}{0.95\textwidth}
        \centering
        \includegraphics[width=\textwidth]{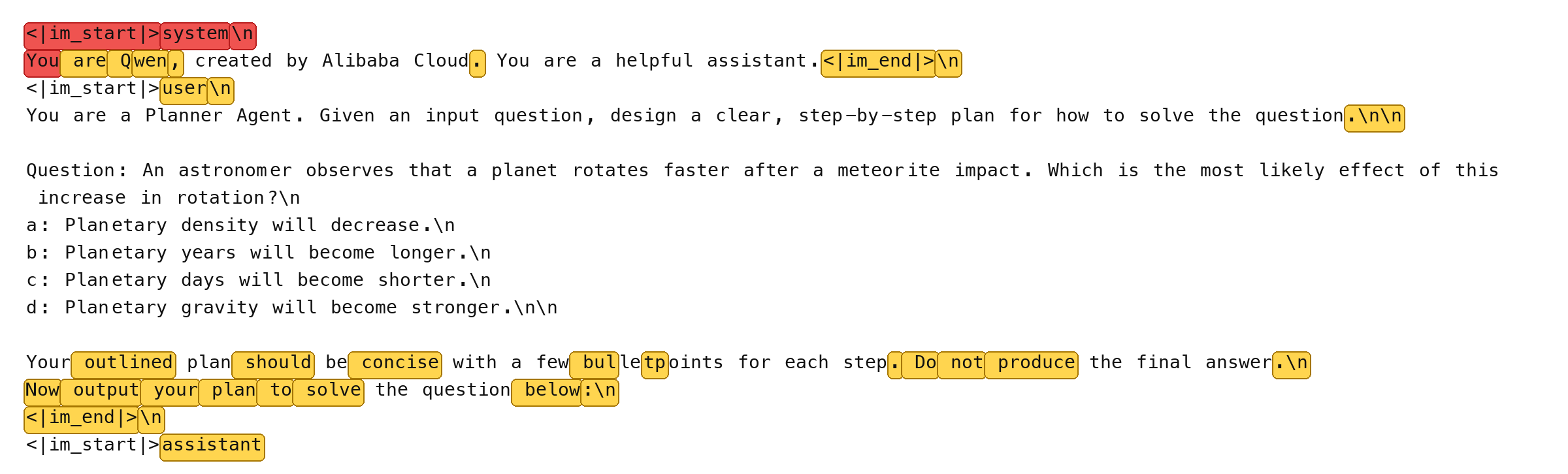}
        \caption{Middle layer (layer 17).}
    \end{subfigure}

    \vspace{0.5em}

    \begin{subfigure}{0.95\textwidth}
        \centering
        \includegraphics[width=\textwidth]{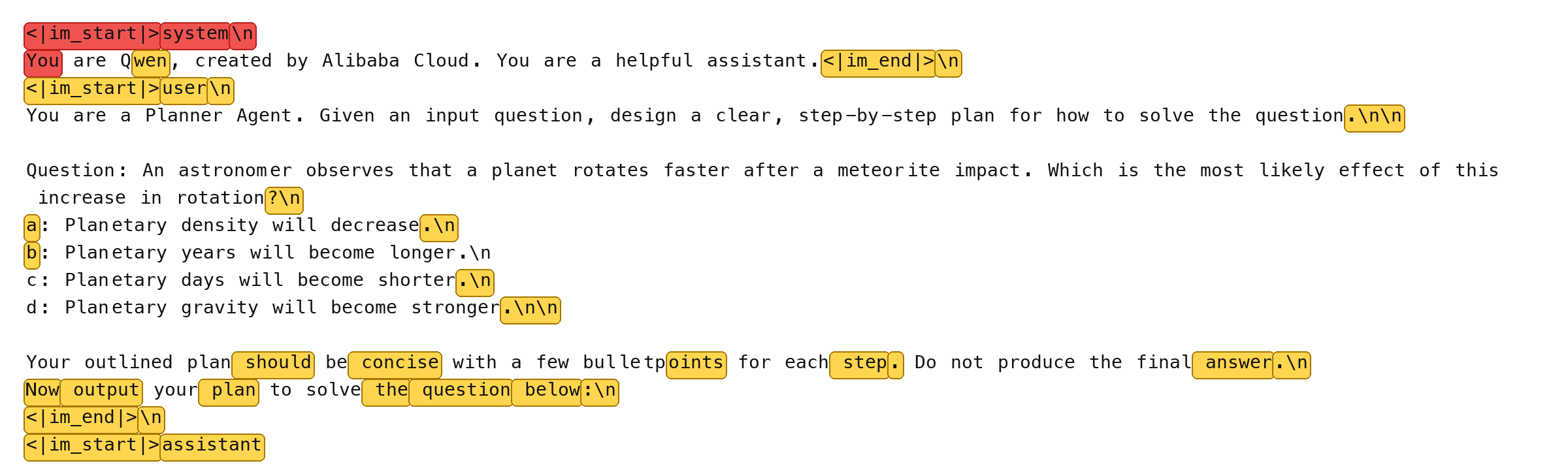}
        \caption{Late layer (layer 35).}
    \end{subfigure}
    \caption{Layerwise token selection for the Planner agent.}
    \label{fig:layerwise-planner}
\end{figure}

\begin{figure}[!htbp]
    \centering
    \begin{subfigure}{0.95\textwidth}
        \centering
        \includegraphics[width=\textwidth]{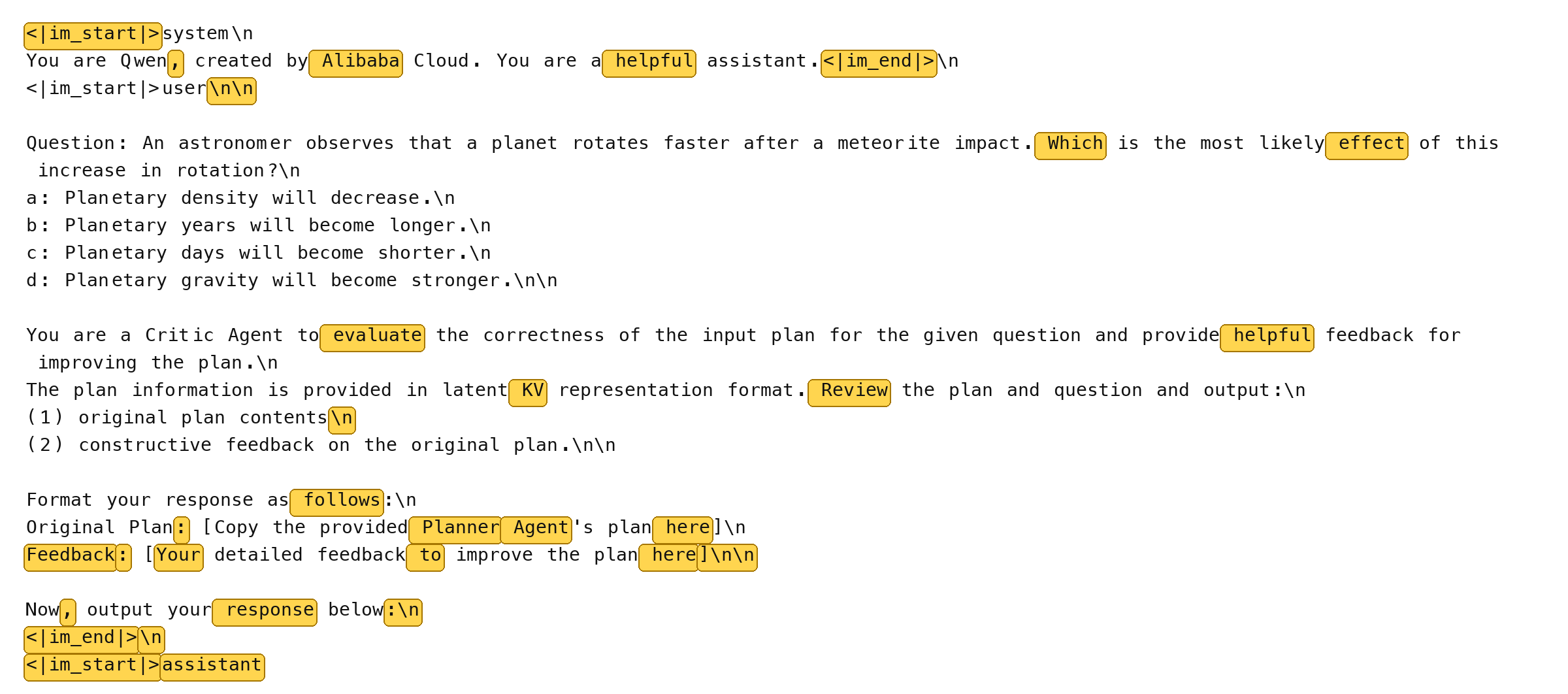}
        \caption{Early layer (layer 0).}
    \end{subfigure}

    \vspace{0.5em}

    \begin{subfigure}{0.95\textwidth}
        \centering
        \includegraphics[width=\textwidth]{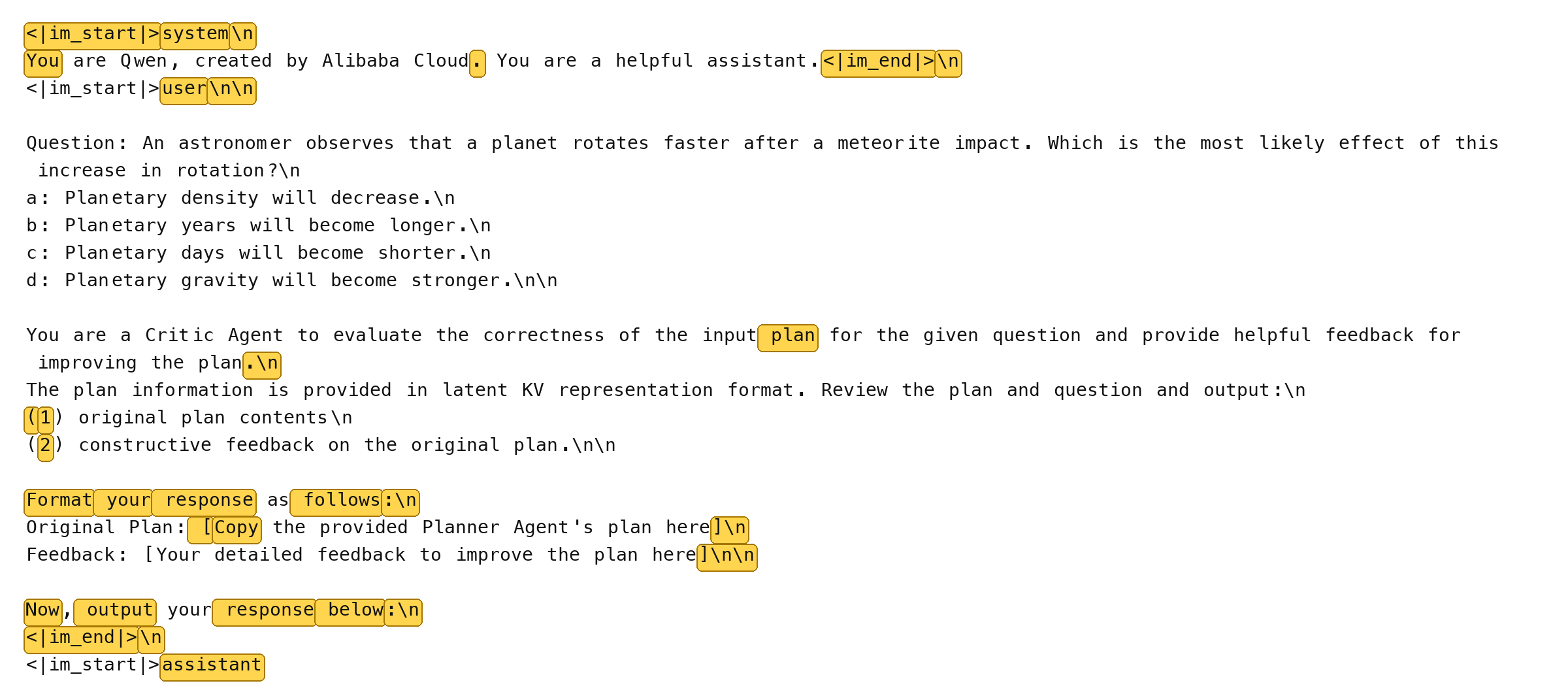}
        \caption{Middle layer (layer 17).}
    \end{subfigure}

    \vspace{0.5em}

    \begin{subfigure}{0.95\textwidth}
        \centering
        \includegraphics[width=\textwidth]{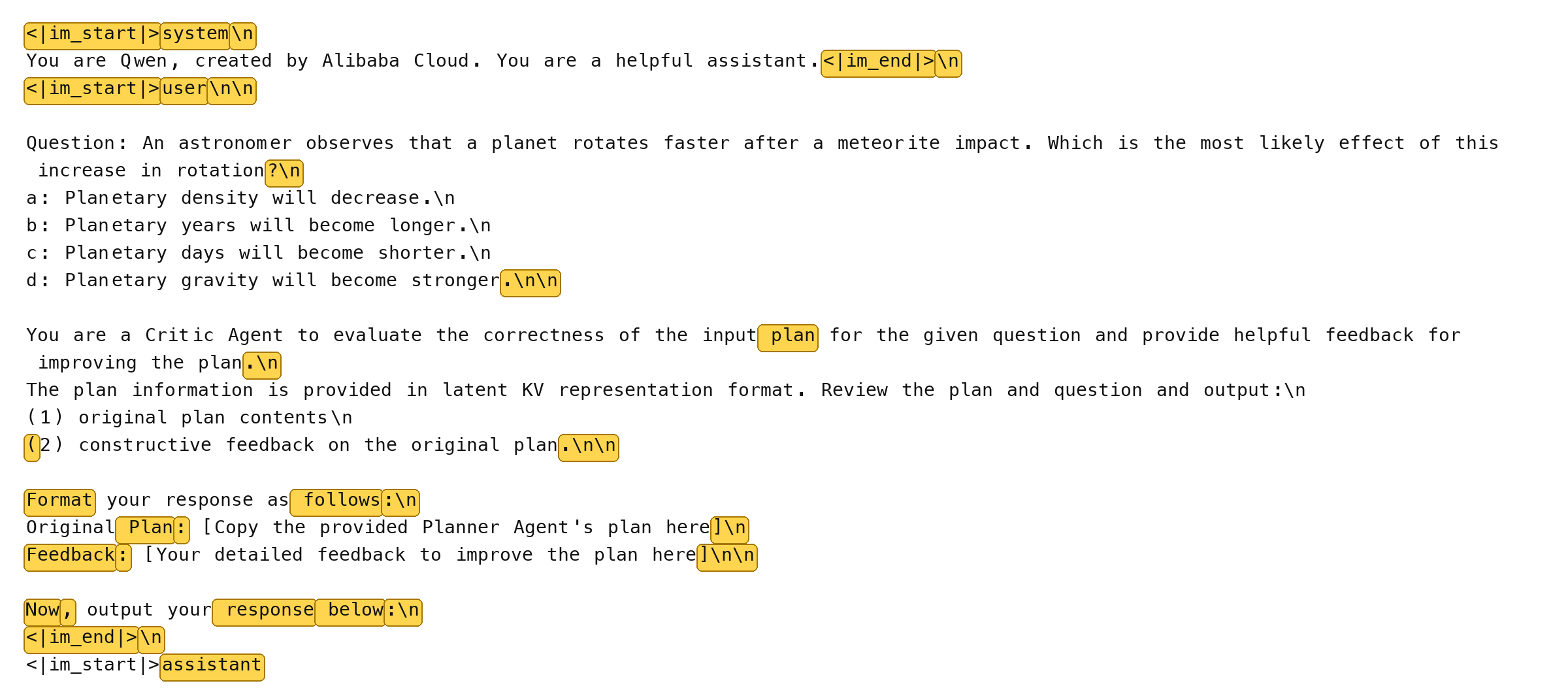}
        \caption{Late layer (layer 35).}
    \end{subfigure}
    \caption{Layerwise token selection for the Critic agent.}
    \label{fig:layerwise-critic}
\end{figure}

\begin{figure}[!htbp]
    \centering
    \begin{subfigure}{0.95\textwidth}
        \centering
        \includegraphics[width=\textwidth]{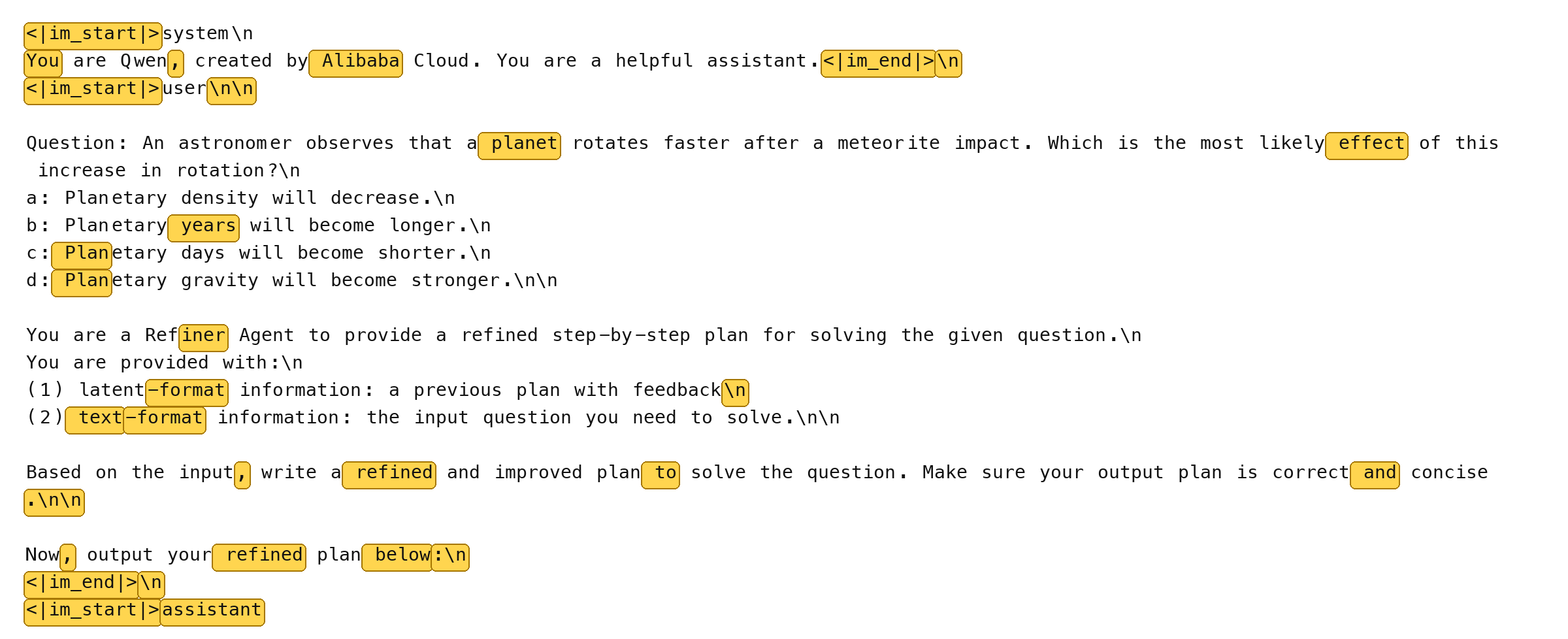}
        \caption{Early layer (layer 0).}
    \end{subfigure}

    \vspace{0.5em}

    \begin{subfigure}{0.95\textwidth}
        \centering
        \includegraphics[width=\textwidth]{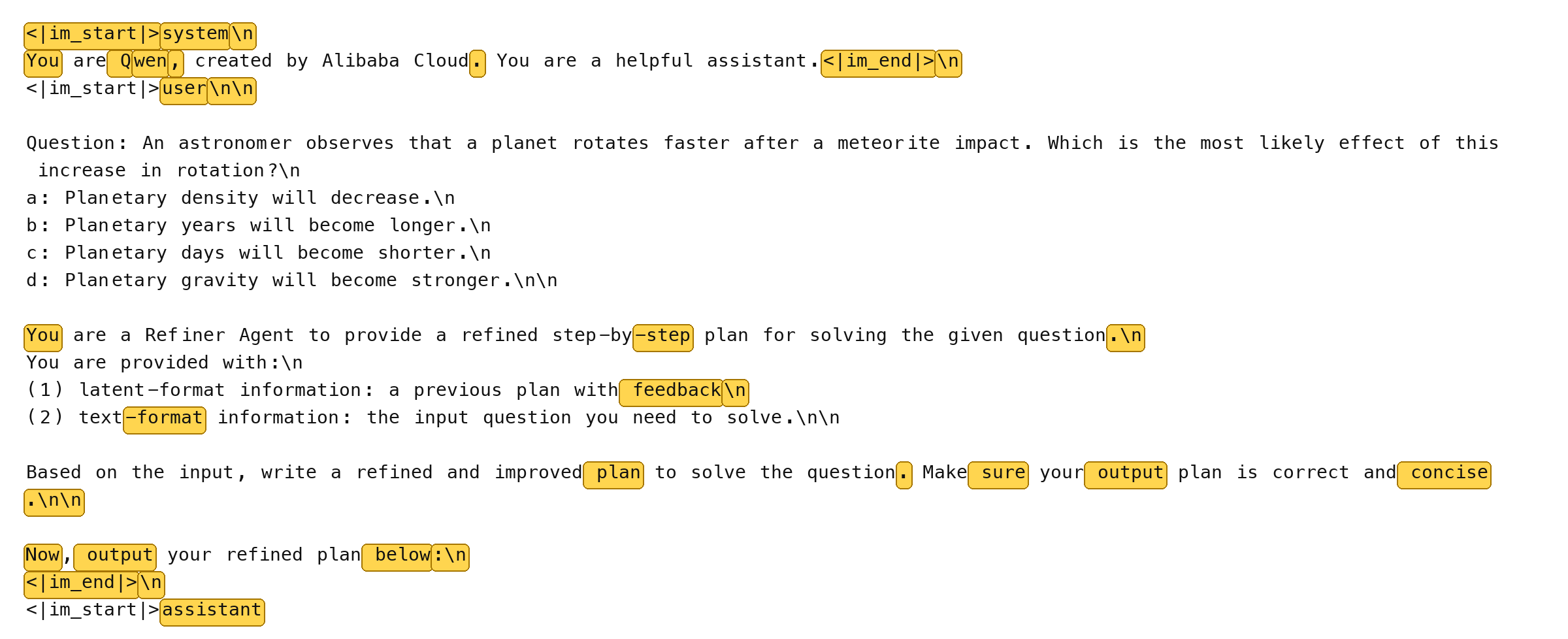}
        \caption{Middle layer (layer 17).}
    \end{subfigure}

    \vspace{0.5em}

    \begin{subfigure}{0.95\textwidth}
        \centering
        \includegraphics[width=\textwidth]{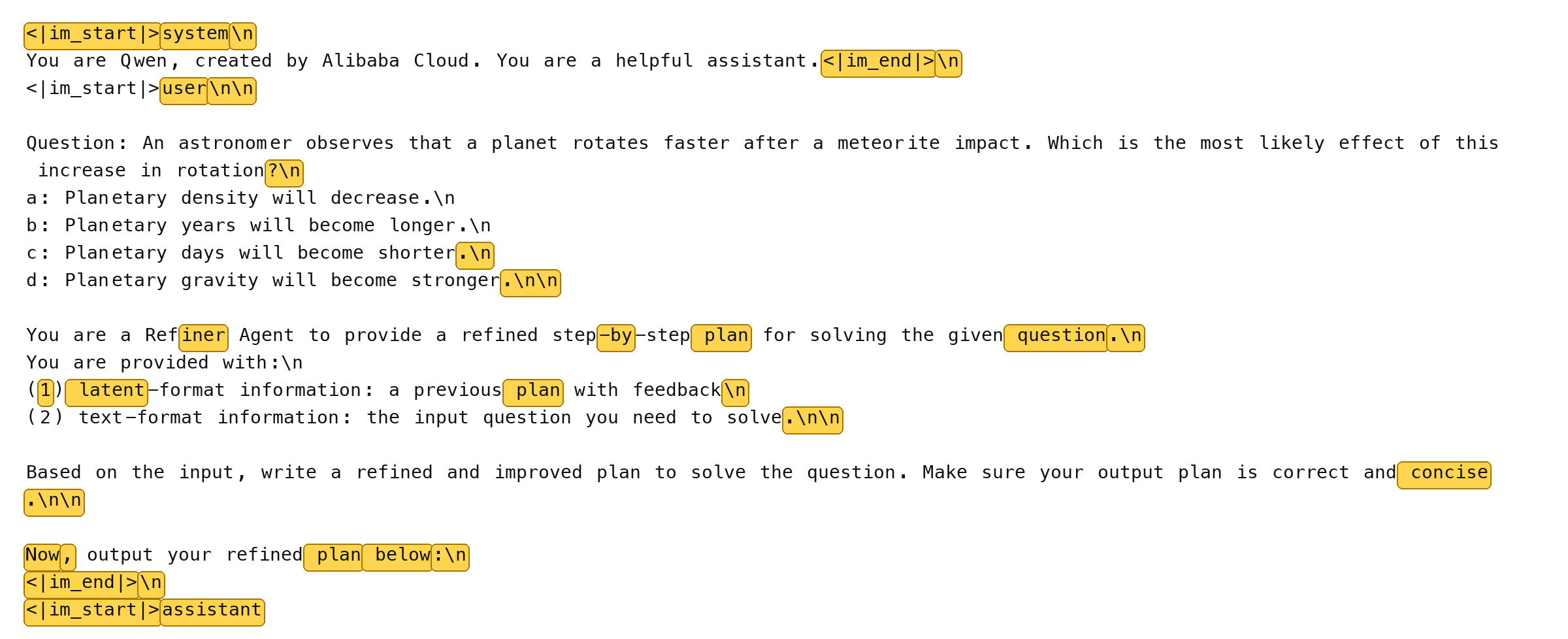}
        \caption{Late layer (layer 35).}
    \end{subfigure}
    \caption{Layerwise token selection for the Refiner agent.}
    \label{fig:layerwise-refiner}
\end{figure}

Figures~\ref{fig:layerwise-planner}, \ref{fig:layerwise-critic}, and \ref{fig:layerwise-refiner} show layerwise selection for the Planner, Critic, and Refiner agents respectively.

\FloatBarrier

\FloatBarrier

\FloatBarrier
\clearpage
\subsection{Headwise Token Selection}

\begin{figure}[!htbp]
    \centering
    \begin{subfigure}{0.95\textwidth}
        \centering
        \includegraphics[width=\textwidth]{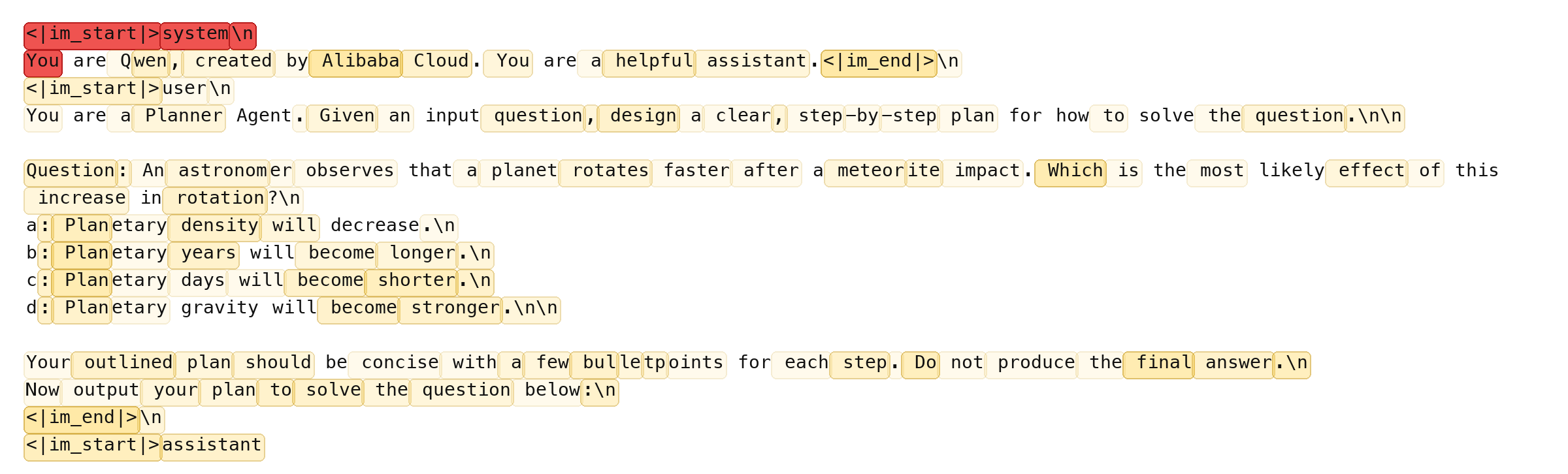}
        \caption{Early layer (layer 0).}
    \end{subfigure}

    \vspace{0.5em}

    \begin{subfigure}{0.95\textwidth}
        \centering
        \includegraphics[width=\textwidth]{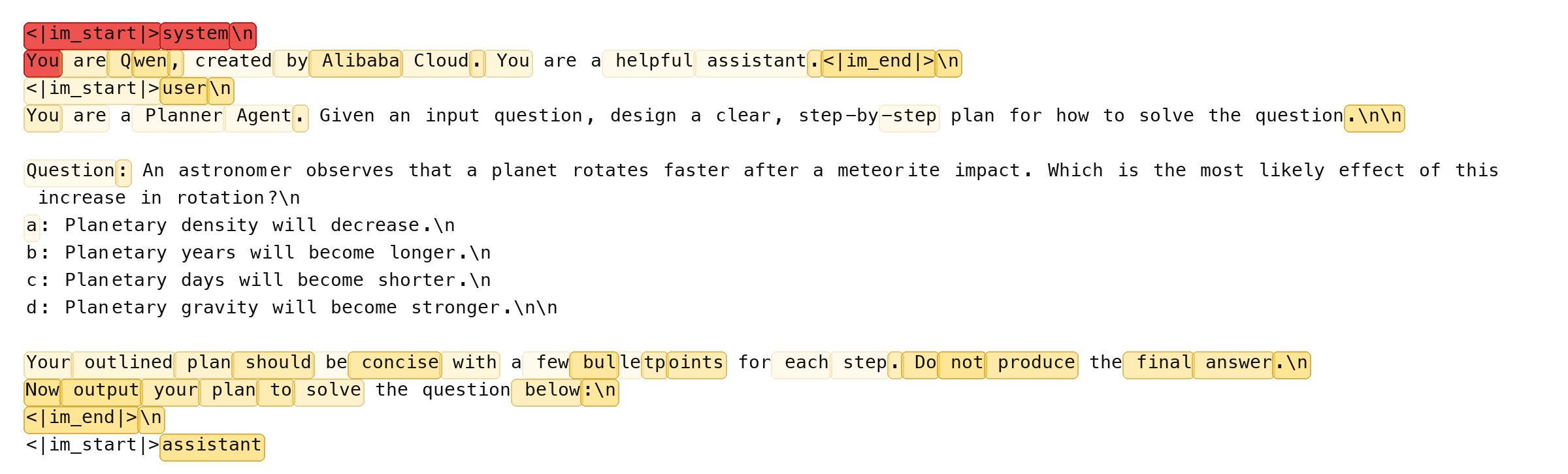}
        \caption{Middle layer (layer 17).}
    \end{subfigure}

    \vspace{0.5em}

    \begin{subfigure}{0.95\textwidth}
        \centering
        \includegraphics[width=\textwidth]{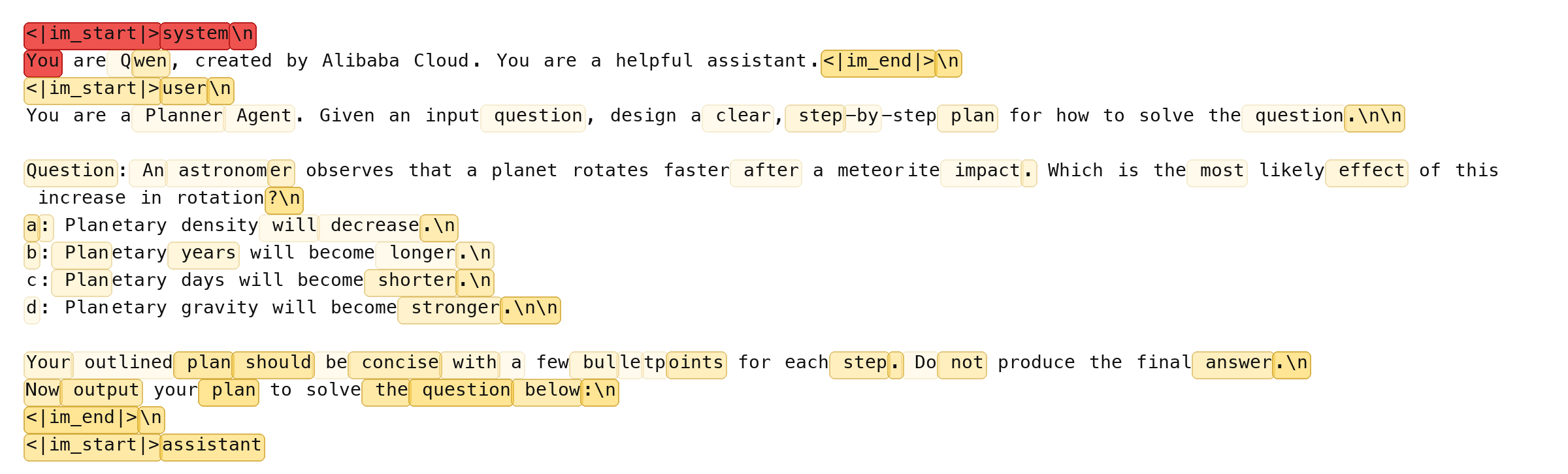}
        \caption{Late layer (layer 35).}
    \end{subfigure}
    \caption{Headwise token selection for the Planner agent.}
    \label{fig:headwise-planner}
\end{figure}

\begin{figure}[!htbp]
    \centering
    \begin{subfigure}{0.95\textwidth}
        \centering
        \includegraphics[width=\textwidth]{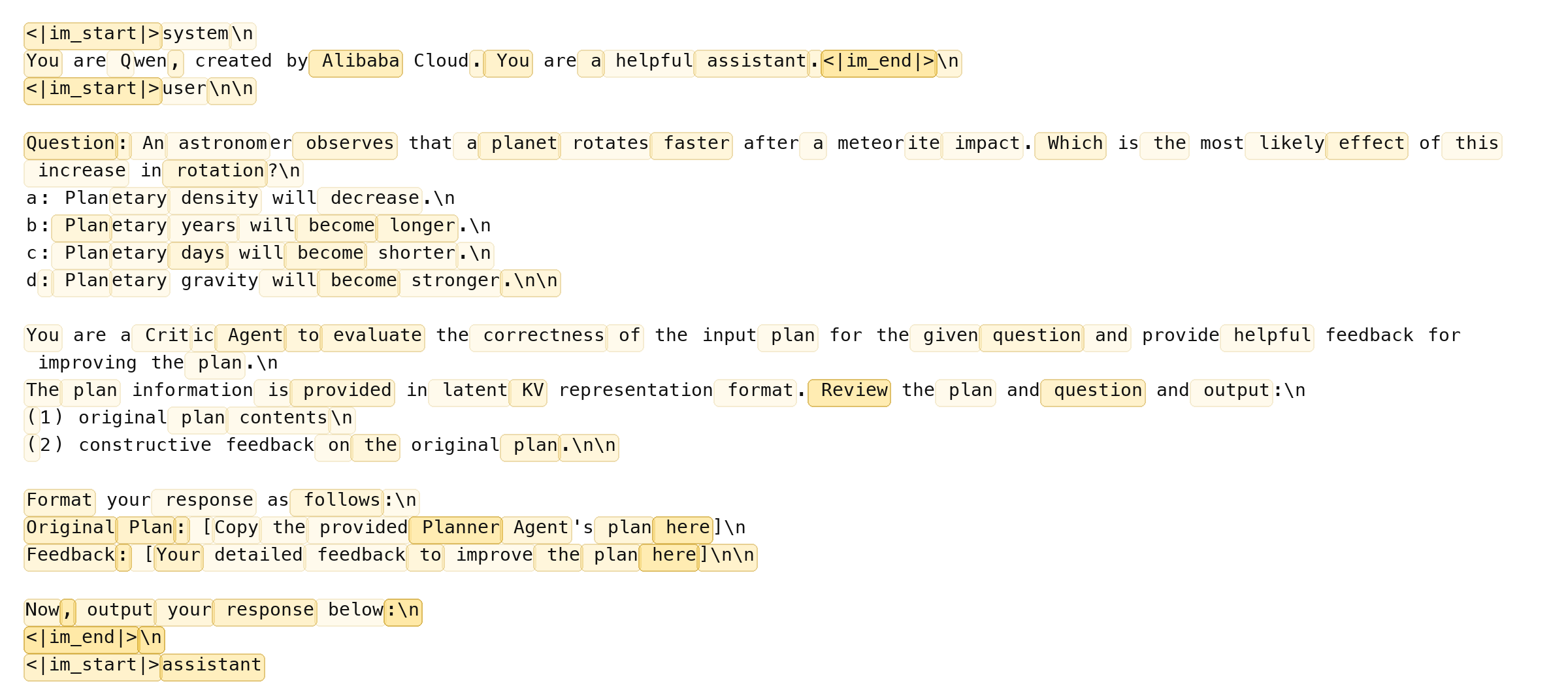}
        \caption{Early layer (layer 0).}
    \end{subfigure}

    \vspace{0.5em}

    \begin{subfigure}{0.95\textwidth}
        \centering
        \includegraphics[width=\textwidth]{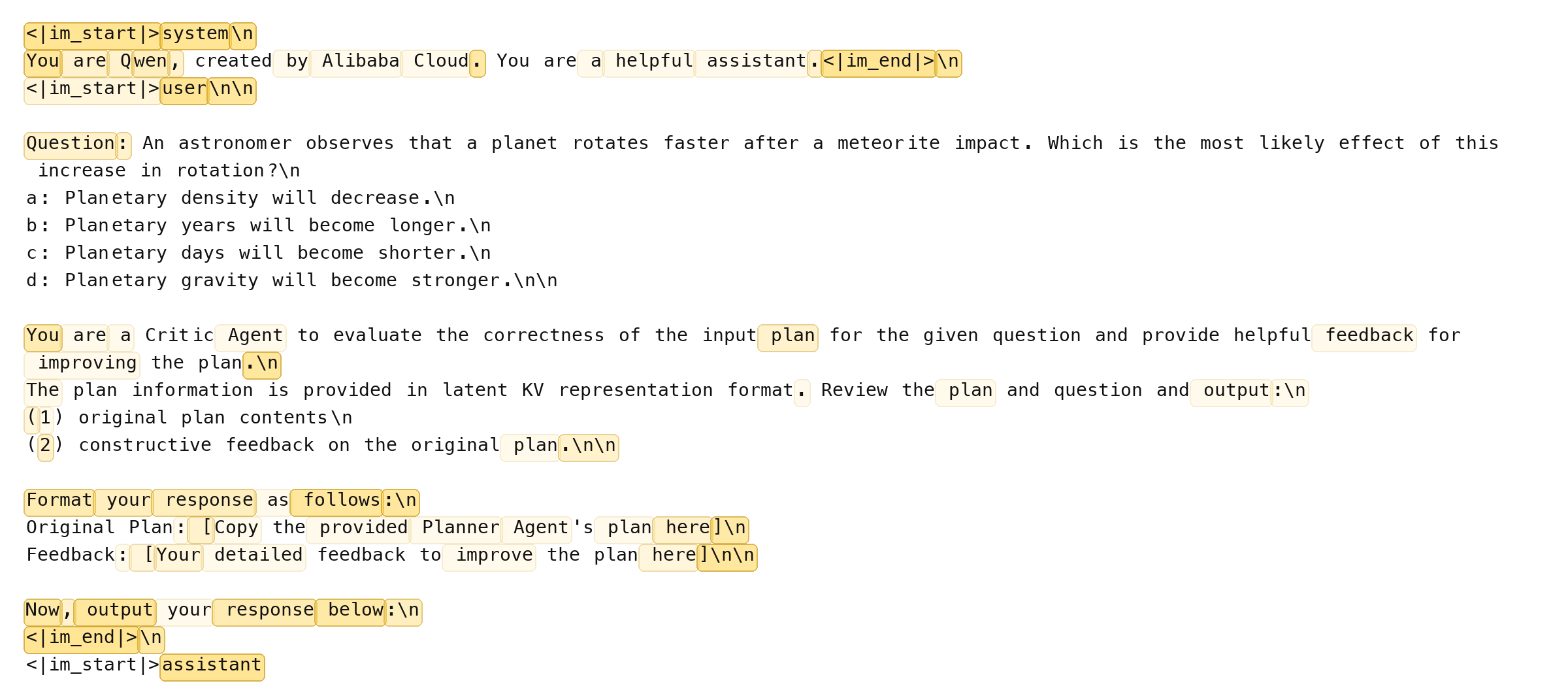}
        \caption{Middle layer (layer 17).}
    \end{subfigure}

    \vspace{0.5em}

    \begin{subfigure}{0.95\textwidth}
        \centering
        \includegraphics[width=\textwidth]{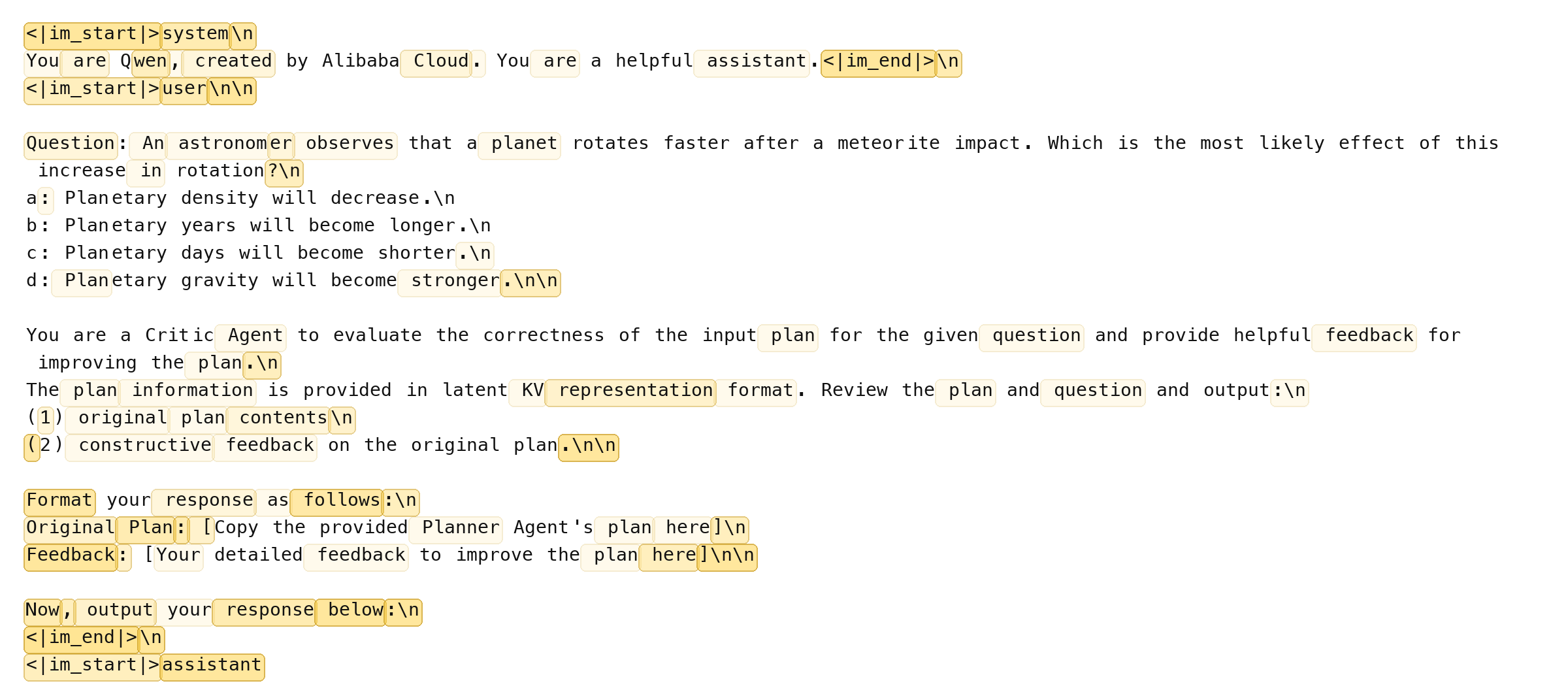}
        \caption{Late layer (layer 35).}
    \end{subfigure}
    \caption{Headwise token selection for the Critic agent.}
    \label{fig:headwise-critic}
\end{figure}

\begin{figure}[!htbp]
    \centering
    \begin{subfigure}{0.95\textwidth}
        \centering
        \includegraphics[width=\textwidth]{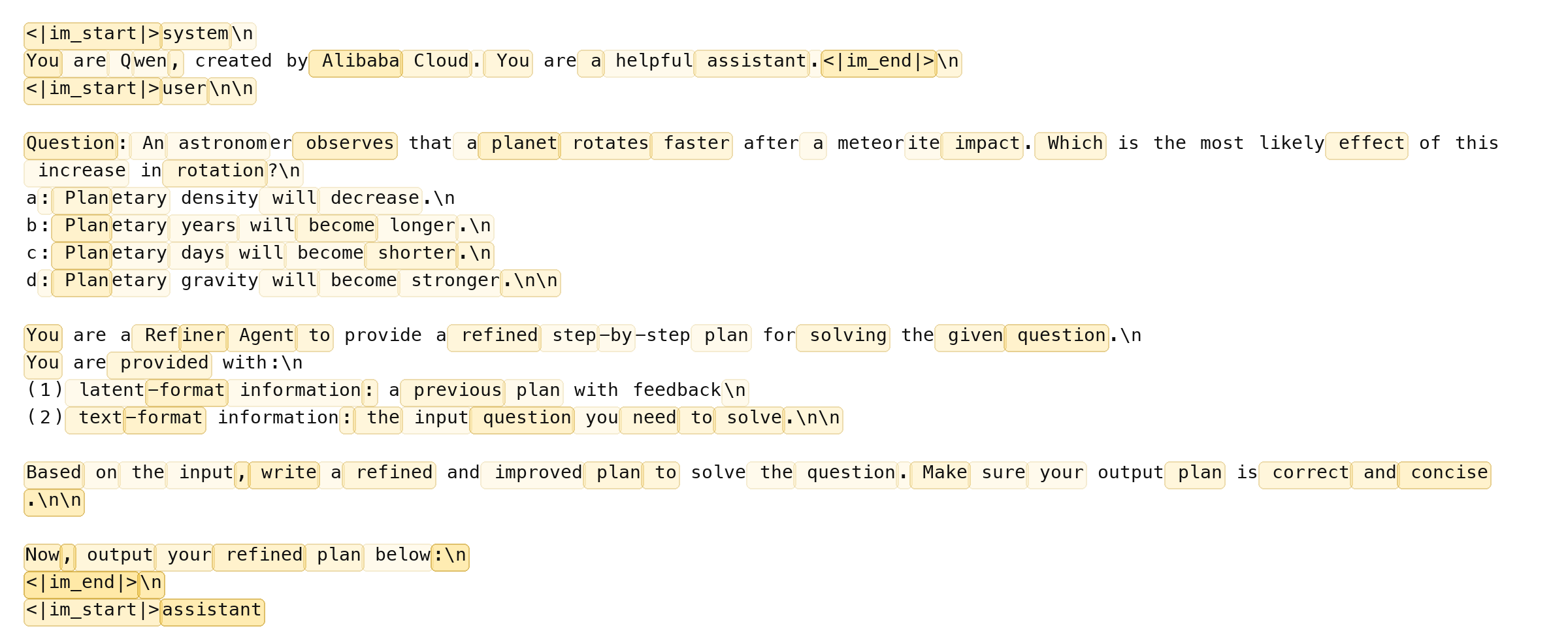}
        \caption{Early layer (layer 0).}
    \end{subfigure}

    \vspace{0.5em}

    \begin{subfigure}{0.95\textwidth}
        \centering
        \includegraphics[width=\textwidth]{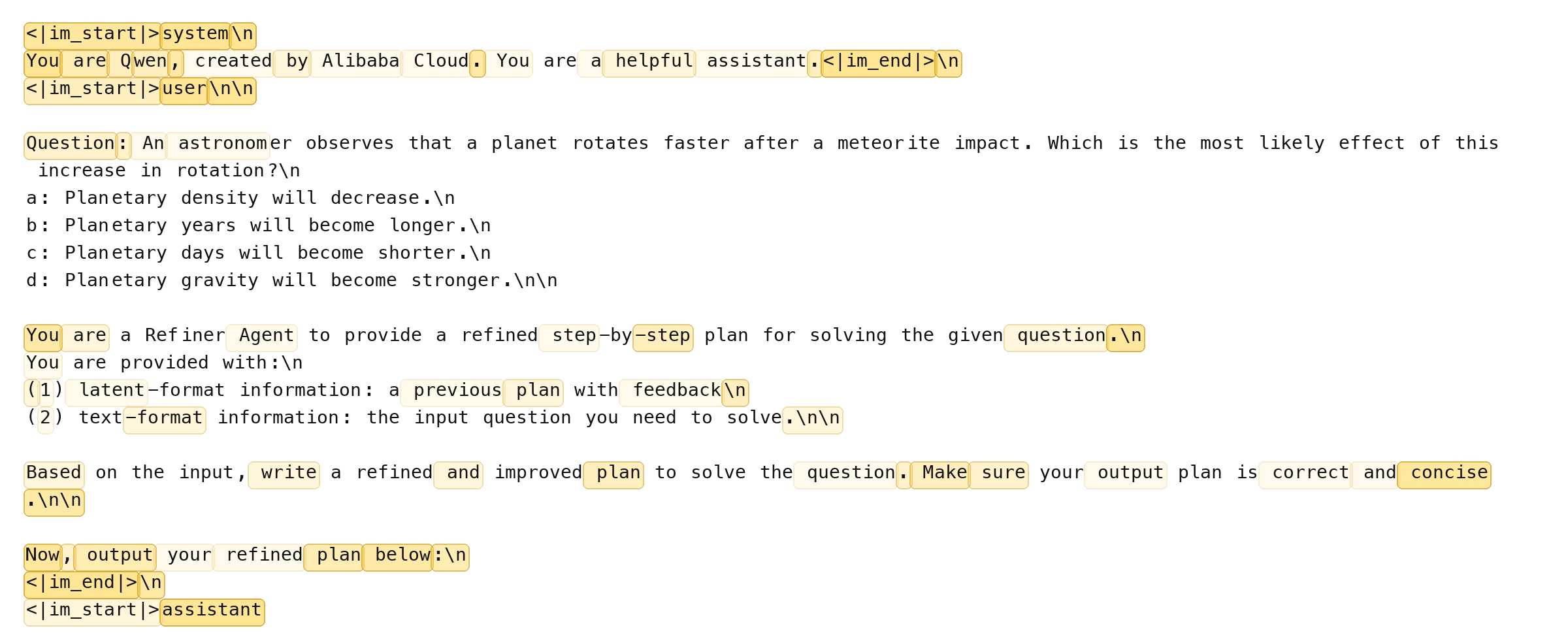}
        \caption{Middle layer (layer 17).}
    \end{subfigure}

    \vspace{0.5em}

    \begin{subfigure}{0.95\textwidth}
        \centering
        \includegraphics[width=\textwidth]{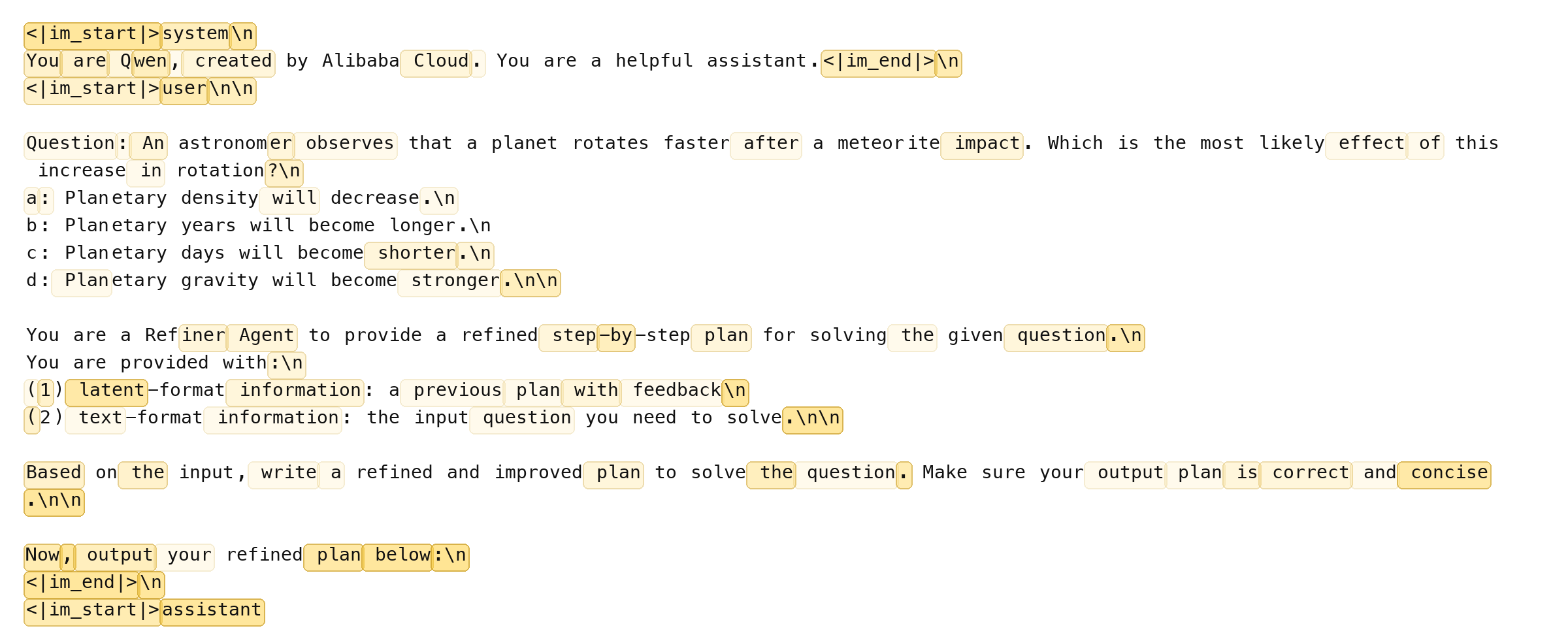}
        \caption{Late layer (layer 35).}
    \end{subfigure}
    \caption{Headwise token selection for the Refiner agent.}
    \label{fig:headwise-refiner}
\end{figure}

Figures~\ref{fig:headwise-planner}, \ref{fig:headwise-critic}, and \ref{fig:headwise-refiner} show headwise selection for the Planner, Critic, and Refiner agents respectively.

\FloatBarrier

\FloatBarrier

\FloatBarrier
\clearpage

\newpage
\end{document}